\documentclass{article}

\usepackage[final]{neurips_2024}

\usepackage{url}
\usepackage{graphicx}

\usepackage{subfigure}

\usepackage{amssymb}
\usepackage{algorithm}
\usepackage{algpseudocode}
\usepackage{amsmath}
\usepackage{ntheorem}
\usepackage{booktabs} % For formal tables

\usepackage{hyperref}
\usepackage{xcolor}
\hypersetup{
	colorlinks,
	linkcolor={red!50!black},
	citecolor={blue!50!black},
	urlcolor={blue!80!black}
}

\newtheorem{definition}{Definition} % 创建一个名为“Definition”的新定理环境
\newtheorem{lemma}{Lemma} % 创建一个名为“Corollary”的新定理环境
\newtheorem{theorem}{Theorem} % 创建一个名为“Proposition”的新定理环境
\title{Preventing Model Collapse in Deep Canonical Correlation Analysis by Noise Regularization}

\author{%
  Junlin He \\
  The Hong Kong Polytechnic University\\
  Hong Kong SAR, China \\
  \texttt{junlinspeed.he@connect.polyu.hk} \\
  \And
  Jinxiao Du \\
  The Hong Kong Polytechnic University\\
  Hong Kong SAR, China \\
  \texttt{jinxiao.du@connect.polyu.hk} \\
  \And
  \And
  Susu Xu \\
  Johns Hopkins University\\
  Maryland, USA \\
  \texttt{sxu83@jhu.edu} \\
  \And
  Wei Ma\thanks{Corresponding author.}\\
  The Hong Kong Polytechnic University\\
  Hong Kong SAR, China \\
  \texttt{wei.w.ma@polyu.edu.hk} \\
}

\begin{document}

\maketitle

\begin{abstract}

Multi-View Representation Learning (MVRL) aims to learn a unified representation of an object from multi-view data.
% into a unified space that reveals complex relationships. 
Deep Canonical Correlation Analysis (DCCA) and its variants share simple formulations and demonstrate state-of-the-art performance. However, with extensive experiments, we observe the issue of model collapse, {\em i.e.}, the performance of DCCA-based methods will drop drastically when training proceeds. The model collapse issue could significantly hinder the wide adoption of DCCA-based methods because it is challenging to decide when to early stop. To this end, we develop NR-DCCA, which is equipped with a novel noise regularization approach to prevent model collapse. Theoretical analysis shows that the Correlation Invariant Property is the key to preventing model collapse, and our noise regularization forces the neural network to possess such a property. A framework to construct synthetic data with different common and complementary information is also developed to compare MVRL methods comprehensively. The developed NR-DCCA outperforms baselines stably and consistently in both synthetic and real-world datasets, and the proposed noise regularization approach can also be generalized to other DCCA-based methods such as DGCCA. Our code will be released at  \href{https://github.com/Umaruchain/NR-DCCA.git}{https://github.com/Umaruchain/NR-DCCA.git}.

%, utilize the powerful nonlinear transformation capabilities of neural networks to project multi-view data into space with maximum correlation. Despite achieving certain accomplishments, these methods are not able to avoid overfitting and be robust to multi-view data with varying degrees of correlation. In this paper, we propose NR-DCCA, which combines noise regularization with DCCA, imposing implicit constraints on the neural networks without altering their structure. We also construct MVRL synthetic datasets, allowing us to evaluate MVRL methods comprehensively. Through experiments conducted on both synthetic datasets and real-world datasets, our method not only achieves competitive performance but also addresses the issues of overfitting and lack of robustness.
\end{abstract}

% \begin{keyword}
% {\bf Keywords:} Multi-view representation learning; Canonical Correlation Analysis; Deep Canonical Correlation Analysis; Noise regularization; Model collapse
% \end{keyword}

\section{Introduction}

In recent years, multi-view representation learning (MVRL) has emerged as a core technology for learning from multi-source data and providing readily useful representations to downstream tasks~\citep{sun2023learning,yan2021deep}, and it has achieved tremendous success in various applications, such as
% In recent decades, multi-view data has become an increasingly popular form of data in many fields, 
video surveillance~\citep{guo2015learning,feichtenhofer2016convolutional,2021Deep},  medical diagnosis~\citep{wei2019m3net,xu2020recognition} and social media~\citep{srivastava2012multimodal,karpathy2015deep,mao2014deep,fan2020deep}. 
Specifically, multi-source data can be collected from the same object, and each data source can be regarded as one view of the object. 
% Generally speaking, multi-view data refers to multiple features of the same object, which are captured from different data sources or through diverse description methods within the same data source. 
For instance, an object can be described simultaneously through texts, videos, and audio, which contain both common and complementary information of the object~\citep{yan2021deep,zhang2019ae2,hwang2021multi,geng2021uncertainty}, and the MVRL aims to learn a unified representation of the object from the multi-view data. 

% while features of an image can be extracted using multiple visual feature extraction methods. The diverse features from these different views contain both common and complementary information, which means that utilizing them directly may not lead to satisfactory performance in various downstream tasks  (e.g. image classification). 
% Therefore, many studies~\citep{kettenring1971canonical,zhang2016multi,sun2010scalable,avron2013efficient,zhang2021crowd} have been conducted on handling multi-view data and generating new representations for objects in an unsupervised way, which is called multi-view representation learning (MVRL). 

The key challenge of MVRL is to learn the intricate relationships of different views. The Canonical Correlation Analysis (CCA), which is one of the early and representative methods for MVRL, transforms all the views into a unified space by maximizing their correlations~\citep{hotelling1992relations,horst1961generalized,hardoon2004canonical,lahat2015multimodal,yan2023robust,sun2023learning}. 
% One of the early MVRL methods that gain significant attention is Canonical Correlation Analysis (CCA)~\citep{kettenring1971canonical}. CCA learns projection matrices that linearly map multi-view data into a shared space by maximizing the correlation among multiple views. 
Through correlation maximization, CCA can identify the common information between different views and extract them to form the representation of the object. On top of CCA, Linear CCA, and DCCA maximize the correlation defined by CCA through gradient descent, while the former uses an affine transformation and the latter uses Deep Neural Networks (DNNs).~\citep{andrew2013deep}.
Indeed, there are quite a few variants of DCCA, such as DGCCA~\citep{benton2017deep}, DCCAE~\citep{wang2015deep}, DVCCA~\citep{wang2016deep}, DTCCA~\citep{wong2021deep} and DCCA\_GHA~\citep{chapman2022generalized}.

However, extensive experimentation reveals that \textbf{DCCA-based methods typically excel during the initial stages of training but suffer a significant decline in performance as training progresses}. This phenomenon is defined as model collapse within the context of DCCA. Notably, our definition is grounded in the performance of the learned representations on downstream tasks. 
Previous studies found that the representations (i.e., final output) of both Linear CCA and DCCA are full-rank~\citep{andrew2013deep,de2003regularization}. Nevertheless, they did not further explore whether merely guaranteeing that the full-rank representations can guarantee that the weight matrices are full-rank.

Though early stopping could be adopted to prevent model collapse \citep{prechelt1998automatic,yao2007early}, it remains challenging when to stop. The model collapse issue of DCCA-based methods prevents the adoption in large models, and currently, many applications still use simple concatenation to combine different views~\citep{yan2021deep,zheng2020feature,nie2017self}.  
Therefore, how to develop a DCCA-based MVRL method free of model collapse remains an interesting and open question.

In this work, we demonstrate that both representations and weight matrices of Linear CCA are full-rank whereas DCCA only guarantees that representations are full-rank but not for the weight matrices. 
Considering that Linear CCA does not show the model collapse while DCCA does, we conjecture that the root cause of the model collapse in DCCA is that the weight matrices in DNNs tend to be low-rank. 
A wealth of research supports this assertion, both theoretically and empirically, demonstrating that over-parameterized DNNs are predisposed to discovering low-rank solutions~\citep{jing2021understanding, saxe2019mathematical, soudry2018implicit, dwibedi2021little}. If the weight matrices in DNNs tend to be low-rank, it means that the weight matrices are highly self-related and redundant, which limits the expressiveness of DNNs and thus affects the quality of representations.

% Though early stopping could be adopted \citep{prechelt1998automatic,yao2007early,prechelt2002early,tian2022comprehensive,mahsereci2017early,bonet2021channel} 

% The correlation learned from DCCA may not come from the views, and it may be attributed to the correlation between the two neural networks. As   

Therefore, this paper develops NR-DCCA, a DCCA-based method equipped with a generalized noise regularization (NR) approach. 
% The proposed noise regularization approach constrains the DNNs to possess CIP, so as to prevent model collapse.
The NR approach ensures that the correlation with random data is invariant before and after the transformation, which we define as the Correlation Invariant Property (CIP).
It is also verified that the NR approach can be applied to other DCCA-based methods.
% To justify the approach, the formulation of CCA is analyzed and rigorous proofs are provided. 
Comprehensive experiments using both synthetic datasets and real-world datasets demonstrate the consistent outperformance and stability of the developed NR-DCCA method.

% We prove that the main reason for Linear CCA not having the model collapse issue is the Correlation Invariant Property(CIP), 
From a theoretical perspective, we derive the equivalent conditions between the full-rank property and CIP of the weight matrix. 
% The CIP defines the condition that 
By forcing DNNs to possess CIP and thus mimicking the behavior of Linear CCA, we introduce random data to constrain the weight matrices in DNNs and expect to avoid them being redundant and thus prevent model collapse.
% We further demonstrate that the NR approach could enforce DNNs to possess CIP, and therefore, model collapse can be prevented.

% Empirically, we observe that this strategy prevents model collapse by ensuring that the weight matrices in the networks are less redundant, which is a primary contributor to model collapse. 
% Note that the proposed noise regularization approach is novel and particularly tailored for DCCA-based methods, which is different from the existing approaches that directly inject noise into the neural networks~\citep{poole2014analyzing,he2019parametric,gong2020maxup}.

% the CCA-based methods 

% This paper advocates a novel idea to employ the noise regularization (NR) in order to prevent model collapse when training the DCCA-based methods. 

% One popular approach to improve training stability and robustness of neural networks is to employ noise regularization during the training process~\citep{bishop1995training,an1996effects,sietsma1991creating}. The most common way is to inject noise into the training data~\citep{poole2014analyzing,he2019parametric,gong2020maxup}.
% In supervised tasks, such a technique can avoid models from overfitting, since models always train in unseen data. In unsupervised tasks, current noise regularization methods must rely on reconstruction loss in autoencoders to impose implicit constraints. Since MVRL is a typical unsupervised task, unless we use additional autoencoder structures, there is no noise regularization method for DCCA-based methods.

In summary, our contributions are four-fold:
% To justify the approach, the formulation of CCA is analyzed and rigorous proofs are provided. Comprehensive experiments using both synthetic datasets and real-world datasets demonstrate the consistent outperformance and stability of the developed NR-DCCA method. 

% In this paper, we focus on ensuring the reliability of DCCA-based methods. Motivated by the success of noise regularization, we use the CCA formula to propose a novel noise regularization approach. We refer to DCCA equipped with our noise regularization approach as NR-DCCA. Our contributions are three-fold:

\begin{itemize}
\item The model collapse issue in DCCA-based methods for MVRL is identified, demonstrated, and explained.
\item A simple yet effective noise regularization approach is proposed and NR-DCCA  is developed to prevent model collapse. Comprehensive experiments using both synthetic datasets and real-world datasets demonstrate the consistent outperformance and stability of the developed NR-DCCA.
\item Rigorous proofs are provided to demonstrate that CIP is the equal condition of the full-rank weight matrix, which justifies the developed NR approach from a theoretical perspective.
\item A novel framework is proposed to construct synthetic data with different common and complementary information for comprehensively evaluating MVRL methods.
% To address the gap of not being able to comprehensively evaluate the reliability of MVRL methods, we define the components that constitute the MVRL benchmarks and thus release the first artificially generated MVRL benchmarks. These benchmarks cover a wide range of scenarios, encompassing various degrees of correlations between views and we conduct extensive experiments with existing methods.
% \item 
% We devise a novel noise regularization method for DCCA / DGCCA, leading to the proposal of NR-DCCA / NR-DGCCA. This approach introduces implicit constraints to the neural networks employed in DCCA / DGCCA. As a result, it effectively prevents the issue of model collapse and empowers the methods to handle data with varying degrees of correlation.
% \item 
% Empirically, to demonstrate the effectiveness of our method and the relevance of our theoretical results, we conduct extensive experiments on both synthetic and real-world datasets, including both natural language and vision, classification, and regression problems. We achieve consistently superior performances in all the experiments, validating the effectiveness of our method.
\end{itemize}

\section{Related Works}
\subsection{Multi-view representation learning}
MVRL aims to uncover relationships among multi-view data in an unsupervised manner, thereby obtaining semantically rich representations that can be utilized for various downstream tasks~\citep{sun2023learning,yan2021deep}.
Several works have been proposed to deal with MVRL from different aspects.
DMF-MVC~\citep{zhao2017multi} utilizes deep matrix factorization to extract a shared representation from multiple views. MDcR~\citep{zhang2016flexible} maps each view to a lower-dimensional space and applies kernel matching to enforce dependencies across the views. CPM-Nets~\citep{zhang2019cpm} formalizes the concept of partial MVRL and many works have been proposed for such issue~\citep{zhang2020deep,tao2019joint,li2022high,yin2021incomplete}. 
$\text{AE}^2$-Nets~\citep{zhang2019ae2} utilizes a two-level autoencoder framework to obtain a comprehensive representation of multi-view data.
DUA-Nets~\citep{geng2021uncertainty} takes a generative modeling perspective and dynamically estimates the weights for different views.
MVTCAE~\citep{hwang2021multi} explores MVRL from an information-theoretic perspective, which can capture the shared and view-specific factors of variation by maximizing or minimizing specific total correlation. Our work focuses on CCA as a simple, classic, and theoretically sound approach as it can still achieve state-of-the-art performance consistently.

% Our work focuses on the domain of CCA in complete MVRL.

\subsection{CCA and its variants}

% Canonical Correlation Analysis (CCA) and its variants are highly effective approaches for addressing such problems due to their ability to uncover relationships among multiple views~
Canonical Correlation Analysis (CCA)  projects the multi-view data into a unified space by maximizing their correlations~\citep{hotelling1992relations,horst1961generalized,hardoon2004canonical,lahat2015multimodal,yan2023robust,sun2023learning}.
It has been widely applied in various scenarios that involve multi-view data, including dimension reduction~\citep{zhang2016multi,sun2010scalable,avron2013efficient}, classification~\citep{kim2007tensor,sun2010canonical}, and clustering~\citep{fern2005correlation,chang2011libsvm}.
To further enhance the nonlinear transformability of CCA, Kernel CCA (KCCA) uses kernel methods, while Deep CCA (DCCA) employs DNNs. Since DNNs is parametric and can take advantage of large amounts of data for training, numerous DCCA-based methods have been proposed.~\cite{benton2017deep} utilizes DNNs to optimize the objective of Generalized CCA, to reveal connections between multiple views more effectively. To better preserve view-specific information,~\cite{wang2015deep} introduces the reconstruction errors of autoencoders to DCCA. Going a step further,~\cite{wang2016deep} proposes Variational CCA and utilizes dropout and private autoencoders to project common and view-specific information into two distinct spaces. 
Furthermore, many studies are exploring efficient methods for computing the correlations between multi-view data when dealing with more than two views such as MCCA, GCCA, and TCCA~\citep{horst1961generalized,nielsen2002multiset,kettenring1971canonical,hwang2021multi}. Some research focuses on improving the efficiency of computing CCA by avoiding the need for singular value decomposition (SVD)~\citep{chang2018scalable,chapman2022generalized}.
However, the model collapse issue of DCCA-based methods has not been explored and addressed.

\subsection{Noise Regularization}
% There has been significant interest in using random noise to regularize the networks during training 
Noise regularization is a pluggable approach to regularize the neural networks during training~\citep{bishop1995training,an1996effects,sietsma1991creating,gong2020maxup}. 
In supervised tasks,~\cite{sietsma1991creating} might be the first to propose that, by adding noise to the train data, the model will generalize well on new unseen data. Moreover,~\cite{bishop1995training,gong2020maxup} analyze the mechanism of the noise regularization, and~\cite{he2019parametric, gong2020maxup} indicate that noise regularization can also be used for adversarial training to improve the generalization of the network.
In unsupervised tasks,~\cite{poole2014analyzing} systematically explores the role of noise injection at different layers in autoencoders, and distinct positions of noise perform specific regularization tasks. 
% However, previous noise regularization methods are not suitable for DCCA / DGCCA since they have requirements for the networks, such as the need for supervised targets or reconstruction loss.
However, how to make use of noise regularization for DCCA-based methods, especially for preventing model collapse, has not been studied.

\section{Preliminaries}
% Recently, there has been a growing interest in learning from data with multiple views to produce useful representations, which we refer to as \textsc{MVRL}.
In this section, we will explain the objectives of the MVRL and then introduce Linear CCA and DCCA as representatives of the CCA-based methods and DCCA-based methods, respectively. Lastly, the model collapse issue in DCCA is demonstrated.

\subsection{Settings for MVRL}

Suppose the set of datasets from $K$ different sources that describe the same object is represented by $X$, and we define $X = \{X_{1},\cdots,X_{k},\cdots,X_{K}\}, X_{k} \in \mathbb{R}^{d_{k} \times n}$, where $x_k$ represents the $k$-th view ($k$-th data source), $n$ is the sample size, and $d_k$ represents the feature dimension for the $k$-th view. And we use $X_{k}'$ to denote the transpose of $X_{k}$. 
We take the Caltech101 dataset as an example and the training set has 6400 images. One image has been fed to three different feature extractors producing three features:  a 1984-d HOG feature, a 512-d GIST feature, and a 928-d SIFT feature. Then for this dataset, we have $X_1 \in \mathbb{R}^{1984 \times 6400}$, $X_2 \in \mathbb{R}^{512 \times 6400}$ , $X_3 \in \mathbb{R}^{928 \times 6400}$.
% Without loss of generality, we assume that all the datasets $X_k$ are zero-centered with respect to row \citep{hotelling1992relations}.
% Consider the following lemma:
% \begin{proposition}
%     \label{prob: zero-center}
%     Given a specific matrix $B$ and a zero-centered $C$ with respect to rows, the product $BC$ is also zero-centered with respect to rows.
% \end{proposition}
% This implies that given a specific matrix $Y$, the product $YX_k$ is also zero-centered with respect to rows. When computing the covariance matrix between such zero-centered matrices, there is no need for an additional subtraction of the mean of row, which simplifies our subsequent derivations.

% We use notations $X = \{X_{1},\cdots,X_{k},\cdots,X_{K}\}, X_{k} \in \mathbb{R}^{d_{k} \times n}$ to represent multi-view data with $K$ views. The $k$-th view data has sample size $n$ and feature dimension $d_{k}$.

% For the data of $k$-th view, let $\bar{X_{k}} \in \mathbb{R}^{d_{k} \times 1}$ represent the row mean of $X_{k}$, and let $\widehat{X_{k}}, \widehat{X_{k}} \in \mathbb{R}^{d_{k} \times n} $ denote $X_{k}-\bar{X_{k}}$, which is also called the zero-centered matrix of $X_k$. When we say that $X_k$ is zero-centered, it means that $\bar{X_k}$ is a zero matrix and $\widehat{X_{k}} = X_{k}$. And we use $X_{k}'$ to denote the transpose of $X_{k}$.

The objective of MVRL is to learn a transformation function $\Psi$ that projects the multi-view data $X$ to a unified representation $Z \in \mathbb{R}^{m \times n}$, where $m$ represents the dimension of the representation space, as shown below:
\begin{equation}
Z=\Psi(X) = \Psi(X_{1},\cdots,X_{k},\cdots,X_{K}).
\end{equation}
After applying $\Psi$ for representation learning, we expect that the performance of using $Z$ would be better than directly using $X$ for various downstream tasks.

% Using $Z$ instead of directly using $X$ can lead to better performance in various downstream tasks.
\subsection{Canonical Correlation Analysis}

Among various MVRL methods, CCA projects the multi-view data into a common space by maximizing their correlations. 
We first define the correlation between the two views as follows:
\begin{equation}
    \text{Corr}(W_1X_1, W_2X_2)  = \text{tr}((\Sigma_{11}^{-1/2}\Sigma_{12}\Sigma_{22}^{-1/2})'\Sigma_{11}^{-1/2}\Sigma_{12}\Sigma_{22}^{-1/2})^{1/2}
\end{equation}
% \begin{equation}

% % \begin{aligned}
%  \text{Corr}(W_1X_1, W_2X_2)  = \text{tr}((\Sigma_{11}^{-1/2}\Sigma_{12}\Sigma_{22}^{-1/2})'\Sigma_{11}^{-1/2}\Sigma_{12}\Sigma_{22}^{-1/2})^{1/2}
% % %  &=\frac{1}{(n-1)^2} \text{tr}((W_2X_2(W_2X_2)')^{-1}(
% % % W_1X_1(W_2X_2)')'(W_1X_1(W_1X_1)')^{-1}(W_1X_1(W_2X_2)'))^{1/2},\\
% % % &=\frac{1}{(n-1)^2} \text{tr}((W_2X_2(W_2X_2)')^{-1}(
% % % W_2X_2(W_1X_1)')(W_1X_1(W_1X_1)')^{-1}(W_1X_1(W_2X_2)'))^{1/2},
% % \end{aligned}

% \end{equation}
where $\text{tr}$ denotes the matrix trace, $\Sigma_{11}$, 
$\Sigma_{22}$ represent the self-covariance matrices of the projected views, and $\Sigma_{12}$ is the cross-covariance matrix between the projected views~\citep{d1994use,andrew2013deep}. The correlation between the two projected views can be regarded as the sum of all singular values of the normalized cross-covariance~\citep{hotelling1992relations,anderson1958introduction}.

% For the $2$-view data,  \textsc{CCA} searches for the linear projection matrices $W_1, W_2$ that maximize correlation using the following steps:
% firstly, \textsc{CCA} calculates the covariance matrices of the projected data, $W_1X_1$ and $W_2X_2$, with themselves as follows:
% $\Sigma_{11} = \frac{1}{n-1} \cdot (W_1X_1)(W_1X_1)'$, 
% $\Sigma_{22} = \frac{1}{n-1} \cdot (W_2X_2)(W_2X_2)'$;
% then the covariance matrix between $W_1X_1$ and $W_2X_2$ is calculated as follows:
% $\Sigma_{12} = \frac{1}{n-1} \cdot (W_1X_1)(W_2X_2)'$; we further maximize the trace norm of $\Sigma_{11}^{-1/2}\Sigma_{12}\Sigma_{22}^{-1/2}$ \citep{martin1979multivariate,andrew2013deep}:
% \begin{equation}
% \begin{aligned}
% (W_1^*, W_2^*) 
% & =\arg\max_{W_1,W_2} \text{Corr}(W_1X_1, W_2X_2), \\
% & = \arg \max_{W_1,W_2} \text{tr}((\Sigma_{11}^{-1/2}\Sigma_{12}\Sigma_{22}^{-1/2})'\Sigma_{11}^{-1/2}\Sigma_{12}\Sigma_{22}^{-1/2})^{1/2}. \\
% \end{aligned}
% \end{equation}
% where $\text{Corr}$ represents the correlation between the two projected representations, and $\text{tr}$ denotes the matrix trace. 

% \begin{equation}
% \resizebox{.95\linewidth}{!}{
% \begin{aligned}
% (W_1^*, W_2^*) 
% & = \arg \max_{W_1,W_2} \text{tr}(\Sigma_{22}^{-1}\Sigma_{12}'\Sigma_{11}^{-1}\Sigma_{12})^{1/2} \\ 
% & = \arg \max_{W_1,W_2} \frac{1}{(n-1)^2} \text{tr}((W_2X_2(W_2X_2)')^{-1}(
% W_1X_1(W_2X_2)')'(W_1X_1(W_1X_1)')^{-1}(W_1X_1(W_2X_2)'))^{1/2}, \\
% & = \arg\max_{W_1,W_2} \frac{1}{(n-1)^2} \text{tr}((W_2X_2(W_2X_2)')^{-1}(
% W_2X_2(W_1X_1)')(W_1X_1(W_1X_1)')^{-1}(W_1X_1(W_2X_2)'))^{1/2}.
% \end{aligned}
% \end{equation}

For multiple views, their correlation is defined as the summation of all the pairwise correlations~\citep{nielsen2002multiset,kettenring1971canonical}, which is shown as follows:
\begin{equation}
\text{Corr}(W_1X_1,\cdots, W_kX_k, \cdots, W_KX_K) = \sum_{k < j} \text{Corr}(W_kX_k, W_jX_j).
\end{equation}

Essentially, Linear CCA searches for the linear transformation matrices $\{W_k\}_k$ that maximize correlation among all the views. Mathematically, it can be represented as follows~\citep{wang2015deep}:
\begin{equation}
\{W_k^*\}_k = \arg \max_{ \{W_k\}_k } \text{Corr}(W_1X_1,\cdots, W_kX_k, \cdots, W_KX_K).
\end{equation}
% For multiple views, the unified representation is obtained by maximizing the summation of the pairwise CCA of all the two views \citep{TODO}. 
Once $W_k^*$ is obtained by backpropagation, the multi-view data are projected into a unified space. Lastly, all projected data are concatenated to obtain $Z=[W_1^* X_1; \cdots; W_k^*X_k; \cdots; W_K^* X_K]$  for downstream tasks.

As an extension of linear CCA, DCCA employs neural networks to capture the nonlinear relationship among multi-view data.
The only difference between DCCA and Linear CCA is that the linear transformation matrix $W_k$ is replaced by multi-layer perceptrons (MLP). Specifically, each $W_k$ is replaced by a neural network $f_k$, which can be viewed as a nonlinear transformation. 
Similar to Linear CCA, the goal of DCCA is to solve the following optimization problem:
\begin{equation}
\{f_k^*\}_k = \arg \max_{ \{f_k\}_k } \text{Corr}\left(f_1(X_1),\cdots, f_k(X_k), \cdots, f_K(X_K)\right).
\end{equation}
The parameters in Linear CCA and DCCA are both updated through backpropagation~\citep{andrew2013deep,wang2015deep}. 
Again, the unified representation is obtained by $Z=[f_1^* (X_1); \cdots; f_k^*(X_k); \cdots; f_K^* (X_K)]$  for downstream tasks.

% \begin{figure}[h]
% \begin{center}
% \includegraphics[width=0.6\linewidth]{figures/syn_mean_std_select.png}
% \end{center}
% \caption{Mean and standard deviation of the testing accuracy for different MVRL methods along the training process.}
% \label{fig: mean and std select}
% \end{figure}

\section{Model Collapse of DCCA}

Despite exhibiting promising performance, DCCA shows a significant decline in performance as the training proceeds.  We define this decline-in-performance phenomenon as the model collapse of DCCA. 

Previous studies found that the representations (i.e., final output) of both Linear CCA and DCCA are full-rank \citep{andrew2013deep,de2003regularization}. However, we further demonstrate that both representations and weight matrices of Linear CCA are full-rank whereas DCCA only guarantees that representations are full-rank but not for the weight matrices. 
Given that Linear CCA has only a single layer of linear transformation $W_k$ and the representations $W_kX_k$ are constrained to be full-rank by the loss function,  $W_k$ in Linear CCA is full-rank (referred to Lemma~\ref{lemma: less-rank} and assume that $W_k$ is a square matrix and $X_k$ is full-rank).
As for DCCA, we consider a simple case when $f_k(X_k) = Relu(W_kX_k)$, and $f_k$ is a single-layer network and uses an element-wise Relu activation function. 
Only the representations $Relu(W_kX_k)$ are constrained to be full-rank, and hence we cannot guarantee that $W_kX_k$ is full-rank. 
For example, when $Relu(W_kX_k) = \left(\begin{smallmatrix} 1, & 0 \\ 0, & 1 \end{smallmatrix}\right)$, it is clear that this is a matrix of rank 2, but in fact $W_kX_k$ can be $\left(\begin{smallmatrix} 1, & -1 \\ -1, & 1 \end{smallmatrix}\right)$, and this is not full-rank.
This reveals that the neural network $f_k$ is overfitted on $X_k$, i.e., making representations $Relu(W_kX_k)$ to be full-rank with the constraint of its loss function, rather than $W_k$ itself being full-rank (verified in Appendix \ref{appendix:Hyper-paramter of ridge regularization}).

\begin{figure*}[h]
	\centering
	\subfigure[100-th epoch (DCCA)]{
		\begin{minipage}[t]{0.5\linewidth}
			\centering
			\includegraphics[width=\linewidth]{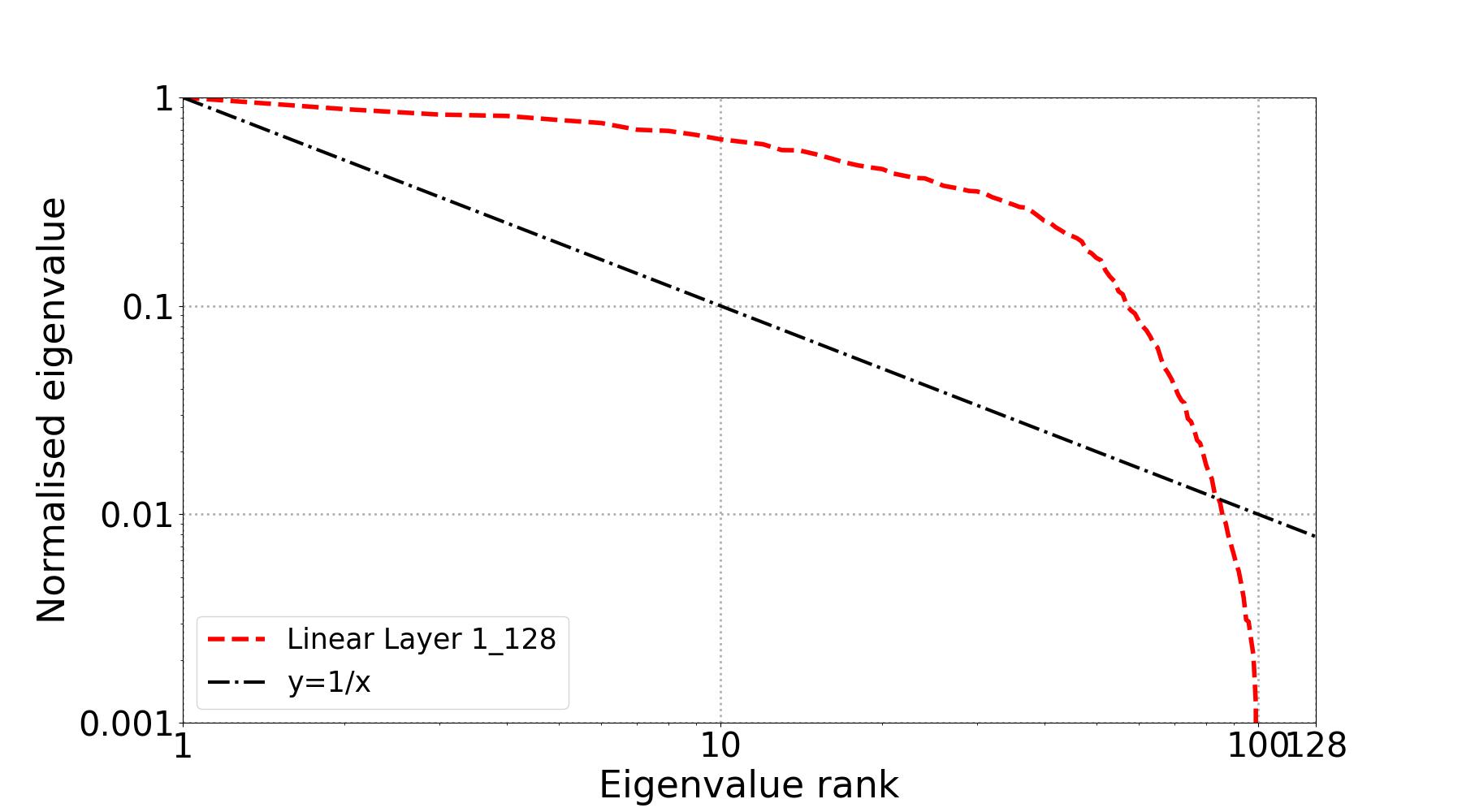}
		\end{minipage}
	}%
        \subfigure[100-th epoch (NR-DCCA)]{
		\begin{minipage}[t]{0.5\linewidth}
			\centering
			\includegraphics[width=\linewidth]{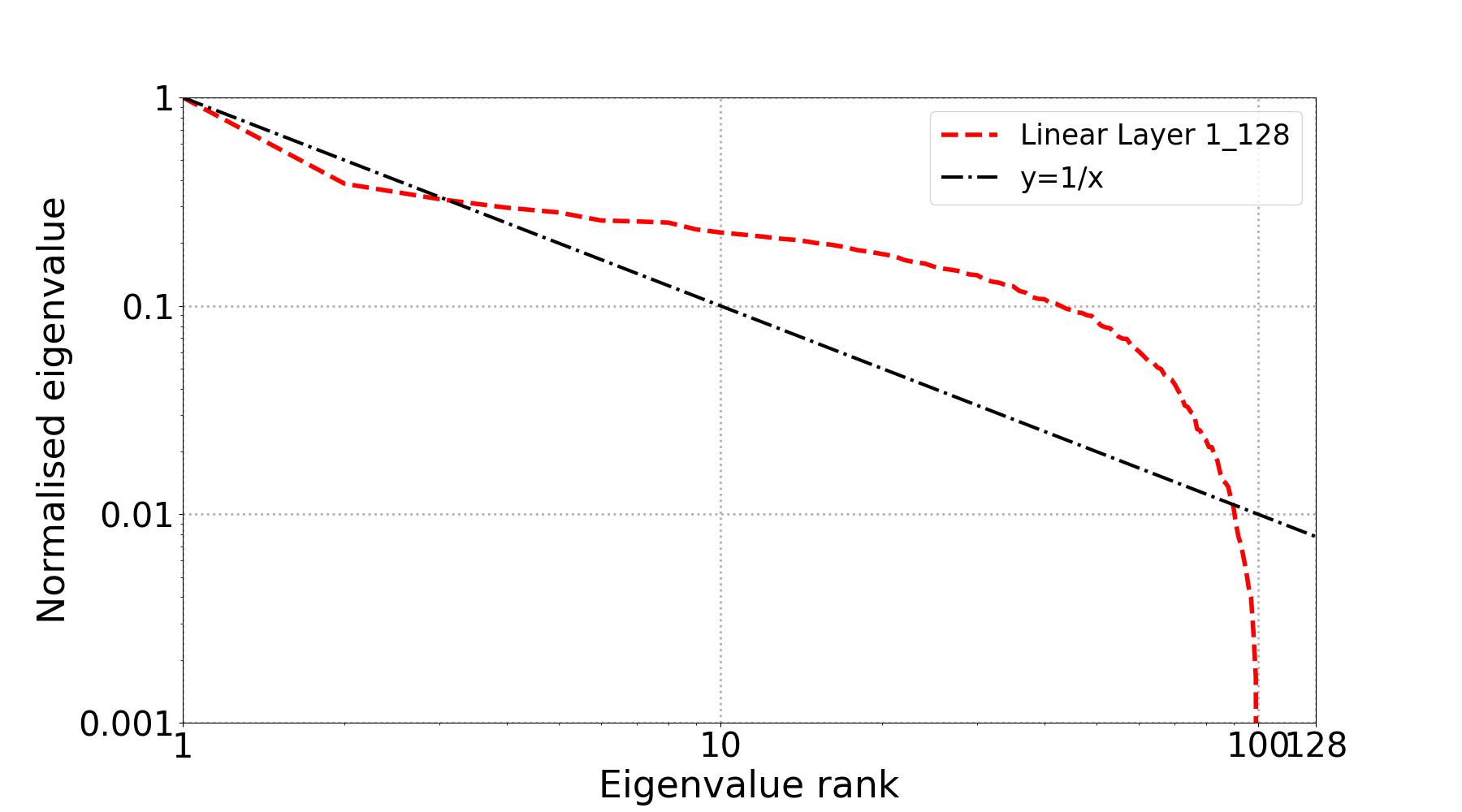}
		\end{minipage}
	}%

        \subfigure[1200-th epoch (DCCA)]{
		\begin{minipage}[t]{0.5\linewidth}
			\centering
			\includegraphics[width=\linewidth]{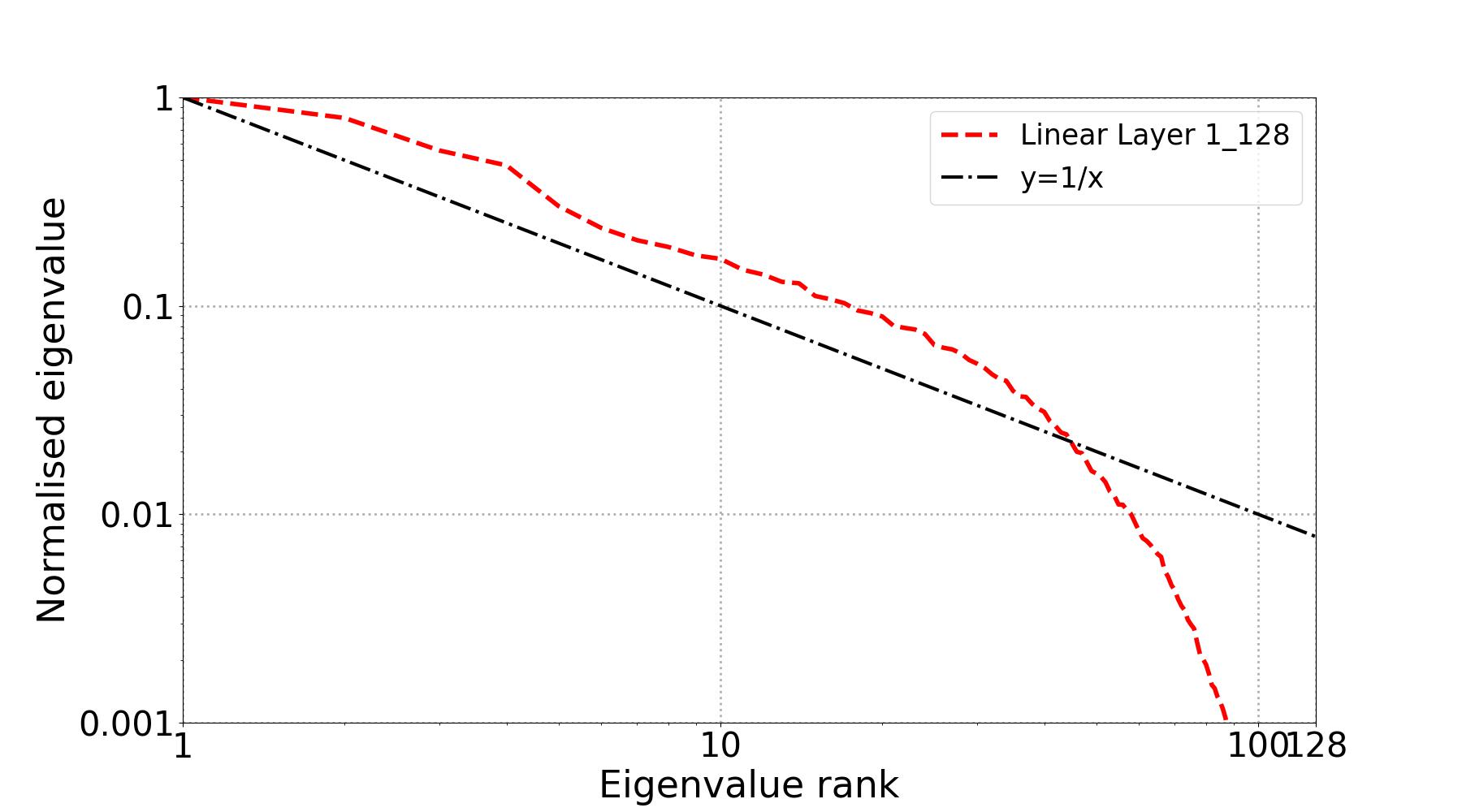}
		\end{minipage}
	}%
        \subfigure[1200-th epoch (NR-DCCA)]{
		\begin{minipage}[t]{0.5\linewidth}
			\centering
			\includegraphics[width=\linewidth]{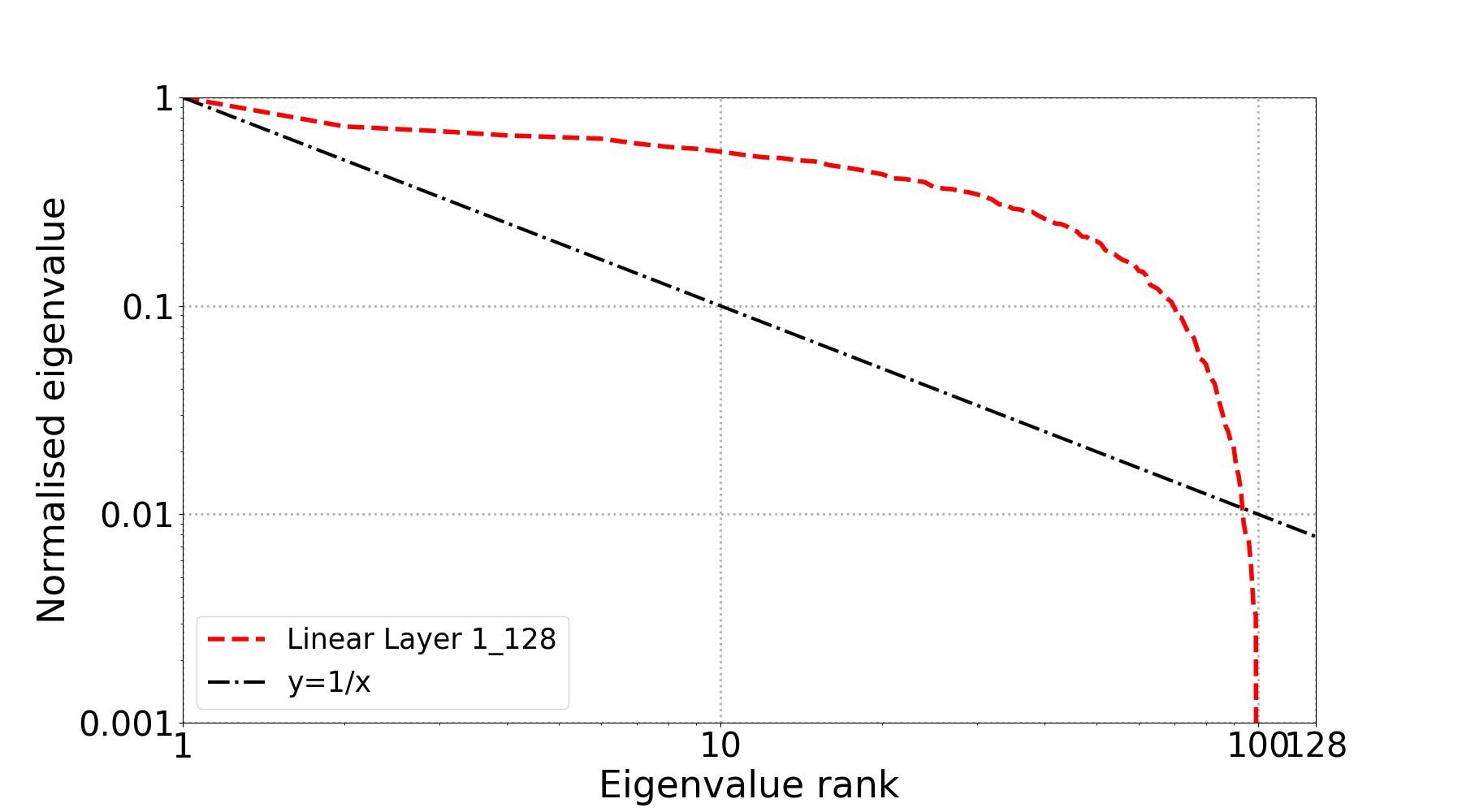}
		\end{minipage}
	}%

	\centering
	\caption{Eigenvalue distributions of the first linear layer's weight matrices in the encoder of $1$-st view.}
 \label{fig: eignvalue_in_training}
\end{figure*}

Thus, we hypothesize that model collapse in DCCA arises primarily due to the low-rank nature of the DNN weight matrices. To investigate this, we analyze the eigenvalue distributions of the first linear layer's weight matrices in both DCCA and NR-DCCA across various training epochs on synthetic datasets. Figure \ref{fig: eignvalue_in_training} illustrates that during the initial training phase (100th epoch), the eigenvalues decay slowly for both DCCA and NR-DCCA. However, by the 1200th epoch, DCCA exhibits a markedly faster decay in eigenvalues compared to NR-DCCA. This observation suggests a synchronization between model collapse in DCCA and increased redundancy of the weight matrices. For more details on the experimental setup and results, please refer to Section~\ref{sec: synthetic performance}.

\section{DCCA with Noise Regularization (NR-DCCA)}

\subsection{Method}
\begin{figure}[h]
    \centering
    \includegraphics[width=0.7\linewidth]{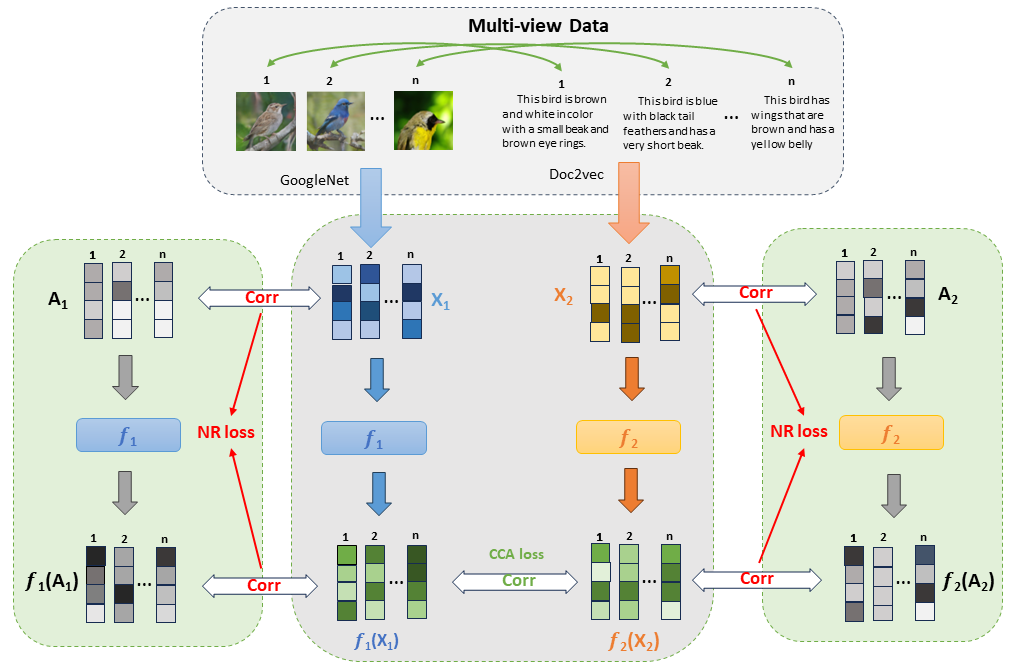}
    \caption
    {Illustration of NR-DCCA. We take the CUB dataset as an example: similar to DCCA, the $k$-th view  $X_k$ is transformed using  $f_k$  to obtain new representation $f_k(X_k)$ and then maximize the correlation between new representations. Additionally, for the $k$-th view, we incorporate the proposed NR loss to regularize $f_k$.}
    \label{fig:framework}
\end{figure}

Based on the discussions in previous sections, we present NR-DCCA, which makes use of the noise regularization approach to prevent model collapse in DCCA. Indeed, the developed noise regularization approach can be applied to variants of DCCA methods, such as Deep Generalized CCA (DGCCA)~\citep{benton2017deep}.
An overview of the NR-DCCA framework is presented in Figure~\ref{fig:framework}.

% : a deep \textsc{MVRL} method based on \textsc{CCA}, where we combine \textsc{DCCA} with noise regularization as illustrated in Fig. \ref{fig:framework}. 

% It is worth noting that we also introduce noise regularization to Deep Genernalize CCA(\textsc{DGCCA}), which we refer to as \textsc{NR-DGCCA}.

The key idea in NR-DCCA is to generate a set of i.i.d Gaussian white noise, denoted as $A = \{A_{1},\cdots, A_{k},\cdots, A_{K}\}, A_{k} \in \mathbb{R}^{d_{k} \times n}$, with the same shape as the multi-view data $X_k$.
% In CCA, $W_kA_k$ is still Gaussian white noise, and hence $\text{Corr}(X_k, A_k) = \text{Corr}(W_kX_k,W_kA_k)$ (rigorous proof provided in Proposition~\ref{prop:ccaequal}). 
In Linear CCA, the correlation with noise is invariant to the linear transformation $W_k$: $\text{Corr}(X_k, A_k) = \text{Corr}(W_kX_k,W_kA_k)$ (rigorous proof provided in Theorem \ref{thm:cca}). 
However, for DCCA, $\text{Corr}(X_k, A_k)$ might not equal $\text{Corr}(f_k(X_k),f_k(A_k))$ because the powerful neural networks $f_k$ have overfitted to the maximization program in DCCA and the weight matrices have been highly self-related. 
Therefore, we enforce the DCCA to mimic the behavior of Linear CCA by adding an NR loss $\zeta_k = \left| Corr(f_{k}(X_{k}),f_{k}(A_{k})) - Corr(X_{k},A_{k}) \right|$, and hence the formulation of NR-DCCA is:
% $$\mathop{max}_{\{f_{k}\}_{k=1}^{K}} Corr(f_{1}(X_{1}),\cdots,f_{k}(X_{k}),\cdots,f_{K}(X_{K})) - \alpha \sum_{k=1}^{K}  \left| Corr(f_{k}(X_{k}),f_{k}(A_{k})) - Corr(X_{k},A_{k}) \right| $$
\begin{equation}
\{f_k^*\}_k = \arg \max_{ \{f_k\}_k } \text{Corr}\left(f_1(X_1),\cdots, f_K(X_K)\right) - \alpha \sum_{k=1}^{K}  \zeta_k.
\end{equation}
where $\alpha$ is the hyper-parameter weighing the NR loss. NR-DCCA can be trained through backpropagation with the randomly generated $A$ in each epoch, and the unified representation is obtained directly using $\{f_k^*\}_k$ in the same manner as DCCA.

\subsection{Theoretical Analysis}

% In this section, we demonstrate that both representations and weight matrices of Linear CCA are full-rank, whereas DCCA only guarantees that representations are full-rank but not for the weight matrices. 
% Considering that Linear CCA does not show the model collapse while DCCA does, we conjecture that the root cause of the model collapse in DCCA is that the weight matrices in DNNs tend to be low-rank.
In this section, we provide the rationale for why the developed noise regularization can help to prevent the weight matrices from being low-rank and thus model collapse. Moreover, we prove the effect of full-rank weight matrices on the representations, which provides a tool to empirically verify the full-rank property of weight matrices by the quality of representations.

% Firstly, considering Linear CCA has only a single layer of linear transformation $W_k$ and the representations $W_kX_k$ are constrained to be full-rank by the loss function,  $W_k$ in Linear CCA is full-rank (referred to Lemma~\ref{lemma: less-rank} and assume that $W_k$ is a square matrix and $X_k$ is full-rank), and thus there is no model collapse in Linear CCA. 

% As for DCCA, we consider a simple case when $f_k(X_k) = Relu(W_kX_k)$, and $f_k$ is a single-layer network and uses an element-wise Relu activation function. 
% Only the representations $Relu(W_kX_k)$ is constrained to be full-rank, and hence we cannot guarantee that $W_kX_k$ is full-rank. 
% For example, when $Relu(W_kX_k) = \left(\begin{smallmatrix} 1, & 0 \\ 0, & 1 \end{smallmatrix}\right)$, it is clear that this is a matrix of rank 2, but in fact $W_kX_k$ can be $\left(\begin{smallmatrix} 1, & -1 \\ -1, & 1 \end{smallmatrix}\right)$, and this is not full-rank.
% This reveals that the neural network $f_k$ is overfitted on $X_k$, i.e., making representations $Relu(W_kX_k)$ to be full-rank with the constraint of its loss function, rather than $W_k$ itself being full-rank.

Utilizing a new Moore-Penrose Inverse (MPI)-based \citep{petersen2008matrix} form of $Corr$ in CCA, we discover that the full-rank property of $W_k$ is equal to CIP:
\begin{theorem}[Correlation Invariant Property (CIP) of $W_k$]
\label{thm:cca}
Given $W_k$ is a square matrix for any $k$ and $\eta_k = \left| Corr(W_{k}X_{k},W_{k}A_{k}) - Corr(X_{k},A_{k}) \right|$, we have $\eta_k = 0$ (i.e. CIP) $\iff W_k$ is full-rank. 
\end{theorem} 
Similarly, we say $f_k$ possess CIP if $\zeta_k = 0$. Under Linear CCA, it is redundant to introduce the NR approach and force $W_k$ to possess CIP, since forcing $W_kX_k$ to be full-rank is sufficient to ensure that $W_k$ is full-rank.  
However, in DCCA,  $f_k$ is overfitted on $X_k$, i.e., making representations $f_k(X_k)$ to be full-rank, rather than weight matrices in $f_k$ being full-rank. 
By forcing $f_k$ to possess CIP and thus mimicking the behavior of Linear CCA, the NR approach constrains the weight matrices to be full-rank and less redundant and thus prevents model collapse.

Next, we show that full-rank weight matrices (i.e., CIP) can greatly affect the quality of representations.
\begin{theorem}[Effects of CIP on the obtained representations]
\label{thm: CIP on Obtained Representations}
For any $k$, if  $W_k$ is a square matrix and CIP holds for $W_k$ (i.e. $W_k$ is full-rank), $W_kX_k$ holds that: 
\begin{equation}
    \min_{P_k} \|P_k W_kX_k - X_k\|_F = 0
\end{equation}
\begin{equation}
    \min_{Q_k} \|Q_k W_k(X_k+A_k) - W_kX_k\|_F \leq \sqrt{n}\|W_kA_k\|_F, E(\|W_KA_k\|_F^2) = \|W_k\|_F^2
\end{equation}
where $\|\cdot\|_F$  denotes the Frobenius norm and $\epsilon$ is a small positive threshold. $P_k$ and $Q_k$ are searched weight matrices of $k$-th view to recover the input and discard noise, respectively. 
And we refer $\|P_k W_kX_k - X_k\|_F$ and $\|Q_k W_k(X_k+A_k) - W_kX_k\|_F$ as reconstruction loss and denosing loss.
\end{theorem}

Theorem~\ref{thm: CIP on Obtained Representations} suggests that the obtained representation is of low reconstruction loss and denoising loss.
Low reconstruction loss suggests that the representations can be linearly reconstructed to the inputs. This implies that $W_k$ preserves distinct and essential features of the input data, which is a desirable property to avoid model collapse since it ensures that the model captures and retains the whole modality of data~\citep{zhang2019ae2,tschannen2018recent,tian2022comprehensive}.
Low denoising loss implies that the model's representation is robust to noise, which means that small perturbations in the input do not lead to significant changes in the output. This condition can be seen as a form of regularization that prevents overfitting the noise in the data~\citep{zhou2017anomaly,yan2023robust,staerman2023functional}. 
Additionally, the theorem also suggests that the rank of weight matrices is a good indicator to assess the quality of representations, which coincides with existing literature \citep{kornblith2019similarity,raghu2021vision,garrido2023rankme,nguyen2020wide,agrawal2022alpha}.

\section{Numerical Experiments}
\label{sec:evaluation}
We conduct extensive experiments on both synthetic and real-world datasets to answer the following research questions:
\begin{itemize}
    \item \textbf{RQ1:} How can we construct synthetic datasets to evaluate the MVRL methods comprehensively? 
    \item \textbf{RQ2:} Does NR-DCCA avoid model collapse across all synthetic MVRL datasets?
    \item \textbf{RQ3:} Does NR-DCCA perform consistently in real-world datasets?
    % \item \textbf{RQ4:} How do different parameter settings affect our model performance?
    % \item \textbf{RQ5:} How do different parameter settings affect our model performance?
\end{itemize}

We follow the protocol described in \cite{hwang2021multi} for evaluating the MVRL methods. For each dataset, we construct a training dataset and a test dataset. The encoders of all MVRL methods are trained on the training dataset. Subsequently, we encode the test dataset to obtain the representation, which will be evaluated in downstream tasks. 
We employ Ridge Regression~\citep{hoerl1970ridge} for the regression task and use $R2$ as the evaluation metric. For the classification task, we use a Support Vector Classifier (SVC)~\citep{chang2011libsvm} and report the average F1 scores. All tasks are evaluated using 5-fold cross-validation, and the reported results correspond to the average values of the respective metrics.

For a fair comparison, we use the same architectures of MLPs for all D(G)CCA methods. To be specific, for the synthetic dataset, which is simple, we employ only one hidden layer with a dimension of 256. For the real-world dataset, we use MLPs with three hidden layers, and the dimension of the middle hidden layer is 1024. We further demonstrate that increasing the depth of MLPs further accelerates the mod collapse of DCCA, while NR-DCCA maintains a stable performance in Appendix ~\ref{appendix:complexity of encoders}. 

Baseline methods include \textbf{CONCAT}, \textbf{PRCCA}~\citep{tuzhilina2023canonical}, \textbf{KCCA}~\citep{akaho2006kernel}, \textbf{Linear CCA}~\cite{wang2015deep},\textbf{Linear GCCA},\textbf{DCCA}~\citep{andrew2013deep},\textbf{DCCA\_EY}, \textbf{DCCA\_GHA} ~\citep{chapman2022generalized}, \textbf{DGCCA}~\citep{benton2017deep}, \textbf{DCCAE/DGCCAE}~\citep{wang2015deep}, \textbf{DCCA\_PRIVATE/DGCCA\_PRIVATE}~\citep{wang2016deep}, and \textbf{MVTCAE}~\citep{hwang2021multi}.
% Details of the experiment setting including hyper-parameters are presented in Appendix~\ref{appendix: Datasets and Baselines}~\ref{appendix:Hyper-paramter Settings}.

It is important to note that our proposed NR approach requires the noise matrix employed to be full-rank, which is compatible with several common continuous noise distributions. In our primary experiments, we utilize Gaussian white noise. Additionally, as demonstrated in Appendix ~\ref{appenidx:Effects of the Distribution of Noise}, uniformly distributed noise is also effective in our NR approach.

Details of the experiment settings including datasets and baselines are presented in Appendix ~\ref{appendix: Datasets and Baselines}. Hyper-parameter settings, including ridge regularization of DCCA, $\alpha$ of NR, are discussed in Appendix~\ref{appendix:Hyper-paramter Settings}.
We also analyze the computational complexity of different DCCA-based methods in Appendix~\ref{appendix:Complexity Analysis} and the learned representations are visualized in Appendix~\ref{appendix:visulization}. 
In the main paper, we mainly compare Linear CCA, DCCA-based methods, and NR-DCCA while other MVRL methods are discussed in Appendix \ref{appendix:additional results}. The results related to DGCCA and are similar and presented in Appendix~\ref{appendix:dgcca}.

\begin{figure}[h]
    \centering
    \includegraphics[width=0.8\linewidth]{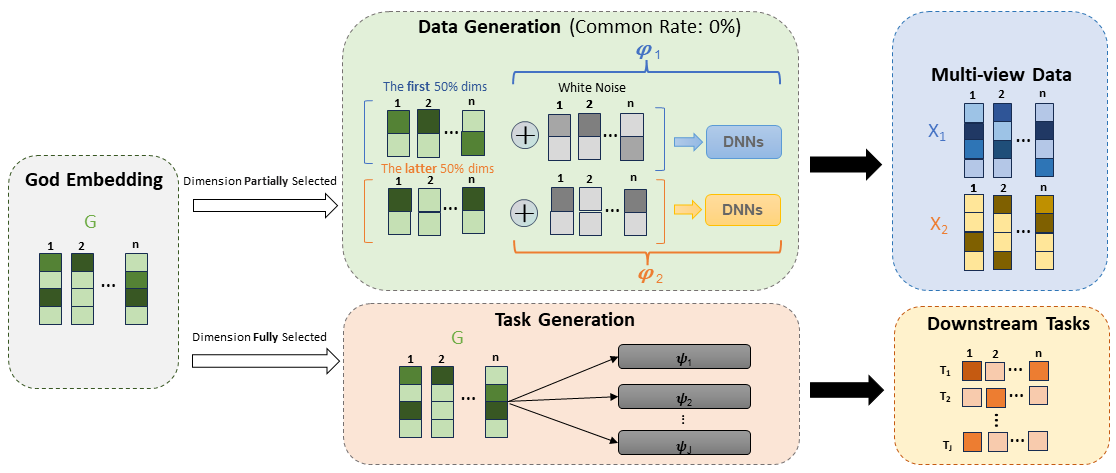}
    \caption{Construction of a synthetic dataset.
     This example consists of $2$ views and $n$ objects, and the common rate is $0\%$.}
    \label{fig: Construction of synthetic benchmarks}
\end{figure}

\subsection{Construction of synthetic datasets (RQ1)}
We construct synthetic datasets to assess the performance of MVRL methods, and the framework is illustrated in Figure~\ref{fig: Construction of synthetic benchmarks}. 
% Representation learning holds that data samples can be represented as high-dimensional embeddings, with all of the sample's feature information stored within the embeddings. 
We believe that the multi-view data describes the same object, which is represented by a high-dimensional embedding $G^{d \times n}$, where $d$ is the feature dimension and $n$ is the size of the data, and we call it God Embedding. 
Each view of data is regarded as a non-linear transformation of part (or all) of $G$. For example, we choose $K=2, d= 100$, and then $X_1 = \phi_1(G[0:50+\text{CR}/2,:]), X_2 = \phi_x(G[50-\text{CR}/2:100],:)$, where 
$\phi_1$ and $\phi_2$ are non-linear transformations, and $CR$ is referred to as common rate. The common rate is defined as follows: 
\begin{definition}[Common Rate]
\label{def:common rate}
For two view data $X = \{X_{1}, X_{2}\}$, common rate is defined as the percentage overlap of the features in $X_1$ and $X_2$ that originate from $G$.
\end{definition}
One can see that the common rate ranges from $0\%$ to $100\%$. The larger the value, the greater the correlation between the two views, and a value of $0$ indicates that the two views do not share any common dimensions in $G$.
Additionally, we construct the downstream tasks by directly transforming the God Embedding $G$. Each task $T_j = \psi_j(G)$, where $\psi_j$ is a transformation, and $T_j$ represents the $j$-th task. 
% We refer to this high-dimensional embedding as the God Embedding $G \in \mathbb{R}^{d \times n}$, where $d$ is the feature dimension and $n$ is the size of the data. 
% For the $k$-th view, its unique information extraction can be understood as a complex and independent nonlinear transformation $O_{k}$, including the selection of feature dimensions, noise addition, linear projection, and nonlinear activation. As a result, by setting transformations set $O = \{O_{1},\cdots,O_{k},\cdots,O_{K}\}$, we can obtain a multi-view data $X$ with $K$ views originated from $G$.
% In practical scenarios, this can be exemplified in the process of collecting data from each view, where the data from each view may only capture partial information about the object. Moreover, the collection process may be hindered by various forms of interference, and the data may undergo complex feature extraction (e.g., through DNNs). 
% In addition, we also need downstream tasks to evaluate the effectiveness of the representations learned from multi-view data. For task $T_j$, we consider it to be obtained by projecting the God Embedding through a task-specific projection matrix $M_j$. Let us denote the set of tasks and their corresponding task-specific projection matrices as $T = \{T_{1},\cdots, T_{j},\cdots, T_{J}\}$ and $M \{M_{1},\cdots, M_{j},\cdots, M_{J}\}$, respectively.
By setting different $G$, common rates, $\phi_k$, and $\psi_j$, we can create various synthetic datasets to evaluate the MVRL methods.
Finally, $X_k$ are observable to the MVRL methods for learning the representation, and the learned representation will be used to classify/regress $T_j$ to examine the performance of each method.
Detailed implementation is given in Appendix~\ref{appendix:implementation of Synthetic datasets}.

\begin{figure*}[h]
	\centering
	\subfigure[Performance]{
		\begin{minipage}[t]{0.8\linewidth}
			\centering
			\includegraphics[width=\linewidth]{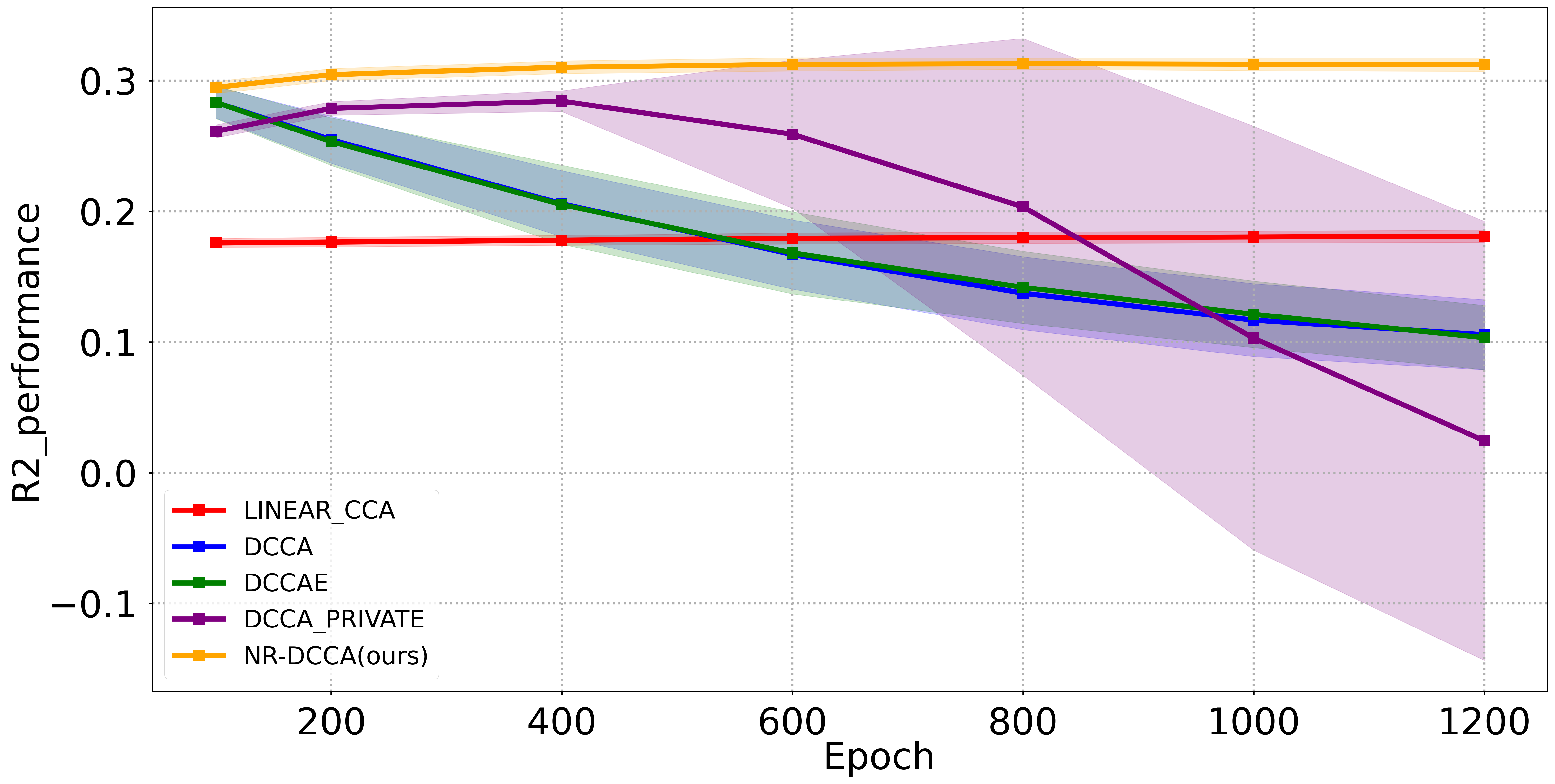}
		\end{minipage}
	}%

        \subfigure[Correlation]{
		\begin{minipage}[t]{0.5\linewidth}
			\centering
			\includegraphics[width=\linewidth]{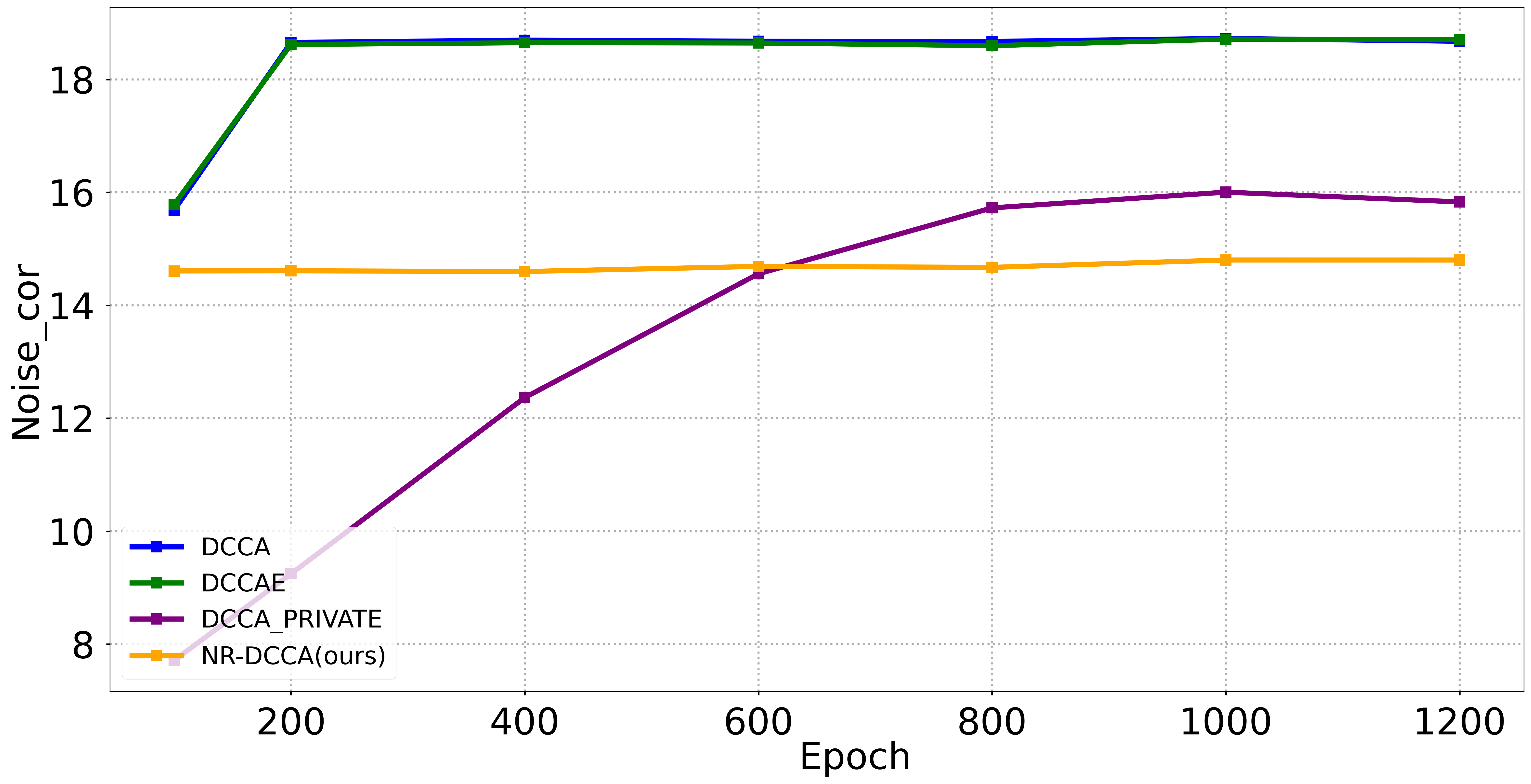}
		\end{minipage}
	}%
        \subfigure[NESum of weight matrices]{
		\begin{minipage}[t]{0.5\linewidth}
			\centering
			\includegraphics[width=\linewidth]{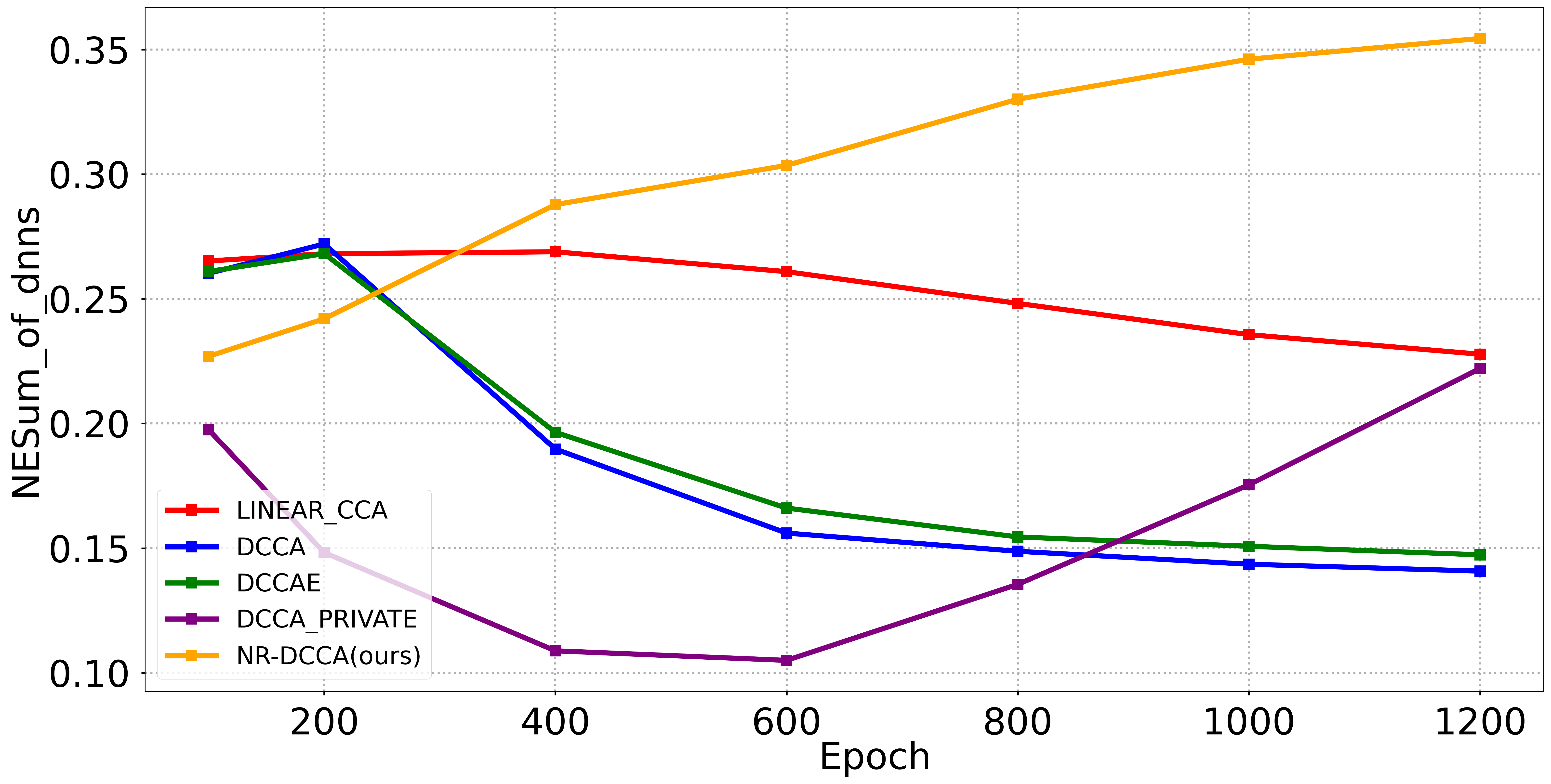}
		\end{minipage}
	}%

 \subfigure[Reconstruction]{
		\begin{minipage}[t]{0.5\linewidth}
			\centering
			\includegraphics[width=\linewidth]{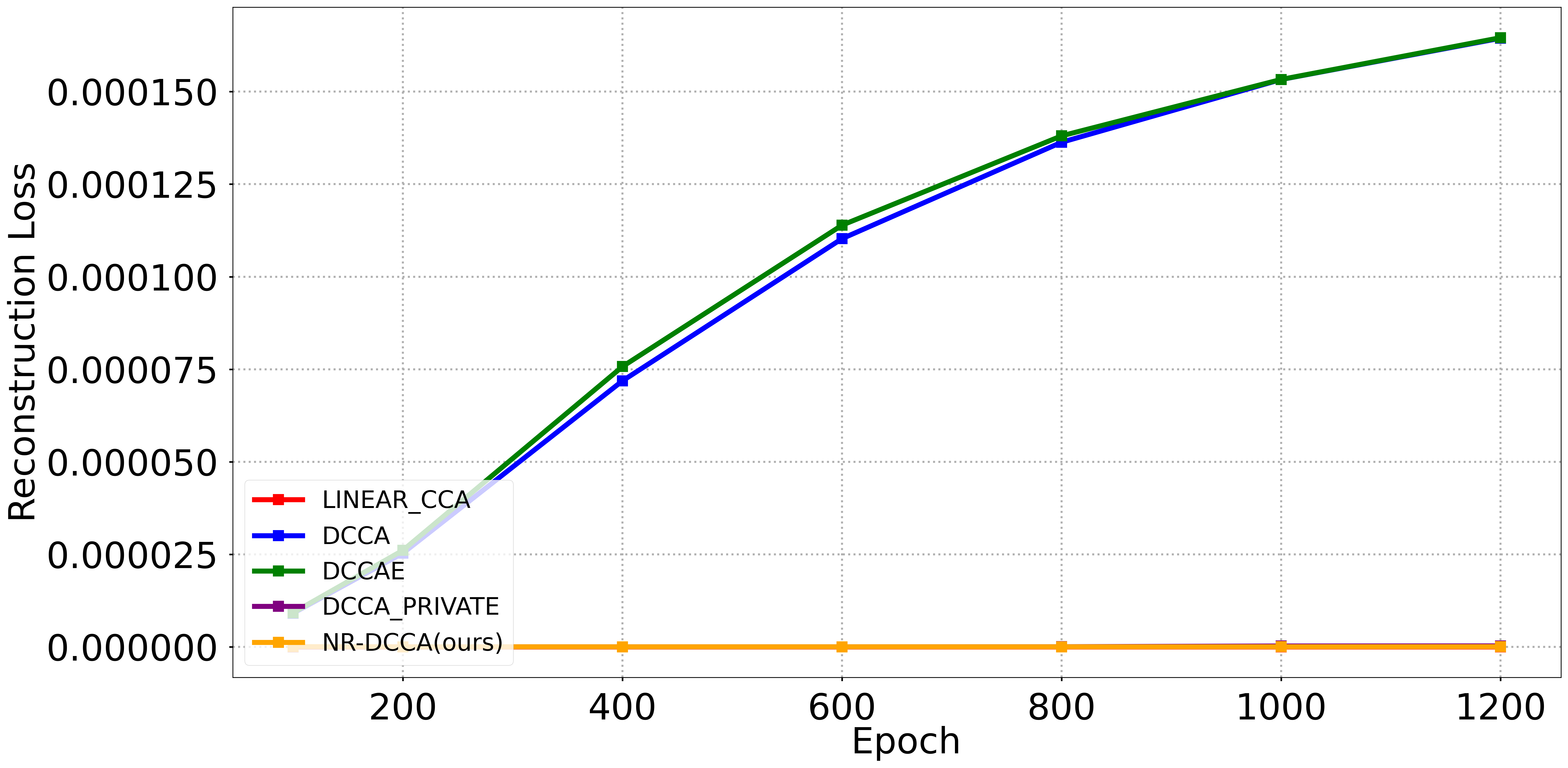}
		\end{minipage}
	}%
    \subfigure[Denoisng]{
		\begin{minipage}[t]{0.5\linewidth}
			\centering
			\includegraphics[width=\linewidth]{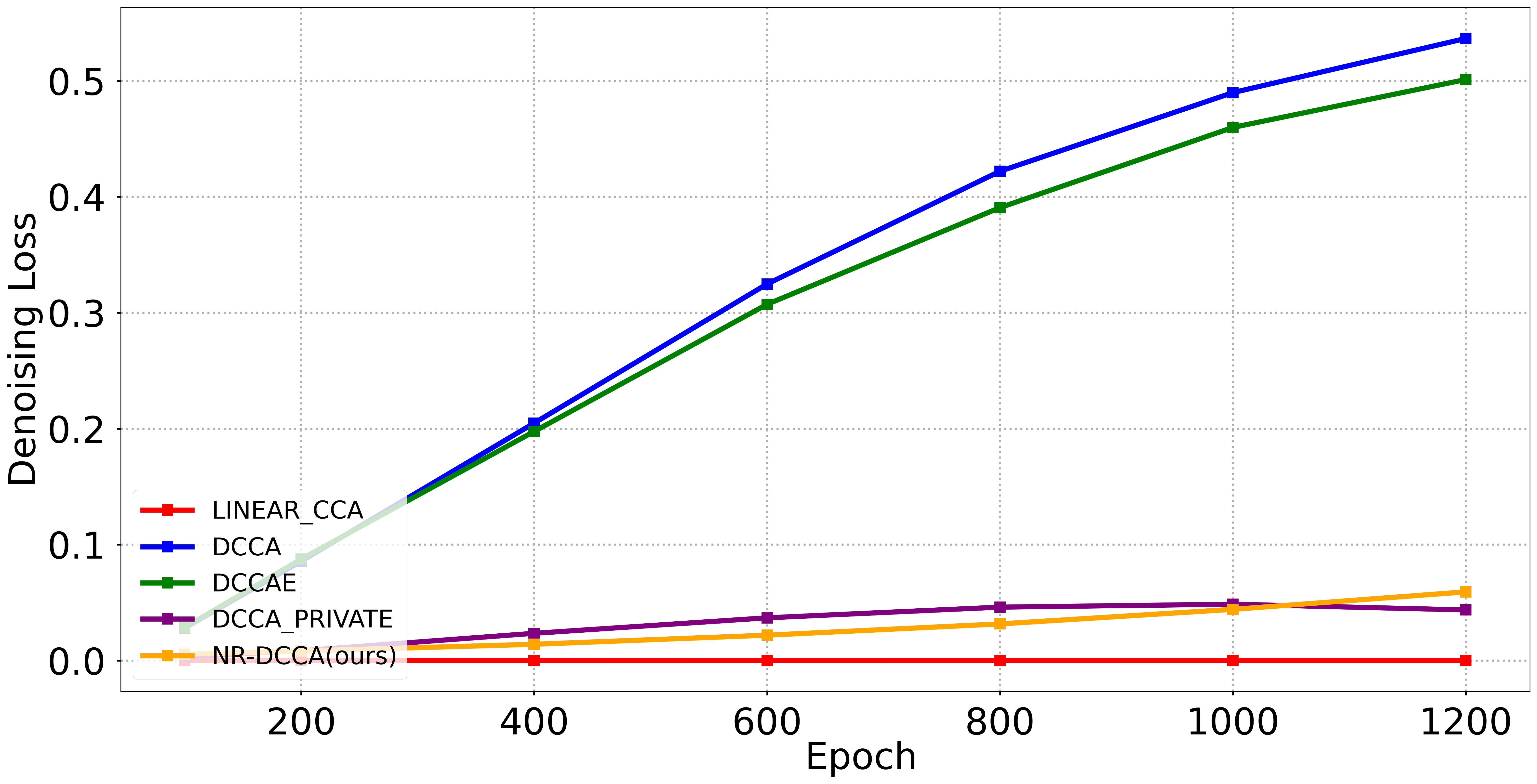}
		\end{minipage}
	}%
	\centering
	\caption{(a) Mean and standard deviation of the (D)CCA-based method performance across synthetic datasets in different training epochs. (b) The mean correlation between noise and real data after transformation varies with epochs. (c) Average NESum across all weights within the trained encoders.
   (d,e) The mean of reconstruction and denoising loss on the test set.}
 \label{fig: mean and std full cca}
\end{figure*}

\subsection{Performance on Synthetic Datasets (RQ2)}
\label{sec: synthetic performance}
We generate the synthetic datasets with different common rates, and the proposed NR-DCCA and other baseline methods are compared.
As shown in Figure~\ref{fig: mean and std full cca}, one can see that the DCCA-based methods (e.g. DCCA, DCCAE, DCCA\_PRIVATE) will encounter model collapse during training, and the variance of accuracy also increases. Linear CCA demonstrates stable performance, while the best accuracy is not as good as DCCA-based methods. Our proposed NR-DCCA achieves state-of-the-art performance as well as training stability to prevent model collapse. 
The results at the final epoch for all common rates are also presented in Table~\ref{table:syn} in Appendix~\ref{appendix:additional results}. 

Considering that we believe that the low-rank property (i.e. highly self-related and redundant) of the weight matrices is the root cause of the model collapse, we utilize NESum to measure the correlation among filters in the weight matrices ( defined in~\ref{appendix: nesum}). Higher NESum represents lower redundancy in weight matrices.
As shown in (b) of Figure \ref{fig: mean and std full cca}, our findings demonstrate that the NR approach effectively reduces filter redundancy, thereby preventing the emergence of low-rank weight matrices and thus averting model collapse.

Moreover, according to our analysis, the correlation should be invariant if neural networks have CIP. Therefore, after training DCCA, DCCAE, and NR-DCCA, we utilize the trained encoders to project the corresponding view data and randomly generated Gaussian white noise and then compute their correlation, as shown in (c) of Figure~\ref{fig: mean and std full cca}. It can be observed that, except for our method (NR-DCCA), as training progresses, other methods increase the correlation between unrelated data. It should be noted that this phenomenon always occurs under any common rates.

Given that the full-rank weight matrix not only produces features that are linearly reconstructed but also discriminates noise in the inputs, we also present the mean value pf Reconstruction and Denoising Loss across different common rates in (d) of Figure~\ref{fig: mean and std full cca}. Notably, NR-DCCA achieves a markedly lower loss, comparable to that observed with Linear CCA, whereas alternative DCCA-based approaches generally lose the above properties.

\subsection{Consistent Performance on Real-world Datasets (RQ3)}
\label{sec: real-world performance}

\begin{figure}[h]
    \centering
    \includegraphics[width=0.99\linewidth]{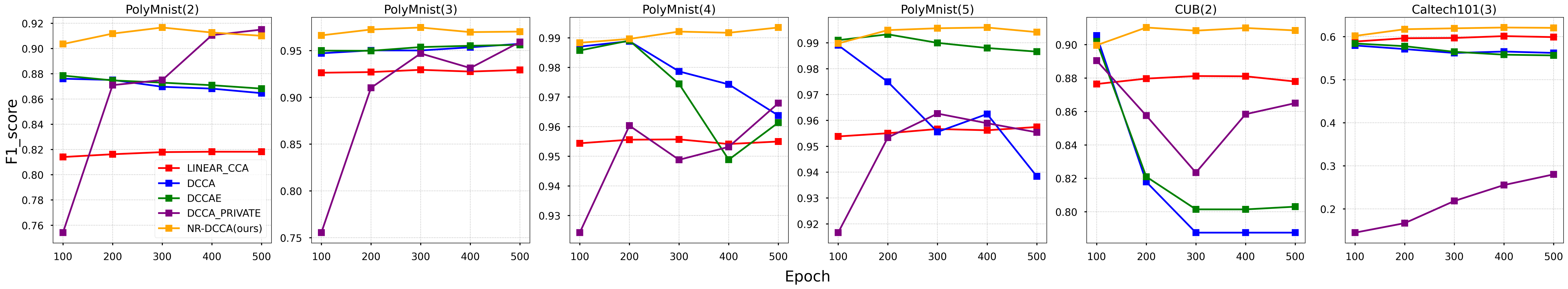}
    \caption
    {Performance of different methods in real-world datasets. 
    % For MVTCAE, we plot its best epoch as the result for each epoch. W
    Each column represents the performance on a specific dataset.
    The number of views in the dataset is denoted in the parentheses next to the dataset name.}
    \label{fig: real_world_cca}
\end{figure}

We further conduct experiments on three real-world datasets: \textbf{PolyMnist}~\citep{sutter2021generalized}, \textbf{CUB}~\citep{2011The}, \textbf{Caltech}~\citep{deng2018triplet}. Additionally, we use a different number of views in PolyMnist.
The results are presented in Figure~\ref{fig: real_world_cca}, and the performance of the final epoch in the figure is presented in Table~\ref{table:syn} in the Appendix~\ref{appendix:additional results}.
Generally, the proposed NR-DCCA demonstrates a competitive and stable performance.
Different from the synthetic data,  the DCCA-based methods exhibit varying degrees of collapse on various datasets, which might be due to the complex nature of the real-world views.
% However, the DCCA-based methods still exhibit varying degrees of collapse as the number of epochs increases.
% It is noteworthy that on the PolyMnist dataset, as the number of views increases, 
% Different from the synthetic data,  the performance of CCA-based methods under-perform the DCCA-based methods, which might be due to the complex nature of the real-world views.
% However, the DCCA-based methods still exhibit varying degrees of collapse as the number of epochs increases.
% It is noteworthy that on the PolyMnist dataset, as the number of views increases, 
% model collapses become more severe for DCCA-based methods, while our NR-DCCA can further improve the performance.
% the performance of all methods improves. However, except for our method, the collapse speeds of other CCA-based deep methods also increase, and the degree of performance decrease becomes more severe, indicating that other methods are not capable enough of handling multiple views. We summarize the results of each method at the $500$-th epoch in Table \ref{real}. KCCA is unable to handle high-dimensional data, so there are no results for this method. PRCCA can only handle views with the same dimension, so there are no results for CUB and Caltech101 datasets.

% \vspace{-1em}

\section{Conclusions}
We propose a novel noise regularization approach for DCCA in the context of MVRL, and it can prevent model collapse during the training, which is an issue observed and analyzed in this paper for the first time. 
Specifically, we theoretically analyze the CIP in Linear CCA and demonstrate that it is the key to preventing model collapse. To this end, we develop a novel NR approach to equip DCCA with such a property (NR-DCCA). Additionally, synthetic datasets with different common rates are generated and tested, which provide a benchmark for fair and comprehensive comparisons of different MVRL methods. The NR-DCCA developed in the paper inherits the merits of both Linear CCA and DCCA to achieve stable and consistent outperformance in both synthetic and real-world datasets. More importantly, the proposed noise regularization approach can also be generalized to other DCCA-based methods ({\em e.g.}, DGCCA).

% apply noise regularization to neural networks, introducing implicit constraints that enhance stability during the training process and improve the model's ability to handle data with varying degrees of correlation.
% To comprehensively evaluate the stability of multi-view representation learning (MVRL) approaches, we construct synthetic MVRL datasets, which have not been previously explored. Furthermore, extensive experiments demonstrate that our approach achieves state-of-the-art results on both synthetic and real-world datasets.

% In future work,  we plan to extend the application of our noise regularization method to other multi-view learning scenarios, such as multi-view classification and clustering.
In future studies, we wish to explore the potential of noise regularization in other representation learning tasks, such as contrastive learning and generative models. It is also interesting to further investigate the difference between our developed NR and other neural network regularization approaches, such as orthogonality regularization~\citep{bansal2018can,huang2020controllable} and weight decay~\citep{loshchilov2017decoupled,zhang2018three,krogh1991simple}. 
% Whether the Gaussian white noise is the optimal choice of the noise distribution for constructing the NR loss is also worth exploring. 
Our ultimate goal is to make the developed noise regularization a pluggable and useful module for neural network regularization.

% Furthermore, we can explore its potential in other areas of representation learning, including contrastive learning and generative models.
\clearpage

% \section*{Reproducibility Statement}
% All of our experiments are conducted with fixed random seeds and all the performance of downstream tasks is the average value of a 5-fold cross-validation. We use one single 3090 GPU. 
% The CCA-zoo package is adopted as the implementation of various CCA/GCCA-based methods, and the original implementation of MVTCAE is employed. Both baselines and our developed NR-DCCA/NR-DGCCA are implemented in the same PyTorch environment (see requirements.txt in the source codes).
% All the datasets used in the paper are either provided or open datasets. Detailed proofs of all the Theorems in the main paper can be found in the Appendix. Both source codes and appendix can be downloaded from the supplementary material. 

% The developed NR-DCCA/NR-DGCCA and other baselines are implemented in PyTorch and source codes can be downloaded from the supplementary materials. All the datasets used in the paper are either provided or open datasets. Detailed proofs of all the propositions in the main paper can be found in the Appendix. 

\section*{Acknowledgments}
The work described in this paper was supported by the Research Grants Council of the Hong Kong Special Administrative Region, China (Project No. PolyU/25209221 and PolyU/15206322), and grants from the Otto Poon Charitable Foundation Smart Cities Research Institute (SCRI) at the Hong Kong Polytechnic University (Project No. P0043552). The contents of this article reflect the views of the authors, who are responsible for the facts and accuracy of the information presented herein.

\bibliography{neurips_2024}

\clearpage
\appendix
\section{Appendix}
\subsection{Proof of Theorem~\ref{thm:cca}}
\subsubsection{Preparations}
To prove Theorem~\ref{thm:cca}, we need first to prove the following Lemmas.
%~\ref{lemma: zero-center},~\ref{lemma:mpi-cca},~\ref{lemma:rank-and-trace},~\ref{lemma: less-rank}:
\begin{lemma}
    \label{lemma: zero-center}
    Given a specific matrix $B$ and a zero-centered $C$ with respect to rows, the product $BC$ is also zero-centered with respect to rows.
\end{lemma}
Proof of Lemma~\ref{lemma: zero-center}:

Let $B_{i,j}$ and $C_{i,j}$  denote the $(i,j)$-th entry of $B$ and $C$, respectively. Then we have:
$$(BC)_{i,j} = \sum_{r=1} B_{i,r} C_{r,j}$$
Since each row of $C$ has a mean of $0$, we have $\sum_{j=1}^{n} C_{r,j} = 0, \forall r$. For the mean value of $i$-th row of $BC$, we can write:
\begin{equation}
\begin{aligned}
\frac{1}{n} \sum_{j=1}^{n} (BC)_{i,j} &= \frac{1}{n} \sum_{j=1}^{n} \sum_{r=1} B_{i,r} C_{r,j} \\
&= \frac{1}{n} \sum_{r=1} \sum_{j=1}^{n}  B_{i,r} C_{r,j} \\
&= \frac{1}{n} \sum_{r=1} B_{i,r} \left( \sum_{j=1}^{n} C_{r,j} \right) \\
&= \frac{1}{n} \sum_{r=1} B_{i,r} \cdot 0 \\
&= 0
\end{aligned}
\end{equation}

The Moore-Penrose Inverse (MPI)~\citep{petersen2008matrix} will be used for analysis, and the MPI is defined as follows:
\begin{definition}
Given a specific matrix $Y$, its Moore-Penrose Inverse (MPI) is denoted as $Y^+$. $Y^+$ satisfies: $YY^+Y = Y$, $Y^+YY^+ = Y^+$, $YY^+$ is symmetric, and $Y^+Y$ is symmetric. 
\end{definition}
The MPI $Y^+$ is unique and always exists for any $Y$. Furthermore, when matrix $Y$ is invertible, its inverse matrix $Y^-$ is exactly $Y^+$. 
Using the definition of MPI, we can rewrite the formulation of CCA. In particular, $\text{Corr}(\cdot, \cdot)$ can be derived by replacing the inverse with MPI. Using $\text{Corr}(X_k,A_k)$ as an example, the following Lemma holds:
\begin{lemma}[MPI-based CCA]
\label{lemma:mpi-cca}
For the  $k$-th view data $X_k$ and the Gaussian white noise $A_k$, we have
\begin{equation}
\text{Corr}(X_k,A_k)  =   \frac{1}{(n-1)^2} \text{tr}(A_k^{+}A_kX_k^{+}X_k)^{1/2}, \forall k. 
\end{equation}
\end{lemma}
Proof of Lemma~\ref{lemma:mpi-cca}:

\begin{equation}
\begin{aligned}
 \text{Corr}(X_k,A_k) 
     & = \text{tr}((\Sigma_{11}^{-1/2}\Sigma_{12}\Sigma_{22}^{-1/2})'\Sigma_{11}^{-1/2}\Sigma_{12}\Sigma_{22}^{-1/2})^{1/2} \\
      & = \text{tr}(\Sigma_{22}^{-1/2}\Sigma_{12}'\Sigma_{11}^{-1/2}\Sigma_{11}^{-1/2}\Sigma_{12}\Sigma_{22}^{-1/2})^{1/2} \\
     & = \text{tr}(\Sigma_{22}^{-1/2}\Sigma_{22}^{-1/2}\Sigma_{12}'\Sigma_{11}^{-1/2}\Sigma_{11}^{-1/2}\Sigma_{12})^{1/2} \\
      & = \text{tr}(\Sigma_{22}^{-1}\Sigma_{12}'\Sigma_{11}^{-1}\Sigma_{12})^{1/2} \\
     & = \frac{1}{(n-1)^2} \text{tr}((A_kA_k')^{-1}(
     X_kA_k')'(X_kX_k')^{-1}(X_kA_k'))^{1/2}\\
    & = \frac{1}{(n-1)^2} \text{tr}((A_kX_k')^{-1}(
    A_kX_k')(X_kX_k')^{-1}(X_kA_k'))^{1/2} \\
     & = \frac{1}{(n-1)^2} \text{tr}((A_kA_k')^{+}(A_kX_k')(X_kX_k')^{+}(X_kA_k'))^{1/2} \\
     & = \frac{1}{(n-1)^2} \text{tr}(A_k'(A_kA_k')^{+}A_kX_k'(X_kX_k')^{+}X_k)^{1/2}     \\
     & =  \frac{1}{(n-1)^2} \text{tr}(A_k^{+}A_kX_k^{+}X_k)^{1/2}   
\end{aligned}
\end{equation}
The first row is based on the definition of $\text{Corr}$, the second row is because the trace is invariant under cyclic permutation, the fifth row is to replace matrix inverse by MPI and the ninth row is due to $Y^+ = Y'(YY^+)$~\citep{petersen2008matrix}.

\begin{lemma}
\label{lemma:rank-and-trace}
Given a specific matrix $Y$ and its MPI  $Y^{+}$, let $\text{Rank}(Y)$ and $\text{Rank}(Y^{+}Y)$ be the ranks of $Y$ and $Y^{+}Y$, respectively. It is true that:
$$\text{Rank}(Y) = \text{Rank}(Y^{+}Y)$$
$$\text{Rank}(Y^{+}Y) = \text{tr}(Y^{+}Y)$$
\end{lemma}
Proof of Lemma~\ref{lemma:rank-and-trace}:

Firstly, the column space of $Y^{+}Y$ is a subspace of the column space of $Y$. Therefore, $\text{Rank}(Y^{+}Y) \leq \text{Rank}(Y)$.
On the other hand, according to the definition of MPI~\citep{petersen2008matrix}, we know that $Y=Y(Y^{+}Y)$. Since the rank of a product of matrices is at most the minimum of the ranks of the individual matrices, we have $\text{Rank}(Y) \leq \text{Rank}(Y^{+}Y)$.
Combining the two inequalities, we have $\text{Rank}(Y) = \text{Rank}(Y^{+}Y)$.
Furthermore, since $(Y^{+}Y)(Y^{+}Y) = Y^{+}Y$ (it holds that $Y^+=Y^{+}YY^{+}$ according to the definition of MPI~\citep{petersen2008matrix}), $Y^{+}Y$ is an idempotent and symmetric matrix, and thus its eigenvalues must be $0$ or $1$. So the sum of its eigenvalues is exactly its rank. Considering matrix trace is the sum of eigenvalues of matrices, we have $\text{Rank}(Y^{+}Y) = \text{tr}(Y^{+}Y)$.

\begin{lemma}
\label{lemma: less-rank}
$\text{Rank}(W_kX_k) < \text{Rank}(X_k)$ 
, when $W_k$ is not a full-rank matrix and $X_k$ is a full-rank matrix.
\end{lemma}
Proof of Lemma~\ref{lemma: less-rank}:
Since the rank of a product of matrices is at most the minimum of the ranks of the individual matrices, we have $\text{Rank}(W_kX_k) \leq min(\text{Rank}(W_k), \text{Rank}(X_k))$.
Considering $X_k$ is full-rank, $\text{Rank}(X_k) = min(d_k,n)$ and then $\text{Rank}(W_kX_k) \leq min(\text{Rank}(W_k), \text{Rank}(X_k)) =  min(\text{Rank}(W_k), min(d_k,n))$.
Since $W_k$ is not full-rank, we have $\text{Rank}(W_k) < d_k$.
In conclusion, $\text{Rank}(W_kX_k) < min(d_k, min(d_k,n))$ and then $\text{Rank}(W_kX_k) < d_k \leq \text{Rank}(X_k)$.

\subsubsection{Main proofs of Theorem~\ref{thm:cca}}
We prove the two directions of Theorem~\ref{thm:cca} in the following two Lemmas. First, we prove CIP holds if $W_k$ is a square and full-rank matrix.
\begin{lemma}
\label{lemma:ccaequal}
For any $k$, if $W_k$ is a square and full-rank matrix, the correlation between $X_k$ and $A_k$ remains unchanged before and after the transformation by $W_k$ (i.e. CIP holds for $W_k$). Mathematically, we have $\text{Corr}(X_k,A_k) = \text{Corr}(W_kX_k, W_kA_k)$.
\end{lemma}

Proof of Lemma~\ref{lemma:ccaequal}:

Firstly, we have the $k$-th view data $X_k$ to be full-rank, as we can always delete the redundant data, and the random noise $A_k$ is full-rank as each column is generated independently. 
Without loss of generality, we assume that all the datasets $X_k$ are zero-centered with respect to row~\citep{hotelling1992relations}, which implies that $W_kA_k$ and $W_kX_k$ are both zero-centered matrices \ref{lemma: zero-center}. When computing the covariance matrix, there is no need for an additional subtraction of the mean of row, which simplifies our subsequent derivations.
And $W_k$ is always full-rank since Linear CCA seeks full-rank $W_kX_k$.
Then by utilizing Lemma~\ref{lemma:mpi-cca}, we derive that the correlation between $X_k$ and $A_k$ remains unchanged before and after the transformation:

\begin{equation}
\resizebox{0.99\linewidth}{!}{$
\begin{aligned}
 \text{Corr}(W_kX_k,W_kA_k) 
          & = \frac{1}{(n-1)^2} 
             \text{tr}((W_kA_k)^{+}W_kA_k(W_kX_k)^{+}W_kX_k)^{1/2} \\
          & = \frac{1}{(n-1)^2} \text{tr}((W_k^{+}W_kA_k)^{+}(W_kA_kA_k^{+})^{+}W_kA_k(W_k^{+}W_kX_k)^{+}(W_{k}X_kX_k^{+})^{+}W_{k}X_k)^{1/2} \\
          & = \frac{1}{(n-1)^2} \text{tr}((W_{k}^{+}W_{k}A_k)^{+}W_{k}^{+}W_{k}A_k(W_{k}^{+}W_{k}X_k)^{+}W_{k}^{+}W_{k}X_k)^{1/2} \\
          & =  \frac{1}{(n-1)^2} \text{tr}(A_k^{+}A_kX_k^{+}X_k)^{1/2}\\
          & = \text{Corr}(X_k,A_k)
\end{aligned}
$}
\end{equation}
The first row is based on Lemma~\ref{lemma:mpi-cca}, the second row is because given two matrices $B$ and $C$, $(BC)^+ = (B^+BC)^+(BC^+C)^+$ always holds~\citep{petersen2008matrix}, and 
the third row utilizes the properties of full-rank and square matrix $W_k$: $W_k^{+} = W_k^{-}$, which means $W_k^+W_k = W_kW_k^+ = I_{d_k}$~\citep{petersen2008matrix}.

Then we prove that  $W_k$ is a full-rank matrix if CIP holds and $W_k$ is square.
\begin{lemma}
\label{lemma:nr loss}
For any $k$, if $\text{Corr}(X_k,A_k) = \text{Corr}(W_kX_k, W_kA_k)$ and $W_k$ is a square matrix, then $W_k$ must be a full-rank matrix.  
\end{lemma}

Proof of Lemma~\ref{lemma:nr loss}:

This Lemma is equivalent to its contra-positive proposition: 
if $W_k$ is not a full-rank matrix, there exists random noise data $A_k$ such that $\eta_k = \left| Corr(W_kX_{k},W_k(A_{k})) - Corr(X_{k}, A_{k}) \right|$ is not $0$.
And we find that when $W_k$ is not full-rank, there exists $A_k = X_k$  such that $ \eta_k  \neq 0$.
We have the following derivation:

\begin{equation}
\resizebox{0.99\linewidth}{!}{$
\begin{aligned}
 \eta_k & = \left| Corr(W_kX_{k},W_kA_{k}) - Corr(X_{k},A_{k}) \right| \\
        & =  \left| \frac{1}{(n-1)^2}\text{tr}((W_kA_k)^{+}(W_kA_k)(W_kX_k)^{+}(WX_k))^{1/2}- \frac{1}{(n-1)^2}\text{tr}(A^{+}AX^{+}X)^{1/2} \right| \\
        & =  \left| \frac{1}{(n-1)^2}\text{tr}((W_k^{+}W_kA_k)^{+}(W_kA_kA_k^{+})^{+}W_kA_k(W_k^{+}W_kX_k)^{+}(W_kX_kX_k^{+})^{+}W_kX_k)^{1/2}  
        - \frac{1}{(n-1)^2}\text{tr}(A_k^{+}A_kX_k^{+}X_k)^{1/2} \right| \\
        & = \left| \frac{1}{(n-1)^2}\text{tr}((W_k^{+}W_kA_k)^{+}W_k^{+}W_kA_k(W_k^{+}W_kX_k)^{+}W_k^{+}W_kX_k)^{1/2}  
        - \frac{1}{(n-1)^2}\text{tr}(A_k^{+}AX_k^{+}X_k)^{1/2} \right|
\end{aligned}
$}
\end{equation}
The first row is the definition of NR loss  with respect to $W_k$, 
the second row is based on the new form of CCA, the third row is because given two specific matrices $B$ and $C$, it holds the equality $(BC)^+ = (B^+BC)^+(BC^+C)^+$~\citep{petersen2008matrix}, and the fourth row utilizes the properties of full-rank matrix: for full-rank matrices $X_k$ and $A_k$, whose sample size is larger than dimension size, they fulfill: $X_k{X_k}^{+} = I_{d_k}, A_k{A_k}^{+} = I_{d_k}$ (given a specific full-rank matrix
$Y$, if its number of rows is smaller than that of cols, it holds that $Y^{+}=Y'(YY')^-$, which means that $YY^{+} = I$)~\citep{petersen2008matrix}. 

Let us analyze the case when  $A_k = X_k$:
\begin{equation}
\label{equal:A_k = X_k}
\resizebox{0.99\linewidth}{!}{$
\begin{aligned}
\eta_k  & = \left| \frac{1}{(n-1)^2}\text{tr}((W_k^{+}W_kX_k)^{+}W_k^{+}W_kX_k(W_k^{+}W_kX_k)^{+}W_k^{+}W_kX_k)^{1/2} - \frac{1}{(n-1)^2}\text{tr}(X_k^{+}X_kX_k^{+}X_k)^{1/2}\right| \\
 & = \left|\frac{1}{(n-1)^2}\text{tr}((W_k^{+}W_kX_k)^{+}W_k^{+}W_kX_k)^{1/2}- \frac{1}{(n-1)^2}\text{tr}(X_k^{+}X_k)^{1/2}\right|.
\end{aligned}
$}
\end{equation}
The first row is to replace $A_k$ with $X_k$, the second row is because $X_k^{+}X_kX_k^{+} = X_k^{+}$ and $(W_k^{+}W_kX_k)^{+}W_k^{+}W_kX_k(W_k^{+}W_kX_k)^{+} = W_k^{+}W_kX_k$, which are based on the definition of MPI that given a specific matrix $Y$, $Y^{+}YY^{+} = Y^{+}$~\citep{petersen2008matrix}.

As a result, we can know that when the random noise data $A_k$ is exactly $X_k$ and $W_k$ is not full-rank, $\eta_k$ can not be zero: 
\begin{equation}
\begin{aligned}
 \eta_k 
 & = \left|\frac{1}{(n-1)^2}\text{tr}((W_k^{+}W_kX_k)^{+}W_k^{+}W_kX_k)^{1/2}- \frac{1}{(n-1)^2}\text{tr}(X_k^{+}X_k)^{1/2}\right| \\
 & = \left|\frac{1}{(n-1)^2} \text{Rank}(W_k^{+}W_kX_k)^{1/2} - \frac{1}{(n-1)^2} \text{Rank}(X)^{1/2}\right| \\
& \neq \left|\frac{1}{(n-1)^2} \text{Rank}(X_k)^{1/2} - \frac{1}{(n-1)^2} \text{Rank}(X_k)^{1/2}\right| \\
& \neq 0
\end{aligned}
\end{equation}
The first row is due to Equation~\ref{equal:A_k = X_k}, the second row is based on Lemma~\ref{lemma:rank-and-trace} that $\text{tr}((W_k^{+}W_kX_k)^{+}W_k^{+}W_kX_k) = \text{Rank}(W_k^{+}W_kX_k)$ and $\text{tr}(X_k^{+}X_k) = \text{Rank}(X)$, and the third row is because of Lemma \ref{lemma: less-rank}.

Finally, if $\eta_k$ is always constrained to $0$ for any $A_k$, then $W_k$ must be a full-rank matrix.

Combining Lemma~\ref{lemma:ccaequal} and~\ref{lemma:nr loss}, we complete the proof.

\subsection{Proof of Theorem~\ref{thm: CIP on Obtained Representations}}
% \in \mathbb{R}^{d_{k} \times n}$ ($d_k<n$)
For linear regression problem : $R^* = \arg \min_{ R } \| RB-C \|_F$, where $R \in \mathbb{R}^{out\_dim \times input\_dim}$ is the weight matrix, and $B \in \mathbb{R}^{input\_dim \times n} ,C \in \mathbb{R}^{output\_dim \times n}$ are the input and target matrix ($input\_dim<n$), respectively. $R^*$ has a closed-form solution:
$R^* = CB'(BB')^{-}$ and therefore, it holds that: 
\begin{equation}
\begin{aligned}
\min \| RB-C \|_F & =  \|R^*B-C \|_F \\
& \overset{\text{a}}{=} \|CB'(BB')^{-}B-C \|_F \\
& \overset{\text{b}}{=} \|CB'(BB')^{+}B-C \|_F  \\ 
& \overset{\text{c}}{=} \|CB^+B-C \|_F 
\end{aligned}
\label{equal: min_mse}
\end{equation}
Equation $a$ is the use of the closed-form solution and Equation $b$ is to replace the matrix inverse by MPI. Equation $c$ is because $Y'(YY')^+ = Y^+$~\citep{petersen2008matrix}.

Given $k$-th view,  when $W_k$ is square possesses CIP, $W_k$ is full-rank.
Using the above equation, we have:
\begin{equation}
\begin{aligned}
\min \| P_kW_kX_k-X_k \|_F & \overset{\text{a}}{=} \|X_k(W_kX_k)^+(W_kX_k)-X_k \|_F \\
& \overset{\text{b}}{=} \|X_k(W_k^+W_kX_k)^+(W_kX_kX_k^+)^+(W_kX_k)-X_k \|_F \\
& \overset{\text{c}}{=} \|X_k(W_k^+W_kX_k)^+W_k^+W_kX_k-X_k \|_F \\
& \overset{\text{d}}{=} \|X_kX_k^+X_k-X_k \|_F \\
& = 0
\end{aligned}
\end{equation}
Equation $a$ is due to Equation \ref{equal: min_mse}~\citep{petersen2008matrix}. Equation $b$ holds because  given two matrices $B$ and $C$, $(BC)^+ = (B^+BC)^+(BC^+C)^+$ always holds and Equation $c$ is because for full-rank matrix $X_k \in \mathbb{R}^{d_k \times n} (d_k<n)$, $X_kX_k^+ = I_{d_k}$. Equation $c$ utilizes the properties of full-rank and square matrix $W_k$: $W_k^{+} = W_k^{-}$, which means $W_k^+W_k = W_kW_k^+ = I_{d_k}$~\citep{petersen2008matrix}.
Equation $d$ is based on the definition of MPI: given a specific matrix $Y$ and its $Y^+$, it holds that $YY^+Y = Y$. We show the first property in Theorem~\ref{thm: CIP on Obtained Representations}.

As for the second property: 
\begin{equation}
\resizebox{0.99\linewidth}{!}{$
\begin{aligned}
\min \| Q_kW_k(X_k+A_k)-W_kX_k \|_F & \overset{\text{a}}{=} \|W_kX_k(W_k(X_k+A_k))^+(W_k(X_k+A_k))-W_kX_k \|_F \\
& \overset{\text{b}}{=} \|W_kX_k(W_k^+W_k(X_k+A_k))^+(W_k(X_KA_k)(X_k+A_k)^+)^+(W_k(X_k+A_k))-W_KX_k \|_F \\
& \overset{\text{c}}{=} \|W_kX_k(W_k^+W_k(X_k+A_k))^+W_k^+W_k(X_k+A_k)-W_KX_k \|_F \\
& \overset{\text{e}}{=} \|W_kX_k(X_k+A_k)^+(X_k+A_k)-W_KX_k \|_F \\
& \overset{\text{f}}{=} \|W_k(X_k+A_k)(X_k+A_k)^+(X_k+A_k) -W_kA_k(X_k+A_k)^+(X_k+A_k) -W_KX_k\|_F \\
& \overset{\text{g}}{=} \|W_k(X_k+A_k) -W_kA_k(X_k+A_k)^+(X_k+A_k) -W_KX_k\|_F \\
& \overset{\text{h}}{=} \|W_kA_k -W_kA_k(X_k+A_k)^+(X_k+A_k)\|_F \\
& \overset{\text{i}}{=} \|W_kA_k(I_{n} - (X_k+A_k)^+(X_k+A_k))\|_F \\
& \overset{\text{j}}{\leq} \|W_kA_k\|_F * \|(I_{n} - (X_k+A_k)^+(X_k+A_k))\|_F \\
& \overset{\text{k}}{=} \|W_kA_k\|_F * \sqrt{\text{tr}(I_{n} - (X_k+A_k)^+(X_k+A_k))} \\
& \overset{\text{l}}{=} \|W_kA_k\|_F * \sqrt{\text{Rank}(I_{n} - (X_k+A_k)^+(X_k+A_k))} \\
& \leq \sqrt{n} \|W_kA_k\|_F
\end{aligned}
$}
\end{equation}
Equation $a$ is because of Equation \ref{equal: min_mse}. Equation $b$ holds because  given two matrices $B$ and $C$, $(BC)^+ = (B^+BC)^+(BC^+C)^+$ always holds and Equation $c$ is because we assume $X_k+A_k$ is a full-rank matrix. Equation $e$ utilizes the properties of full-rank and square matrix $W_k$: $W_k^+W_k = W_kW_k^+ = I_{d_k}$. Equation $g$ is based on the definition of MPI: $(X_k+A_k)(X_k+A_k)^+(X_k+A_k) = (X_k+A_k)$. Equation $j$ holds because given two specific matrices $B$ and $C$, $\|BC\|_F \leq \|B\|_F * \|C\|_F$~\citep{belitskii2013matrix}. Equation $k$ and $l$ is because given a specific matrix $B$, $I - B^{+}B$ is an idempotent matrix and $\|I -  B^{+}B\|_F = \sqrt{\text{tr}((I -  B^{+}B)'(I -  B^{+}B))} = \sqrt{\text{tr}(I -  B^{+}B)}$.

Now, we use $(W_kA_k)_(i,j)$ to donate the $(i,j)$-th entry of $W_kA_k$ and the expected value of the square of the Frobenius norm of $W_kA_k$ is:
\begin{equation}
    \mathbb{E}\left[ \| WA_k \|_F^2  \right] = \mathbb{E}\left[ \sum_{i}\sum_{j}  \left[(W_kA_k)_{i,j} \right]^2 \right]
\end{equation}
Expanding the product \( (W_kA_k)_{i,j} \), we have:
\begin{equation}
    (W_kA_k)_{i,j} = \sum_{r} (W_k)_{i,r} (A_k)_{r,j}
\end{equation}
Substituting back into the expectation, we get:
\begin{equation}
    \mathbb{E}\left[ \| WA_k \|_F^2  \right] = \mathbb{E}\left[ \sum_{i}\sum_{j}  \left[ \sum_{r} (W_k)_{i,r} (A_k)_{r,j} \right]^2 \right] =  \sum_{i}\sum_{j}   \mathbb{E}\left[\left[ \sum_{r} (W_k)_{i,r} (A_k)_{r,j} \right]^2 \right]
\end{equation}
Since the elements of \( A_k \) are i.i.d. with zero mean and unit variance, the expectation of their squares is 1, and the cross terms vanish due to the zero mean. Therefore, we have:
\begin{equation}
    \sum_{i}\sum_{j} \mathbb{E}\left[\left[ \sum_{r} (W_k)_{i,r} (A_k)_{r,j} \right]^2 \right] = \sum_{i}\sum_{r} (W_k)_{ir}^2 = \| W_k \|_F^2
\end{equation}
Hence, we have shown that

\begin{equation}
    \mathbb{E}\left[ \| W_kA_k \|_F^2  \right] = \| W_k \|_F^2
\end{equation}
This completes the proof.

% The first row is based on the definition of $\text{Corr}$ and the replacement of matrix inverse by MPI, the second row is because trace is invariant under cyclic permutation, and the third row is due to $Y^+ = Y'(YY^+)$ \citep{petersen2008matrix}.

\subsection{Details of Datasets and Baselines}
\label{appendix: Datasets and Baselines}
\textbf{Synthetic datasets}:

All the datasets used in the paper are either provided or open datasets. Detailed proofs of all the Theorems in the main paper can be found in the Appendix. Both source codes and appendix can be downloaded from the supplementary material. 

We make $6$ groups of multi-view data originating from the same $G \in \mathbb{R}^{d \times n}$ (we set $n=4000, d=100$). Each group consists of tuples with $2$ views ($2000$ tuples for training and $2000$ tuples for testing) and a distinct common rate. Common rates of these sets are from $\{0\%, 20\%, 40\%, 60\%, 80\%, 100\%\}$ and there are $50$ downstream regression tasks. We report the mean and standard deviation of $\text{R2}$ score across all the tasks.

\textbf{Real-world datasets}:

\textbf{PolyMnist}~\citep{sutter2021generalized}:
A dataset consists of tuples with $5$ different MNIST images ($60,000$ tuples for training and $10,000$ tuples for testing). 
Each image within a tuple possesses distinct backgrounds and writing styles, yet they share the same digit label. The background of each view is randomly cropped from an image and is not used in other views. Thus, the digit identity represents the common information, while the background and writing style serve as view-specific factors. The downstream task is the digit classification task.
\textbf{CUB}~\citep{2011The}:
A dataset consists of tuples with deep visual features ($1024$-d) extracted by \textsc{GoogLeNet} and text features ($300$-d) obtained through \textsc{Doc2vec}~\citep{le2014distributed} ($480$ tuples for training and $600$ tuples for testing).
This MVRL task utilizes the first 10 categories of birds in the original dataset and the downstream task is the bird classification task.
\textbf{Caltech}~\citep{deng2018triplet}:
A dataset consists of tuples with traditional visual features extracted from images that belong to 101 object categories, including an additional background category ($6400$ tuples for training and $9144$ tuples for testing). Following \cite{yang2021partially}, three features are used as views: a $1,984$-d HOG feature, a $512$-d GIST feature, and a $928$-d SIFT feature. 

\textbf{Baselines}:
All of our experiments are conducted with fixed random seeds and all the performance of downstream tasks is the average value of a 5-fold cross-validation. We use one single 3090 GPU. 
The CCA-zoo package is adopted as the implementation of various CCA/GCCA-based methods, and the original implementation of MVTCAE is employed. Both baselines and our developed NR-DCCA/NR-DGCCA are implemented in the same PyTorch environment (see requirements.txt in the source codes).

Direct method:
\begin{itemize}
    \item \textbf{CONCAT} straightforwardly concatenates original features from different views.
\end{itemize}
CCA methods:
\begin{itemize}
    % \item  \textbf{CCA}~\citep{hotelling1992relations} maps multiple views' data into a common space that maximizes their correlation and concatenates the new representations of different views. 
    \item  \textbf{PRCCA}~\citet{tuzhilina2023canonical} preserves the internal data structure by grouping high-dimensional data features while applying an l2 penalty to CCA,
    \item  \textbf{Linear CCA}~\citep{wang2015deep} employs individual linear layers to project multi-view data and then maximizes their correlation defined in \citep{hotelling1992relations}. 
    \item  \textbf{Linear GCCA}  uses linear layers to maximize the correlation of multi-view data defined in \citep{benton2017deep}.
\end{itemize}
Kernel CCA Methods:
\begin{itemize}
    \item \textbf{KCCA}~\citep{akaho2006kernel} employs CCA methods through positive-definite kernels. 
\end{itemize}
DCCA-based methods:
\begin{itemize}
    \item \textbf{DCCA}~\citep{andrew2013deep} employs neural networks to individually project multiple sets of views, obtaining new representations that maximize the correlation between each pair of views.
    \item \textbf{DGCCA}~\citep{benton2017deep} constructs a shared representation and maximizes the correlation between each view and the shared representation.
    \item \textbf{DCCA\_EY}~\citep{chapman2022generalized} optimizes the objective of CCA via a sample-based EigenGame.
    \item \textbf{DCCA\_GHA}~\citep{chapman2022generalized} solves the objective of CCA by a sample-based generalized Hebbian algorithm.

    \item \textbf{DCCAE/DGCCAE}~\citep{wang2015deep} introduces reconstruction objectives to DCCA, which simultaneously optimize the canonical correlation between the learned representations and the reconstruction errors of the autoencoders. 
    \item \textbf{DCCA\_PRIVATE/DGCCA\_PRIVATE}~\citep{wang2016deep} incorporates dropout and private autoencoders, thus preserving both shared and view-specific information. 
\end{itemize}
Information theory-based methods:
\begin{itemize}
    \item \textbf{MVTCAE}~\citep{hwang2021multi} maximizes the reduction in Total Correlation to capture both shared and view-specific factors of variations. 
\end{itemize}
All CCA-based methods leverage the implementation of CCA-Zoo~\citep{chapman2021cca}. To ensure fairness, we use the official implementation of MVTCAE  while replacing the strong CNN backbone with MLP.
    
% Privous

% \subsection{Hyper-parameter Settings}
% \label{appendix:Hyper-paramter Settings}
% To ensure a fair comparison, we tune the hyper-parameters of all baselines within the ranges suggested in the original papers, except for the following fixed settings: 

% The embedding size for the real-world dataset is set as $200$, while the size for synthetic datasets is set as $100$. Batch size is $\min(2000, \text{full-size})$. The same MLP architectures are used for DCCA-based methods.

% In the synthetic datasets, DCCA, DGCCA, DCCAE, and DGCCAE methods utilize a minimum learning rate of $5e-3$.  DCCA\_PRIVATE/DGCCA\_PRIVATE employ a slightly higher learning rate of $1e-2$. In contrast, our proposed methods, NR-DCCA/NR-DGCCA, utilize the maximum learning rate of $1.5e-2$. And the regularization weight $\alpha$ is set as $200$.

% In the real-world datasets, the learning rates in PolyMnist and CUB are set to $1e-4$. For Caltech101, a slightly lower learning rate of $5e-5$ is used. And the regularization weight $\alpha$ is tuned from $\{1.0,1.5,5,15\}$.
% To expedite the computation of $\text{Corr}(X_k, A_k)$, on the PolyMnist dataset, we utilize the initialized $f_k$ to reduce the feature dimensions of $X_k$ and $A_k$ separately. Subsequently, we calculate their correlation. For the extracted features in the CUB and Caltech101 datasets, we simply employ $X_k[:out_dim,:]$ and $A_k[:out_dim,:]$ to compute of $\text{Corr}$.

% Revised
\subsection{Implementation Details of Synthetic Datasets}
\label{appendix:implementation of Synthetic datasets}

We draw $n$ random samples with dim $d$ from a Gaussian distribution as $G \in \mathbb{R}^{d \times n}$ to represent complete representations of $n$ objects.
We define the non-linear transformation $\phi_k$ as the addition of noise to the data, followed by passing it through a randomly generated MLP. To generate the data for the $k$-th view, we select specific feature dimensions from $G$ based on a given common rate~\ref{def:common rate} and then apply $\phi_k$ to those selected dimensions.
And we define $\psi_j$ as a linear layer, and task $T_j$ is generated by directly passing G through $\psi_j$.

\subsection{Hyper-parameter Settings}
\label{appendix:Hyper-paramter Settings}
To ensure a fair comparison, we tune the hyper-parameters of all baselines within the ranges suggested in the original papers, including hyper-parameter $r$ of ridge regularization, except for the following fixed settings: 

The embedding size for the real-world datasets is set as $200$, while the size for synthetic datasets is set as $100$. Batch size is $\min(2000, \text{full-size})$. The same MLP architectures are used for D(G)CCA-based methods.
The hyper-parameter $r$ of ridge regularization is set as $0$ in our NR-DCCA and NR-DGCCA. And the best $r$ for Linear (G)CCA and D(G)CCA-based methods is tuned on the validation data (synthetic datasets and PolyMnist: 1e-3, CUB and Caltech101 : 0). 

In the synthetic datasets, Linear CCA and Linear GCCA use a minimum learning rate of $1e-4$, DCCA, DGCCA, DCCAE, and DGCCAE methods utilize a bigger learning rate of $5e-3$.  DCCA\_PRIVATE/DGCCA\_PRIVATE employ a slightly higher learning rate of $1e-2$. In contrast, our proposed methods, NR-DCCA/NR-DGCCA, utilize the maximum learning rate of $1.5e-2$. And the regularization weight $\alpha$ is set as $200$.

In the real-world datasets, the learning rates for all deep methods are set to $1e-4$ while that of Linear CCA and Linear GCCA are $1e-5$. 
To expedite the computation of $\text{Corr}(X_k, A_k)$,  in the real-world datasets, we simply employ $X_k[:outdim,:]$ and $A_k[:outdim,:]$ to compute of $\text{Corr}(X_k, A_k)$.  The optimal $\alpha$ values of NR-DCCA for the CUB, PolyMnist, and Caltech datasets are found to be 1.5, 2, and 10,  respectively.

\subsubsection{Hyper-parameter $r$ in Ridge Regularization}
\label{appendix:Hyper-paramter of ridge regularization}
\begin{figure}[h]
    \centering
    \includegraphics[width=1\linewidth]{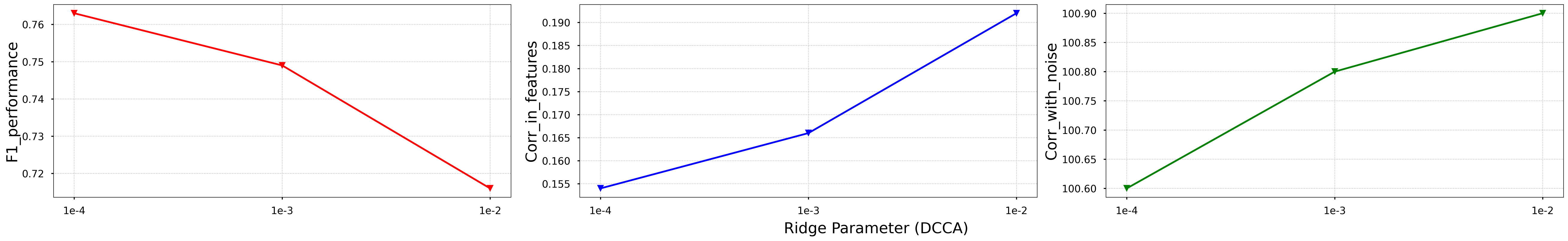}
    \caption
    {The effects of hyper-parameter $r$ of DCCA in the CUB dataset.}
    \label{fig:ridge_in_cub}
\end{figure}

In this section, we discuss the effects of hyper-parameter $r$ in ridge regularization.
Ridge regularization is commonly used across almost all (D)CCA methods, which improves numerical stability. It works by adding an identity matrix $I$ to the estimated covariance matrix. 
% Here is an example for regularize covariance matrix $\Sigma_{11}$ of data $X_1$:
% $$\Sigma_{11}' = \Sigma_{11} + r*I$$, where $r$ is the hyper-parameter of ridge regularization. 
However, ridge regularization mainly regularizes the features, rather than the transformation (i.e., $W_k$ in Linear CCA and $f_k$ in DCCA) and it cannot prevent the weight matrices in DNNs from being low-rank or redundant.
To further support our arguments, we provide the experimental results with different ridge parameters on a real-world dataset CUB as shown in Figure \ref{fig:ridge_in_cub}.
One can see that the ridge regularization even damages the performance of DCCA and also leads to an increase in the internal correlation within the feature and the correlation between the feature and noise.
In our NR-DCCA, we set the ridge parameter to zero. We conjecture the reason is that the large ridge parameter could make the neural network even “lazier” to actively project the data into a better feature space, as the full-rank property of features and covariance matrix are already guaranteed.

\subsubsection{Hyper-parameter $\alpha$ of NR-DCCA}
\begin{figure}[h]
    \centering
    \includegraphics[width=1\linewidth]{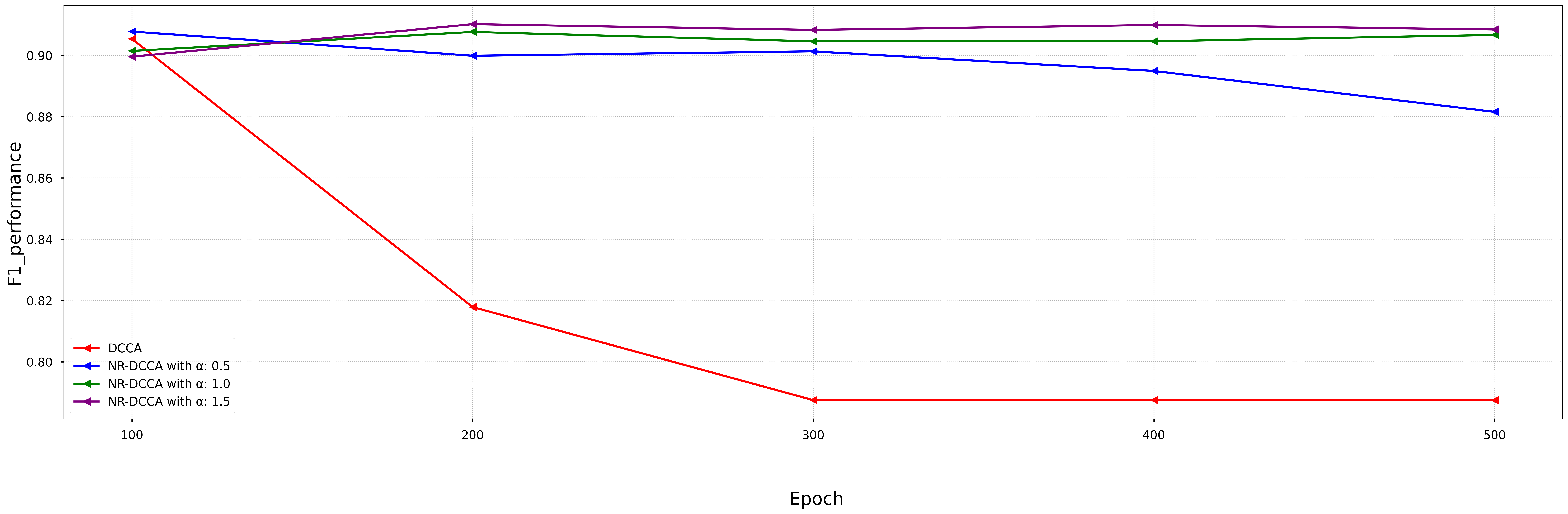}
    \caption
    {The effects of hyper-parameter $\alpha$ of NR-DCCA in the CUB dataset.}
    \label{fig:alpha}
\end{figure}

The choice of the hyper-parameter $\alpha$ is essential in NR-DCCA. Different from the conventional hyper-parameter tuning procedures, the determination of $\alpha$ is simpler, as we can search for the smallest $\alpha$ that can prevent the model collapse, and the model collapse can be directly observed on the validation data. Specifically, we increase the $\alpha$ adaptively until the model collapse issue is tackled, i.e., the correlation with noise will not increase or the performance of DCCA will not drop with increasing training epochs, then the optimal $\alpha$ is found.
To further illustrate the influence of $\alpha$ in NR-DCCA, we present performance curves of NR-DCCA in CUB under different $\alpha$. As shown in Figure~\ref{fig:alpha}, if $\alpha$ is too large, the convergence of the training becomes slow; if $\alpha$ is too small, model collapse remains. Additionally, one can see the NR-DCCA outperforms DCCA robustly with a wide range of $\alpha$.

\subsection{Effects of the Distribution of Noise}
\label{appenidx:Effects of the Distribution of Noise}

From our theoretical analysis, the most important feature of noise in NR is that the sampled noise matrix is a full-rank matrix. Therefore, continuous distributions such as the uniform distribution can also be applied to NR, which demonstrates the robustness of the proposed NR method. We compare NR-DCCA with different noise distributions on synthetic datasets, and both noises are effective in suppressing model collapse as shown in Table ~\ref{tab:noise_effects}.

\begin{table}[ht]
\centering
\caption{Effect of Noise on DCCA and NR-DCCA}
\label{tab:noise_effects}
\begin{tabular}{@{}cccc@{}}
\toprule
Epoch & DCCA & NR-DCCA (Gaussian Noise) & NR-DCCA (Uniform Noise) \\ \midrule
100   & $0.284 \pm 0.012$ & $0.295 \pm 0.005$ & $0.291 \pm 0.004$ \\
800   & $0.137 \pm 0.028$ & $0.313 \pm 0.004$ & $0.313 \pm 0.005$ \\
1200  & $0.106 \pm 0.027$ & $0.312 \pm 0.005$ & $0.316 \pm 0.005$ \\ \bottomrule
\end{tabular}
\end{table}

\subsection{Effects of depths of Encoders}

\label{appendix:complexity of encoders}
In this section, we test the effects of depths of encoders (i.e. MLPs) on model collapse and NR.
Specifically, we increase the depth of MLPs to observe the variation in the performance of DCCA and NR-DCCA on synthetic datasets. As shown in Table ~\ref{tab: Effects of Complexity of Encoders}, The increase in network depth results in a faster decline in DCCA performance, while NR-DCCA still maintains a stable performance.

\begin{table}[htbp]
\centering
\caption{Performance comparison of DCCA and NR-DCCA across different network depths.}
\label{tab: Effects of Complexity of Encoders}
\resizebox{\linewidth}{!}{
\begin{tabular}{@{}ccccccc@{}}
\toprule
Epoch/R2 & \multicolumn{2}{c}{\textbf{1 hidden layer}} & \multicolumn{2}{c}{\textbf{2 hidden layers}} & \multicolumn{2}{c}{\textbf{3 hidden layers}} \\ 
\cmidrule(r){2-3} \cmidrule(r){4-5} \cmidrule(r){6-7}
 & DCCA & NR-DCCA & DCCA & NR-DCCA & DCCA & NR-DCCA \\  % 这里去掉了一个多余的 `{`
\midrule
100  & $0.284 \pm 0.012$ & $\textbf{0.295} \pm 0.005$ & $0.161 \pm 0.013$ & $\textbf{0.304} \pm 0.006$ & $0.071 \pm 0.084$ & $\textbf{0.299} \pm 0.010$ \\
800  & $0.137 \pm 0.028$ & $\textbf{0.313} \pm 0.004$ & $-0.072 \pm 0.071$ & $\textbf{0.307} \pm 0.005$ & $-0.975 \pm 0.442$ & $\textbf{0.309} \pm 0.005$ \\
1200 & $0.106 \pm 0.027$ & $\textbf{0.312} \pm 0.005$ & $-0.154 \pm 0.127$ & $\textbf{0.303} \pm 0.006$ & $-1.412 \pm 0.545$ & $\textbf{0.308} \pm 0.006$ \\
\bottomrule
\end{tabular}
}
\end{table}

\subsection{Effects of NR Loss on Filter Redundancy}
\label{appendix: nesum}
\begin{figure}[h]
    \centering
    \includegraphics[width=0.8\linewidth]{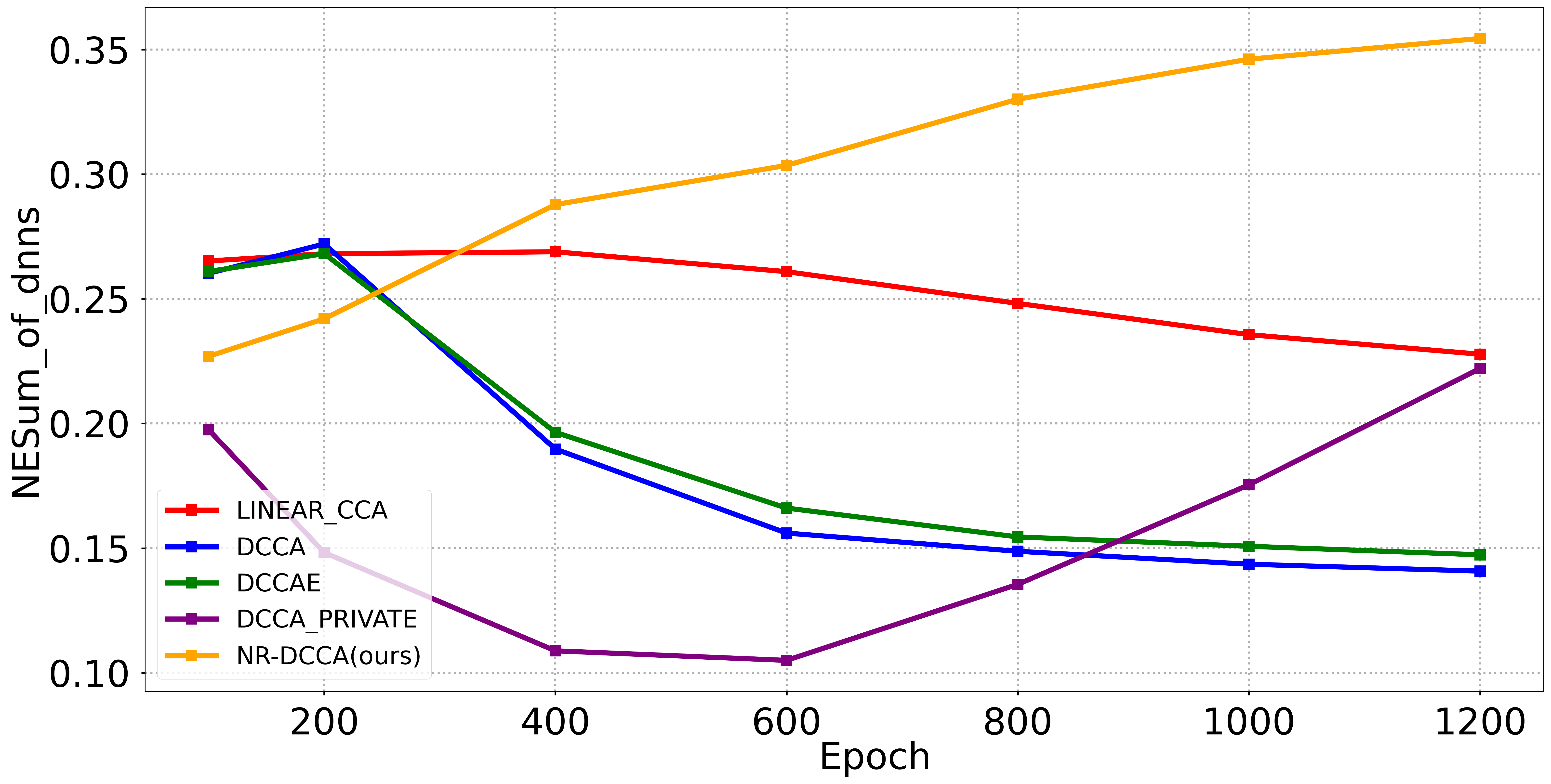}
    \caption
    {Average NESum across all weights within the trained encoders.}
    \label{fig:nesum}
\end{figure}

Extensive research has established a significant correlation between the redundancy of neurons or filters and the compromised generalization capabilities of neural networks, indicating a propensity for overfitting~\citep{wang2020mma, morcos2018importance, zhu2018improving}. 
Considering the fully connected layer with 1024 units in a MLP network as a paradigm, the initial layer's weights, denoted by \(W \in \mathbb{R}^{1024 \times 3 \times 28 \times 28}\), can be interpreted as 1024 discriminative filters. These filters operate on images with 3 channels, each of size \(28 \times 28\), with every filter representing a vector in a \(3 \times 28 \times 28\) dimensional space. Subsequently, a similarity matrix \(S\) is constructed, wherein each element \(S_{ij}\) quantifies the cosine similarity between the \(i^{th}\) and \(j^{th}\) filters, with higher values indicating greater redundancy.
To further assess filter redundancy in \(W\), we employ NESum, a metric designed for evaluating redundancy and whiten degrees of features~\citep{zhang2023orthoreg}.
\begin{definition}[NESum of Weight]
\label{def:NESum_of_Weight}
Given a weight matrix \(W \in \mathbb{R}^{output \times input}\) with an accompanying output-wise similarity matrix \(S \in \mathbb{R}^{output \times output}\) and eigenvalues \(\{\lambda_i\}_{i=1}^{output}\) sorted in descending order, the normalized eigenvalue sum is defined as follows:
\[
NESum(W) = \frac{1}{output} \sum_{i=1}^{output} \frac{\lambda_i}{\lambda_1}
\]
\end{definition}
In Figure~\ref{fig:nesum}, we present the evolution of the average NESum across all weights within the trained encoders. Notably, we observe a sustained increase in NESum exclusively in NR-DCCA throughout prolonged training epochs. This phenomenon underscores the efficacy of the loss of NR in reducing filter redundancy, crucially preventing low-rank solutions.

\subsection{Visualization of Learned Representations}
\label{appendix:visulization}
\begin{figure*}[h]
	\centering
	\subfigure[CONCAT]{
		\begin{minipage}[t]{0.2\linewidth}
			\centering
			\includegraphics[width=1.2in]{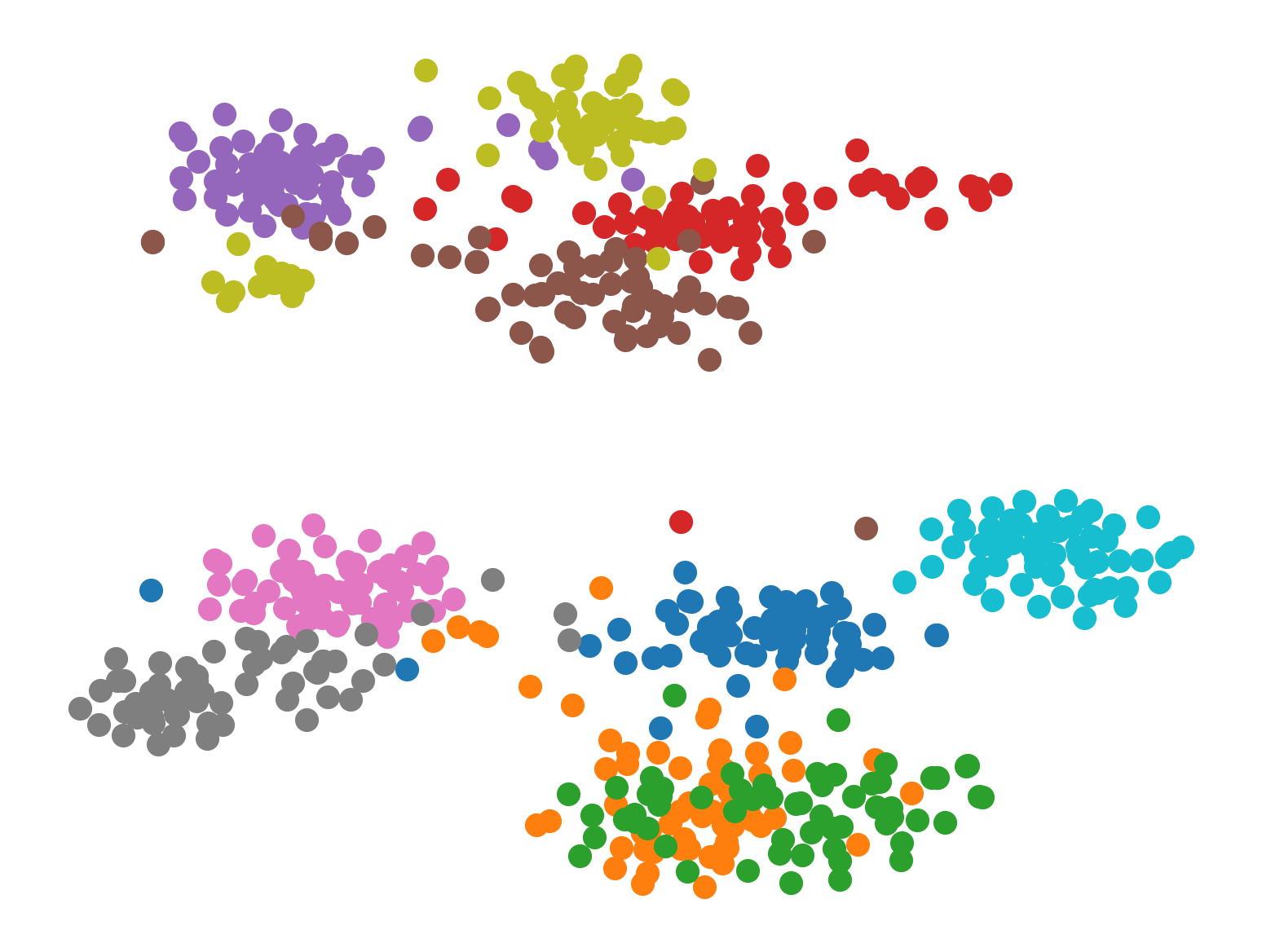}
		\end{minipage}
	}%
	\subfigure[CCA]{
		\begin{minipage}[t]{0.2\linewidth}
			\centering
			\includegraphics[width=1.2in]{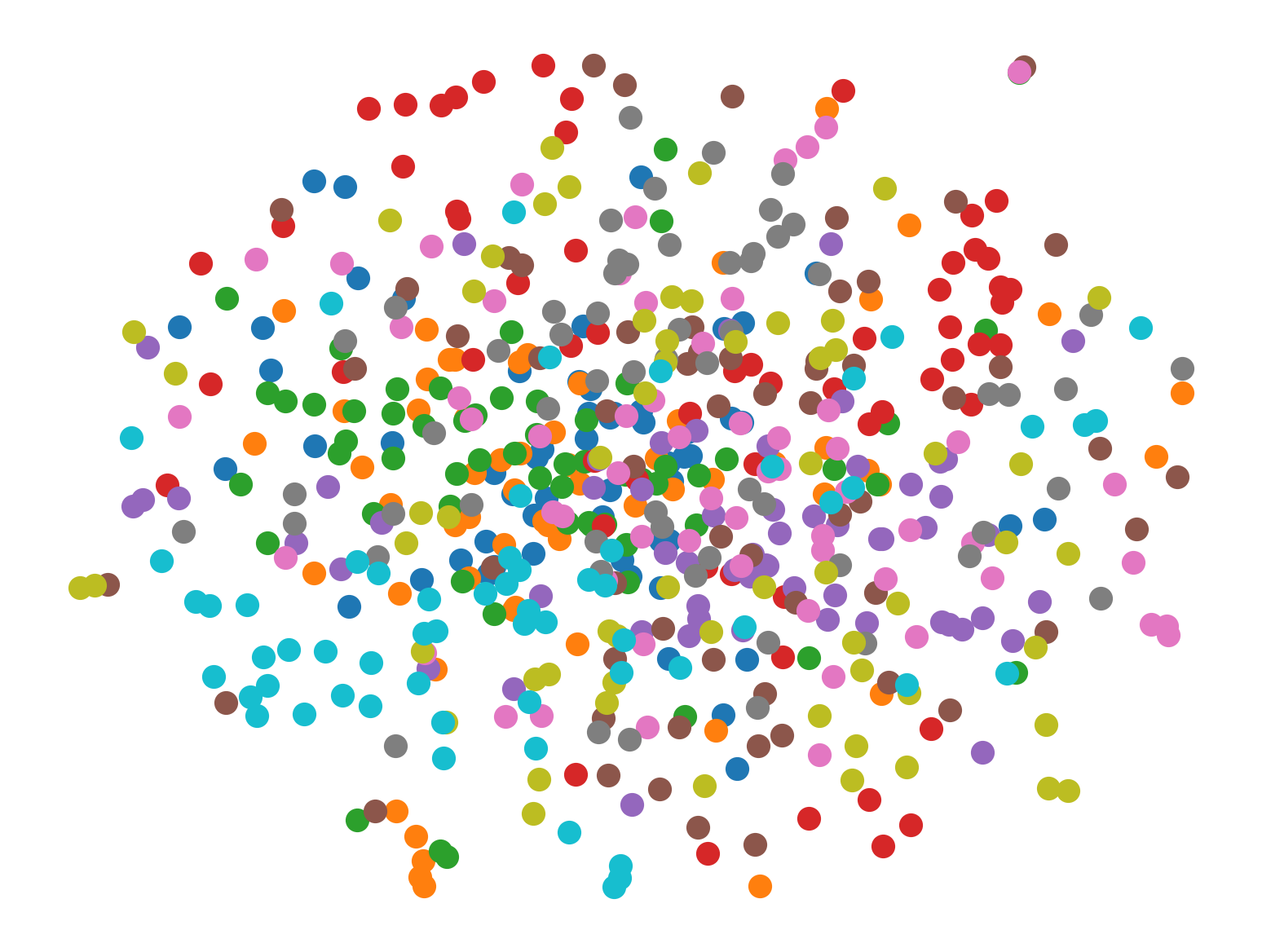}
		\end{minipage}
	}%
	\subfigure[MVTCAE]{
		\begin{minipage}[t]{0.2\linewidth}
			\centering
			\includegraphics[width=1.2in]{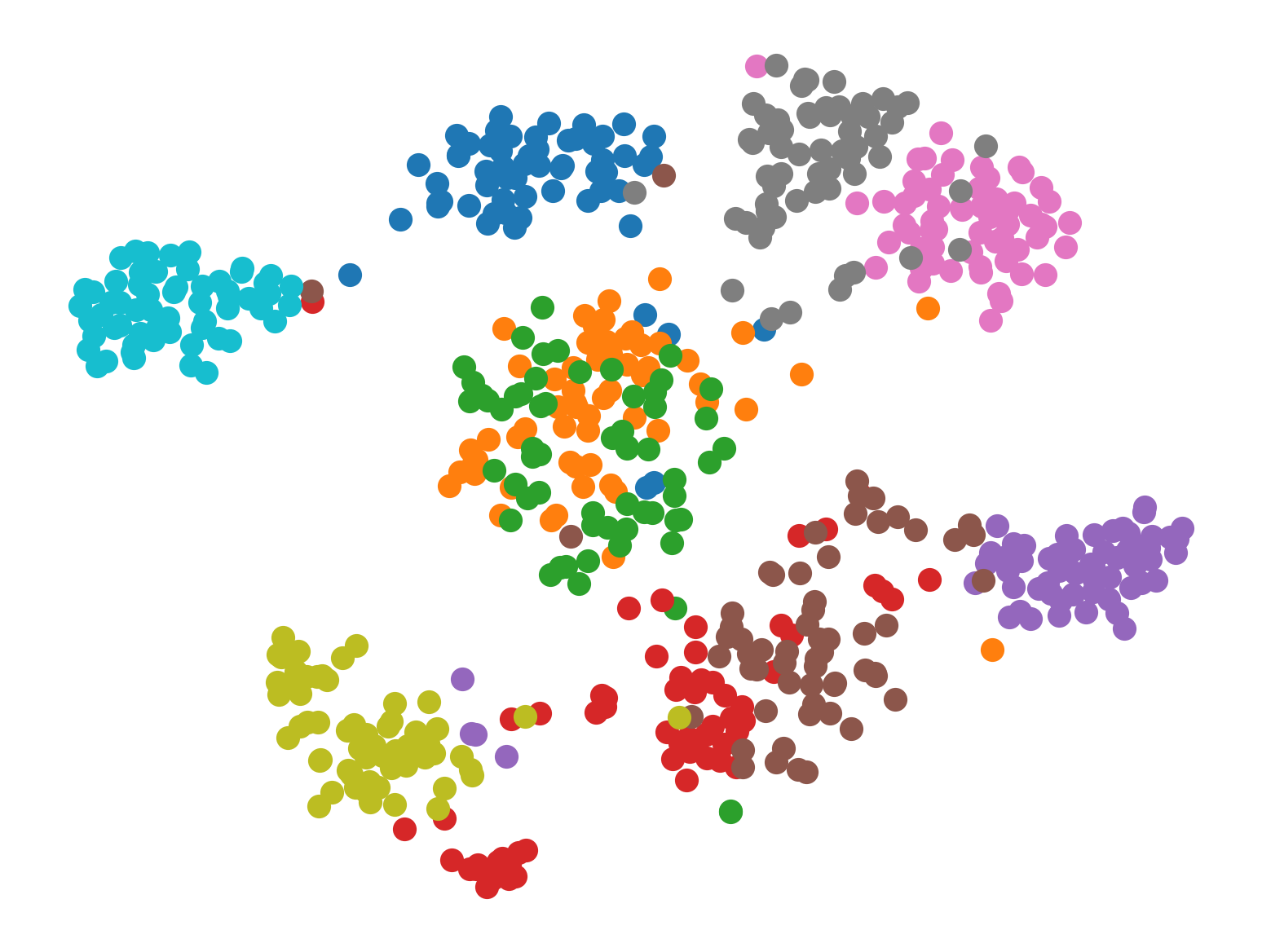}
		\end{minipage}
	}%
 
	\subfigure[DCCA]{
		\begin{minipage}[t]{0.2\linewidth}
			\centering
			\includegraphics[width=1.2in]{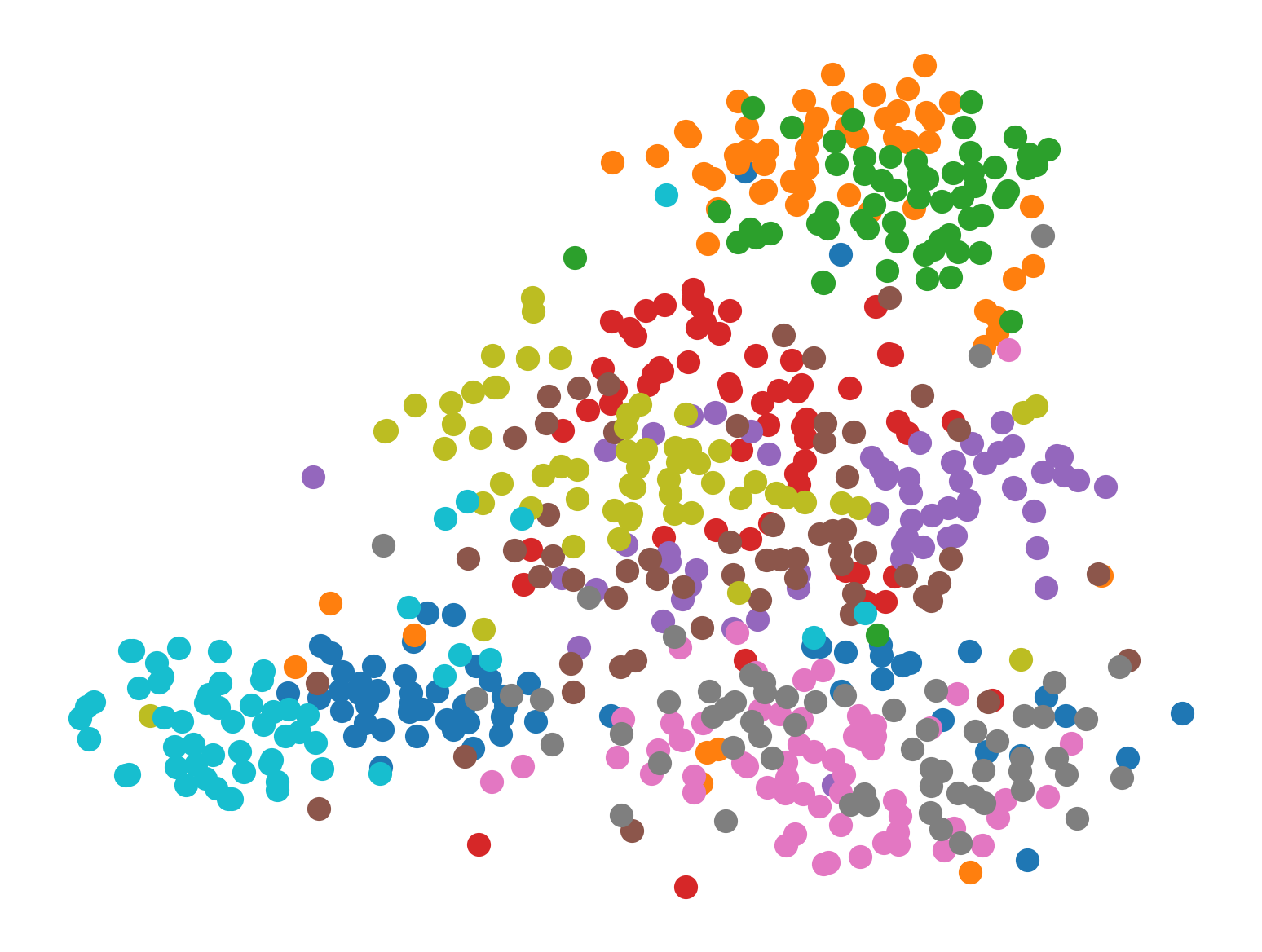}
		\end{minipage}
	}%
        \subfigure[DCCAE]{
		\begin{minipage}[t]{0.2\linewidth}
			\centering
			\includegraphics[width=1.2in]{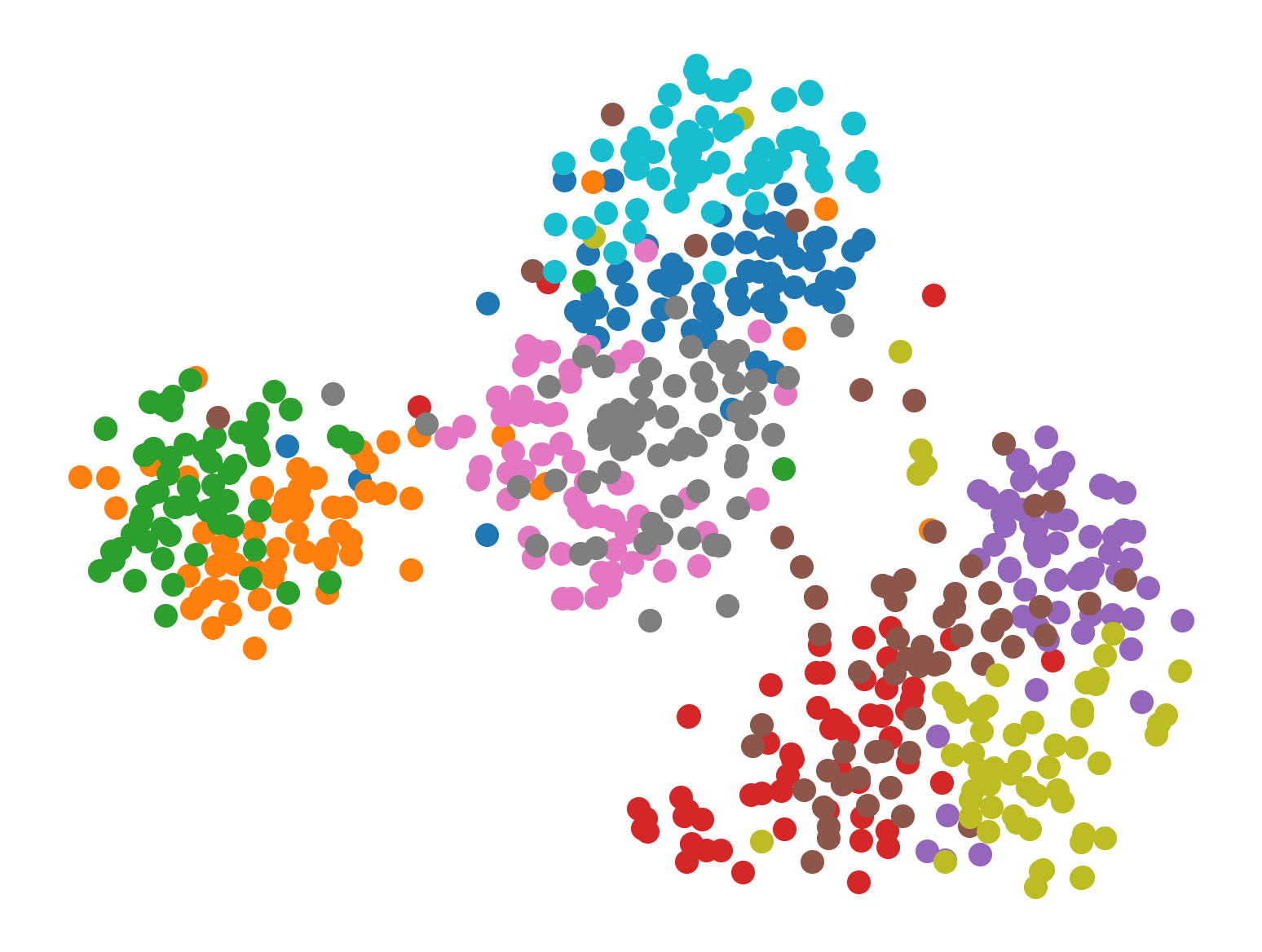}
		\end{minipage}
	}%
        \subfigure[NR-DCCA]{
		\begin{minipage}[t]{0.2\linewidth}
			\centering
			\includegraphics[width=1.2in]{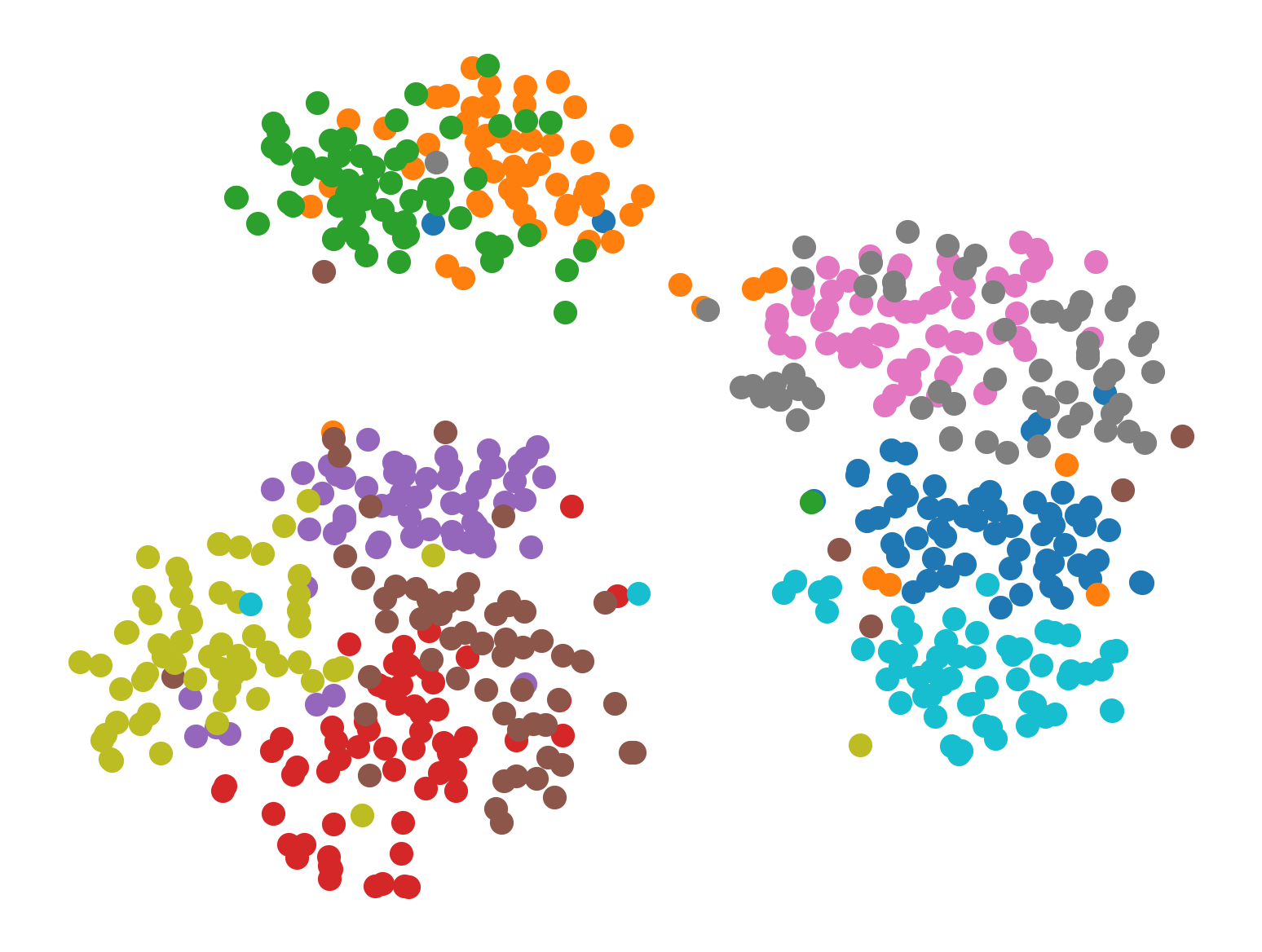}
		\end{minipage}
	}%
        \subfigure[DCCA\_PRIVATE]{
		\begin{minipage}[t]{0.2\linewidth}
			\centering
			\includegraphics[width=1.2in]{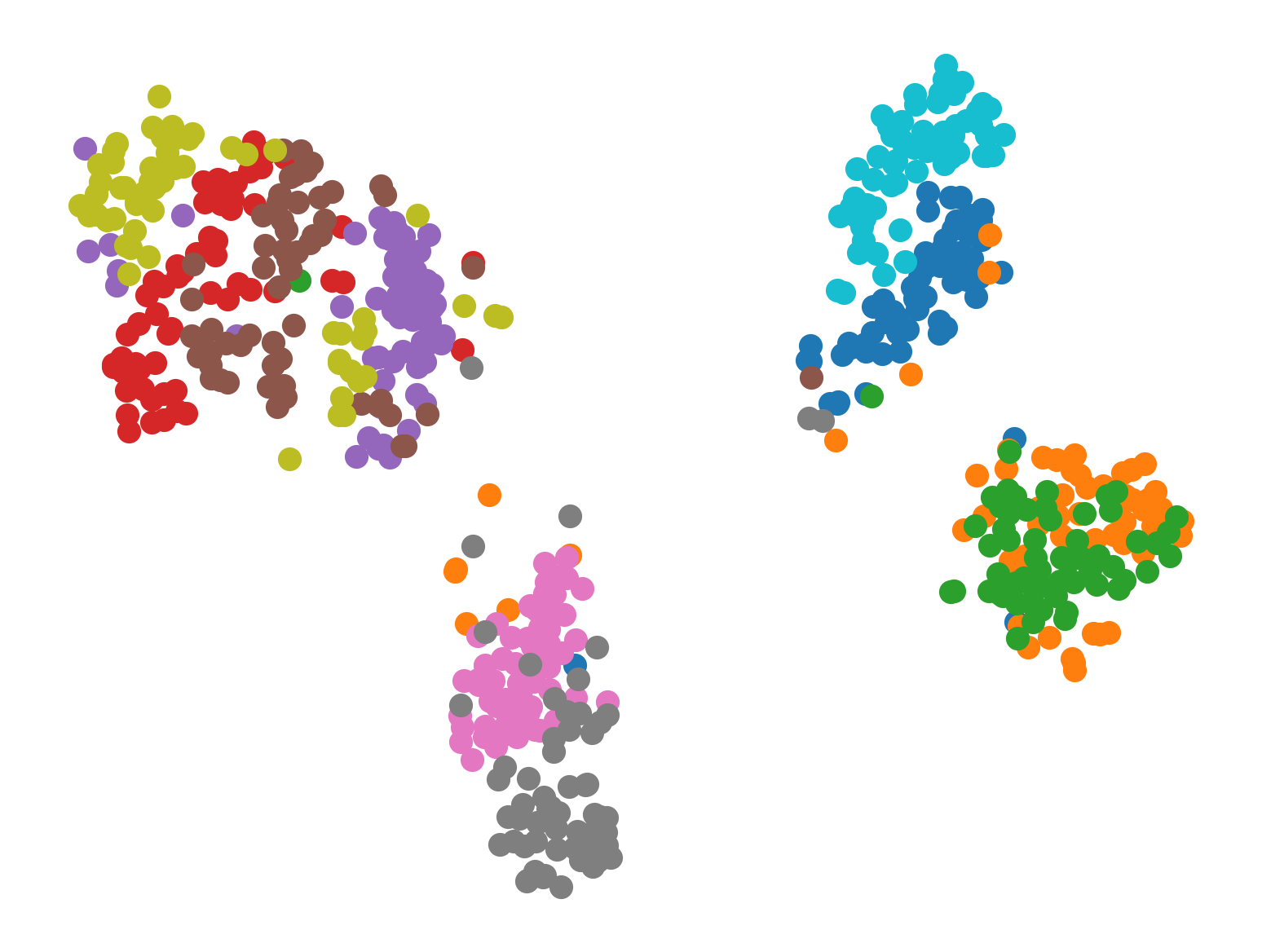}
		\end{minipage}
	}%

        \subfigure[DGCCA]{
		\begin{minipage}[t]{0.2\linewidth}
			\centering
			\includegraphics[width=1.2in]{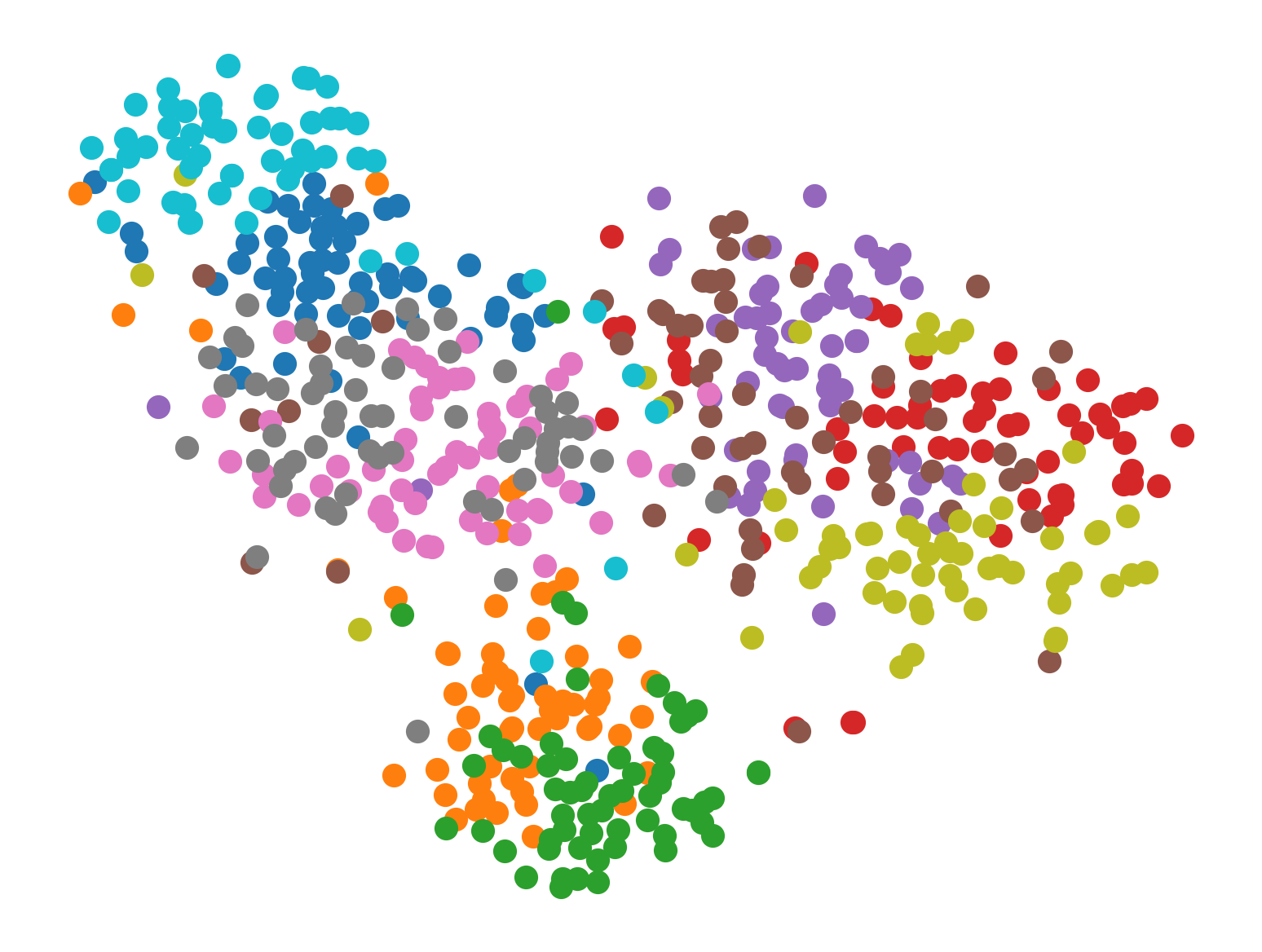}
		\end{minipage}
	}%
        \subfigure[DGCCAE]{
		\begin{minipage}[t]{0.2\linewidth}
			\centering
			\includegraphics[width=1.2in]{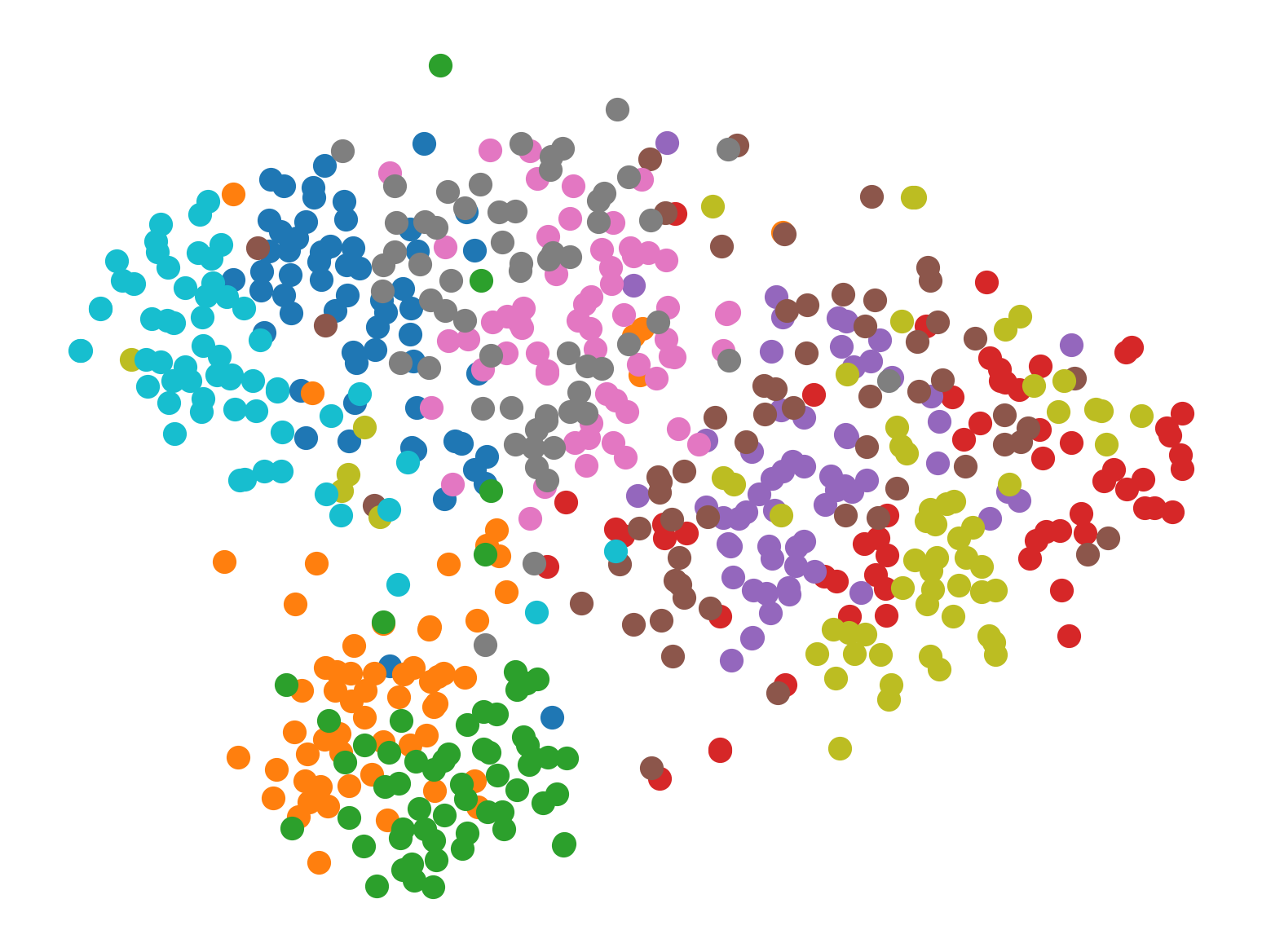}
		\end{minipage}
	}%
        \subfigure[NR-DGCCA]{
		\begin{minipage}[t]{0.2\linewidth}
			\centering
			\includegraphics[width=1.2in]{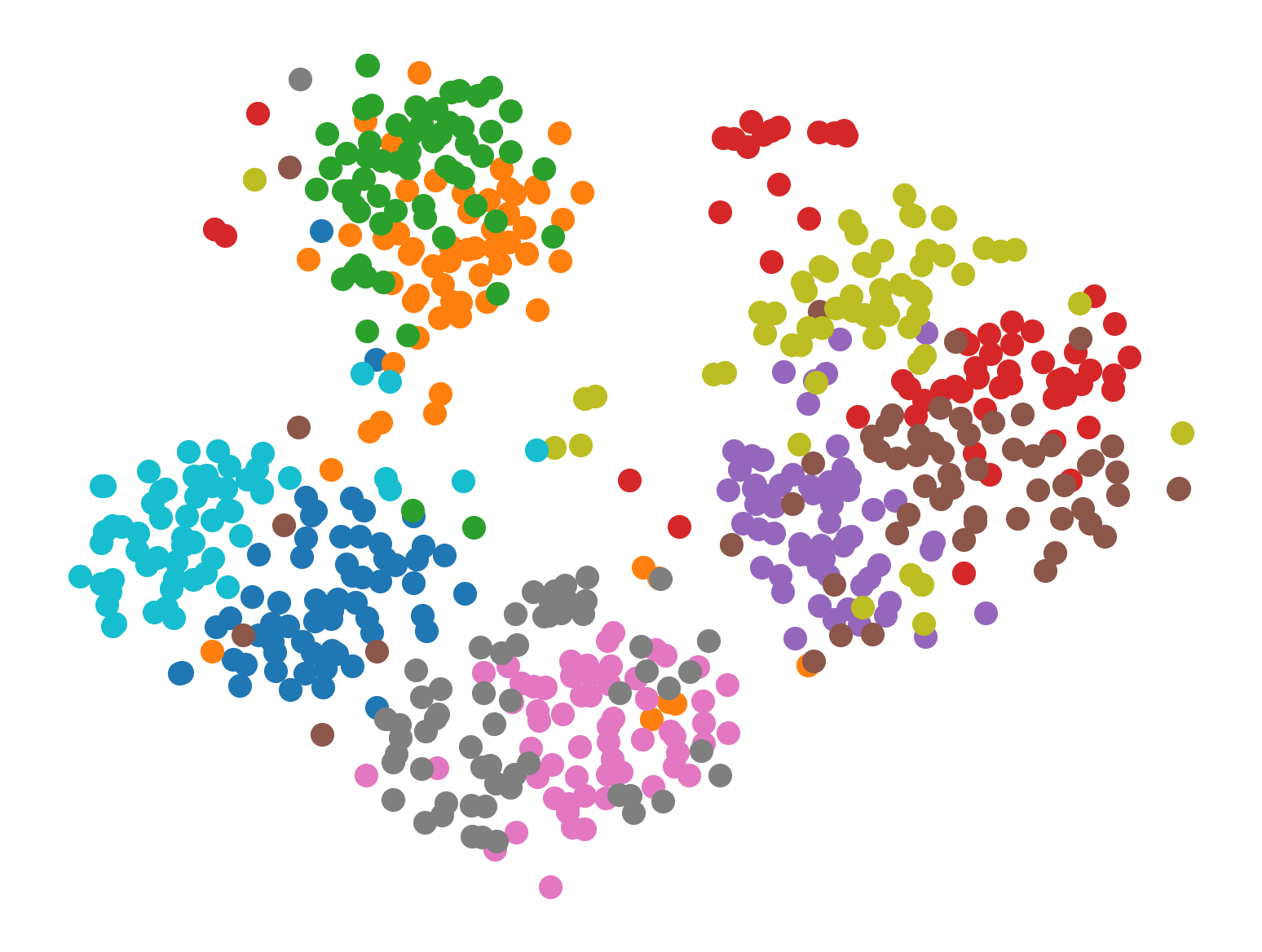}
		\end{minipage}
	}%
        \subfigure[DGCCA\_PRIVATE]{
		\begin{minipage}[t]{0.2\linewidth}
			\centering
			\includegraphics[width=1.2in]{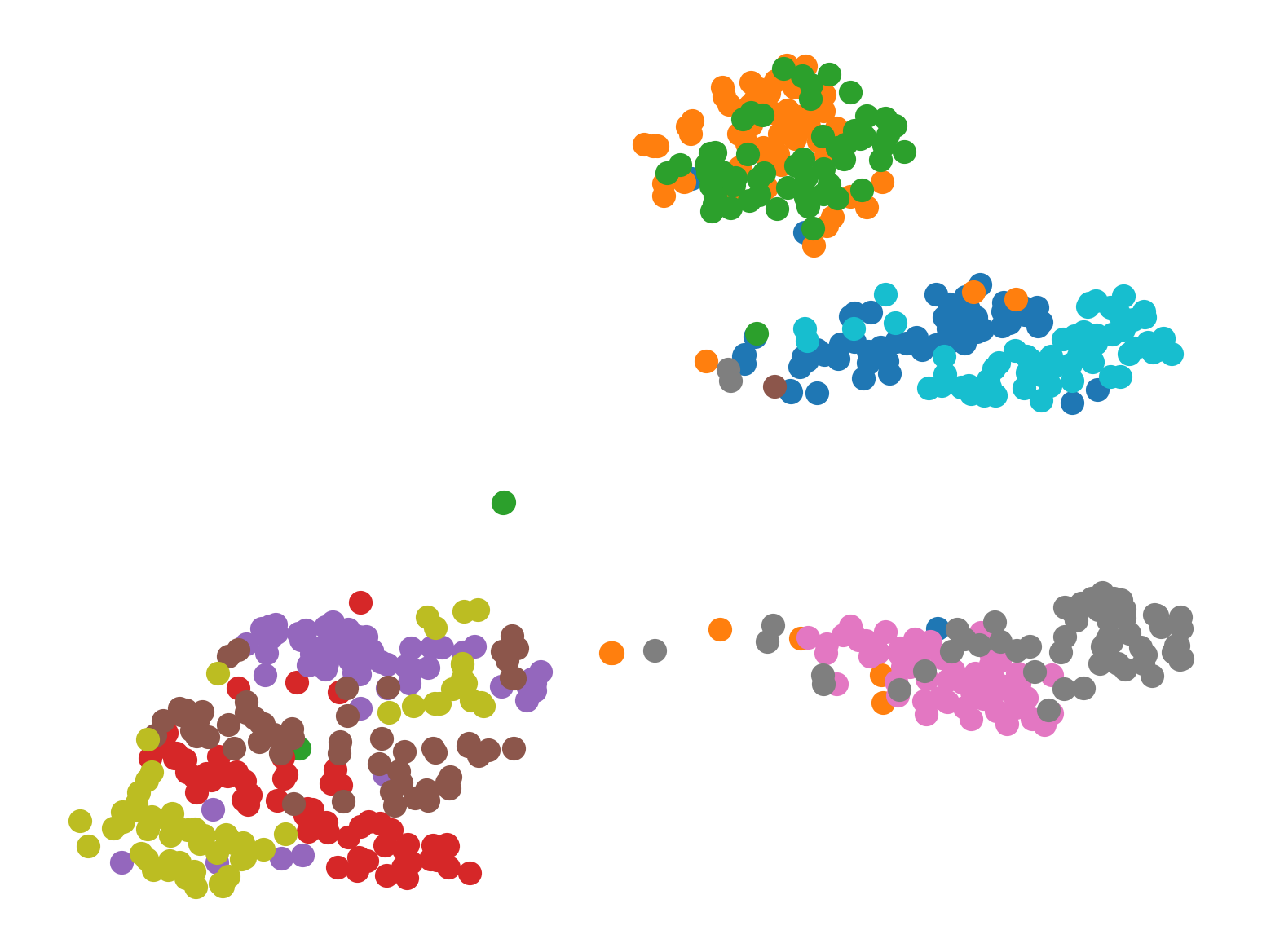}
		\end{minipage}
	}%
        
	\centering
	\caption{Visualization of the learned representations with t-SNE in the \textbf{CUB} dataset.}
	\label{fig:visualization}
\end{figure*}

To further demonstrate the effectiveness of our method, we employ 2D-tSNE visualization to depict the learned representations of the CUB dataset (test set) under different methods. Each data point is colored based on its corresponding class, as illustrated in Figure \ref{fig:visualization}. 
There are a total of 10 categories, with 60 data points in each category. A reasonable distribution of learned representations entails that data points belonging to the same class are grouped in the same cluster, which is distinguishable from clusters representing other classes. Additionally, within each cluster, the data points should exhibit an appropriate level of dispersion, indicating that the data points within the same class can be differentiated rather than collapsing into a single point. This dispersion is indicative of the preservation of as many distinctive features of the data as possible.

From Figure.~\ref{fig:visualization}, we can observe that CCA, DCCA / DGCCA have all confused the data from different categories. Specifically, CCA  completely scatters the data points as it cannot handle non-linear relationships. By incorporating autoencoders, DCCAE / DGCCAE and DCCA\_PRIVATE / DGCCA\_PRIVATE have partially separated the data; however, they have not fully separated the green and orange categories. NR-DCCA / NR-DGCCA is the only method that successfully separates all categories.

It is worth noting that our approach not only separates the data into different clusters but also maintains dispersion within each cluster. Unlike DCCA\_PRIVATE / DGCCA\_PRIVATE, where the data points within a cluster form a strip-like distribution, our method ensures that the data points within each cluster remain appropriately scattered.

\clearpage
\subsection{DGCCA and NR-DGCCA}
\label{appendix:dgcca}

\begin{figure*}[h]
	\centering
	\subfigure[Performance]{
		\begin{minipage}[t]{0.8\linewidth}
			\centering
			\includegraphics[width=\linewidth]{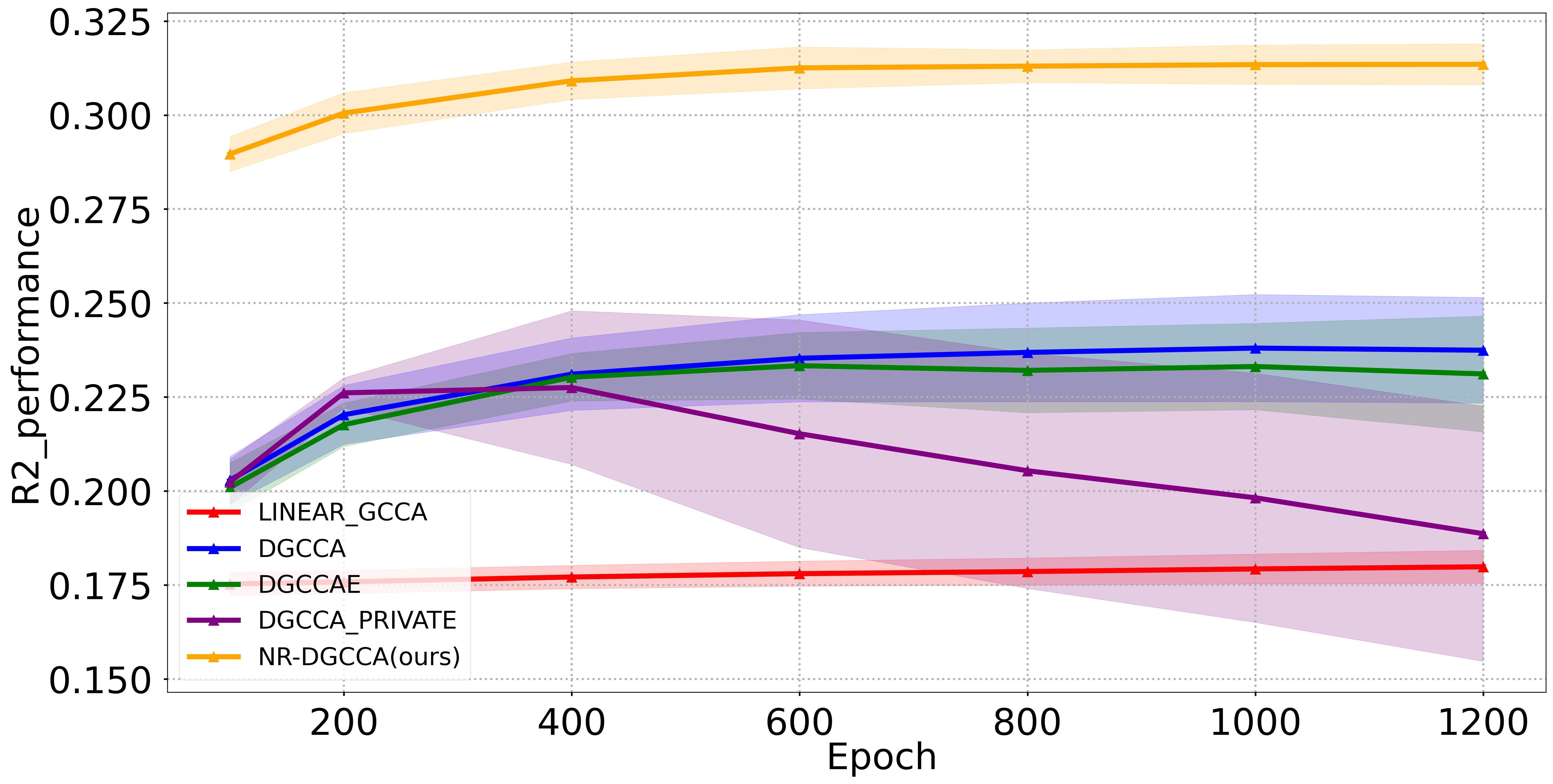}
		\end{minipage}
	}%

        \subfigure[Correlation]{
		\begin{minipage}[t]{0.5\linewidth}
			\centering
			\includegraphics[width=\linewidth]{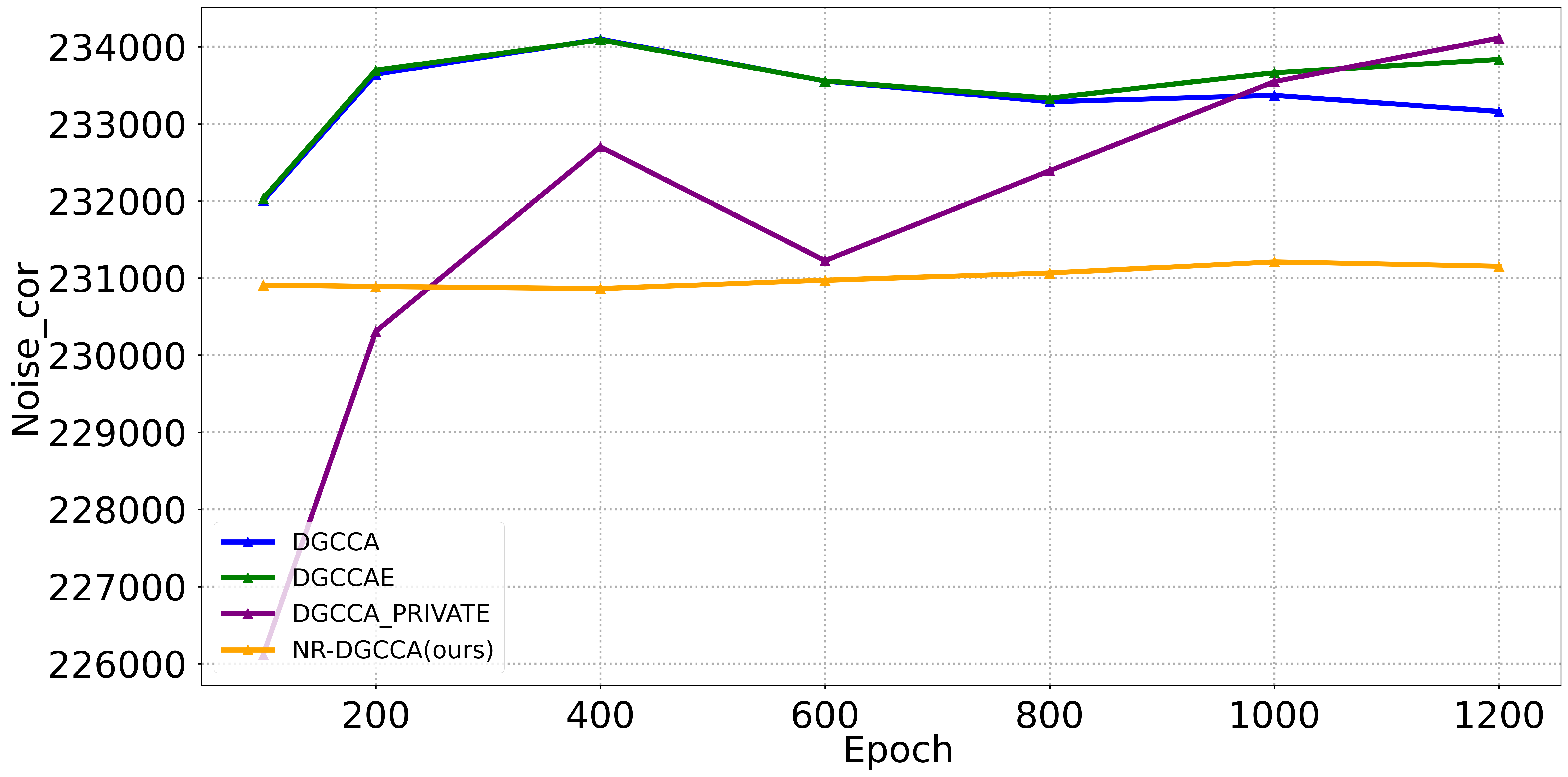}
		\end{minipage}
	}%
        \subfigure[NESum of weight matrices]{
		\begin{minipage}[t]{0.5\linewidth}
			\centering
			\includegraphics[width=\linewidth]{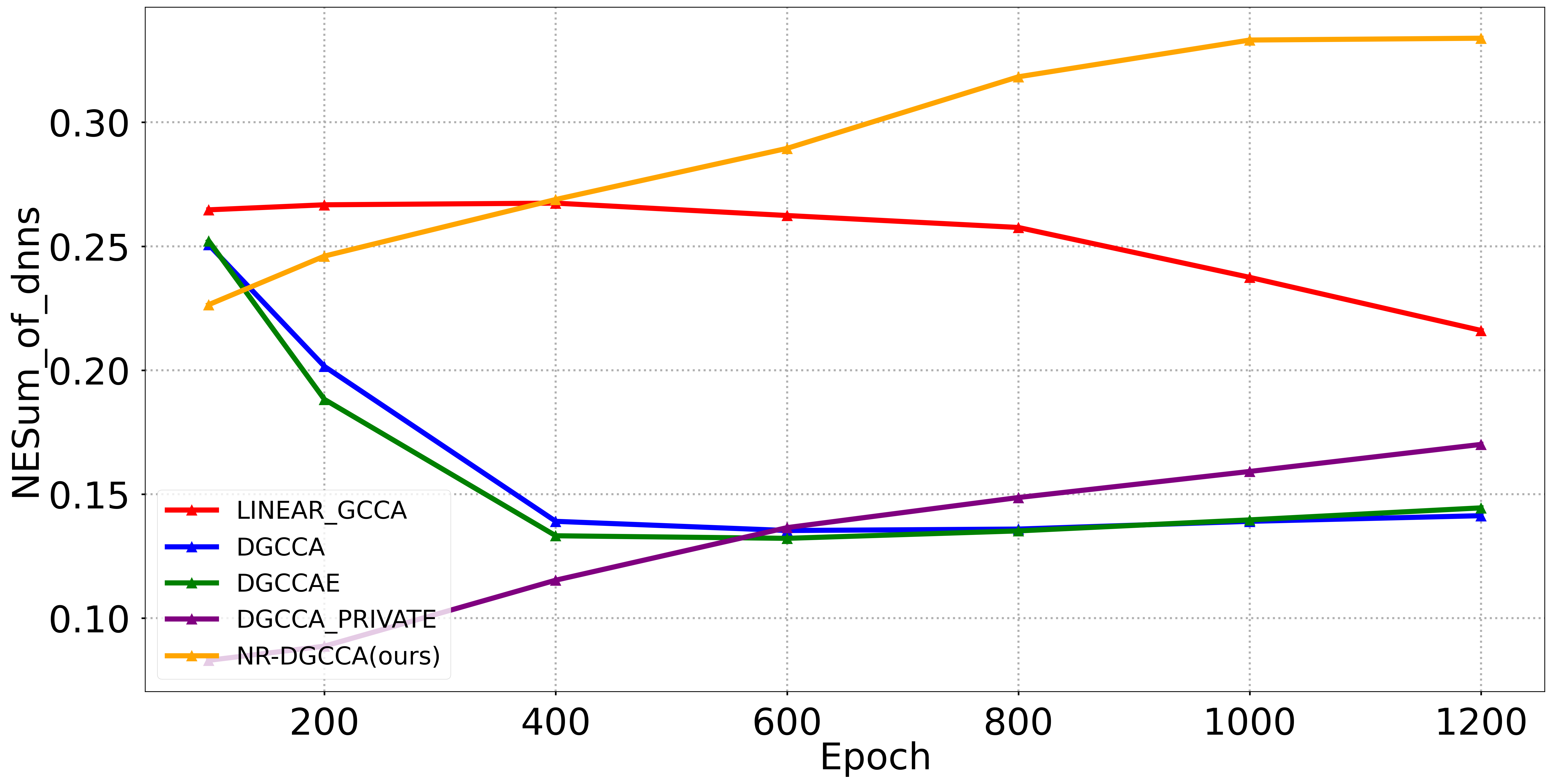}
		\end{minipage}
	}%

 \subfigure[Reconstruction]{
		\begin{minipage}[t]{0.5\linewidth}
			\centering
			\includegraphics[width=\linewidth]{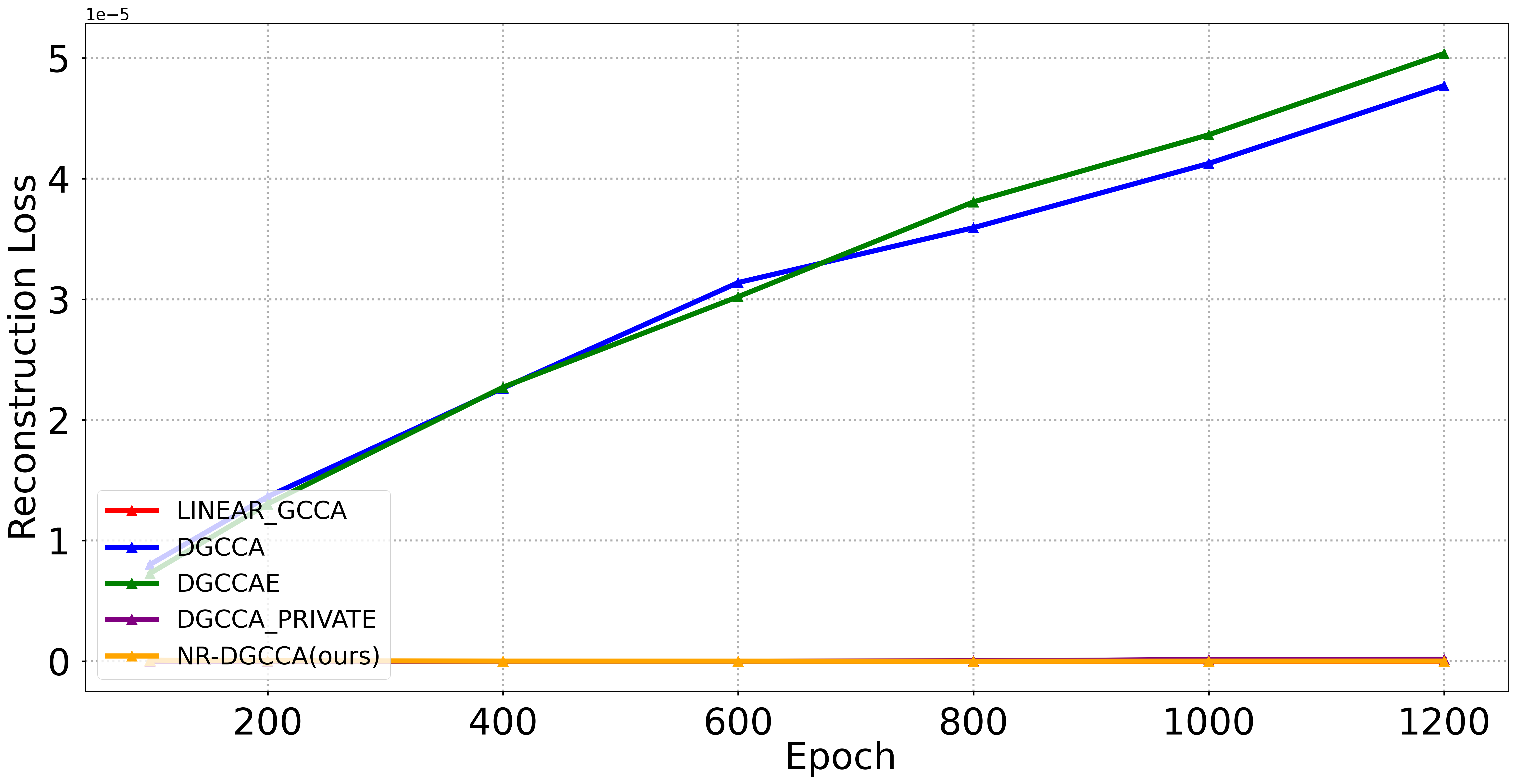}
		\end{minipage}
	}%
    \subfigure[Denoisng]{
		\begin{minipage}[t]{0.5\linewidth}
			\centering
			\includegraphics[width=\linewidth]{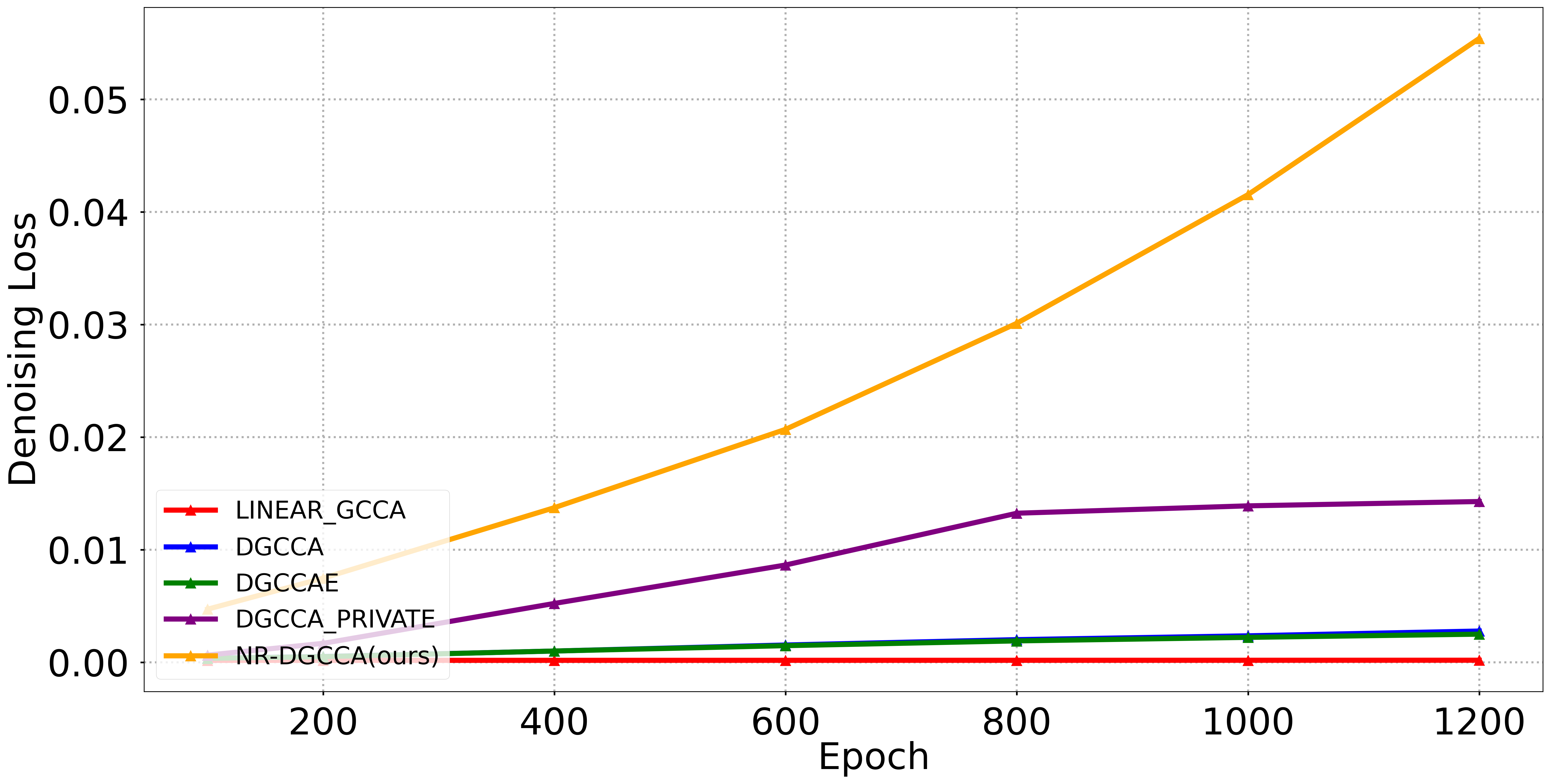}
		\end{minipage}
	}%
	\centering
	\caption{(a) Mean and standard deviation of the (D)GCCA-based method performance across synthetic datasets in different training epochs.(b) The mean correlation between noise and real data after transformation varies with epochs. (c) Average NESum across all weights within the trained encoders.
   (d,e) The mean of Reconstruction and Denoising Loss on the test set.}
 \label{fig: mean and std full Gcca}
\end{figure*}

\begin{figure}[h]
    \centering
    \includegraphics[width=0.99\linewidth]{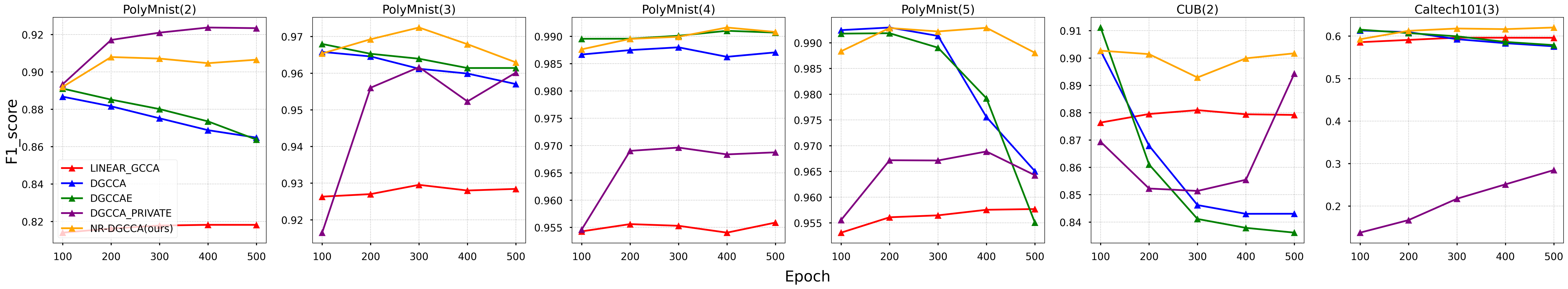}
    \caption
    {Performance of DGCCA-based methods in real-world datasets. 
    % For MVTCAE, we plot its best epoch as the result for each epoch. W
    Each column represents the performance on a specific dataset.
    The number of views in the dataset is denoted in the parentheses next to the dataset name.}
    \label{fig: real_world_gcca}
\end{figure}

This section presents the experimental results for DGCCA and NR-DGCCA, which supplement the results of GCCA and NR-DCCA presented in the main paper. In general, DGCCA is a variant of DCCA, and hence the proposed noise regularization approach can also be applied. We repeat the experiments in Figures~\ref{fig: mean and std full cca},and~\ref{fig: real_world_cca}, and hence we have the results for DGCCA in Figure~\ref{fig: mean and std full Gcca}, and~\ref{fig: real_world_gcca}. One can see that the proposed noise regularization approach can also help DGCCA prevent model collapse, proving its generalizability.

% \begin{figure}[h]
%     \centering
%     \includegraphics[width=0.99\linewidth]{figures/Syn_rate_r2_GCCA.png}
%     \caption
%     {Performance of DGCCA-based methods with respect to different common rates during the training. Each column represents the testing accuracy of the method at a specific training epoch.}
%     \label{fig: syn_full_GCCA}
% \end{figure}

\clearpage

% \subsection{Effects of depths of Encoders}
% \label{appendix:complexity of encoders}

% \clearpage

\subsection{Additional Experimental Results}
\label{appendix:additional results}

Table~\ref{table:syn} and~\ref{table:real} present the model performance of various MVRL methods in synthetic and real-world datasets, and both tables correspond to the final epoch of the results presented in Figure~\ref{fig: mean and std full cca}and~\ref{fig: real_world_cca}.
It should be noted that the values in Table \ref{table:syn} represent the mean and standard deviation of the methods across different tasks, indicating their performance and variability.

\begin{table}[h]
\caption{Performance in synthetic datasets.} % title of Table
\centering % used for centering table
\resizebox{\linewidth}{!}{
\begin{tabular}{c|c|c|c|c|c|c}
\hline
R2/Common Rate      & 0\%   & 20\%   & 40\% & 60\% & 80\% & 100\% \\ \hline
CONCAT  & 0.253\(\pm\)0.038 & 0.255\(\pm\)0.039 & 0.250\(\pm\)0.040 & 0.254\(\pm\)0.040 & 0.256 \(\pm\)0.042 & 0.264 \(\pm\) 0.033        \\
% \textsc{MVTCAE}        & -0.001 & -0.001 & -0.001         \\ \hline
Linear\_CCA & 0.179\(\pm\)0.030 & 0.184\(\pm\)0.035 & 0.172\(\pm\)0.033 & 0.182\(\pm\)0.034 & 0.182\(\pm\)0.034 & 0.188\(\pm\)0.031 \\
Linear\_GCCA & 0.177\(\pm\)0.030 & 0.184\(\pm\)0.036 & 0.171\(\pm\)0.033 & 0.182\(\pm\)0.034 & 0.182\(\pm\)0.033 & 0.182\(\pm\)0.031 \\
KCCA & 0.243\(\pm\)0.047 & 0.261\(\pm\)0.046 & 0.260\(\pm\)0.043 & 0.272\(\pm\)0.045 & 0.276\(\pm\)0.049 & 0.288\(\pm\)0.038 \\

PRCCA & 0.212\(\pm\)0.053 & 0.249\(\pm\)0.046 & 0.216\(\pm\)0.055 & 0.267\(\pm\)0.046 & 0.256\(\pm\)0.052 & 0.284\(\pm\)0.039 \\
\hline
MVTCAE & 0.065\(\pm\)0.015 & 0.071\(\pm\)0.016 & 0.067\(\pm\)0.016 & 0.069\(\pm\)0.016 & 0.071\(\pm\)0.016 & 0.069\(\pm\)0.015 \\
\hline
DCCA          & 0.053\(\pm\)0.044 & 0.094\(\pm\)0.046  & 0.123\(\pm\)0.047 &0.107\(\pm\)0.046 &0.125\(\pm\)0.052 &0.133\(\pm\)0.044       \\
DCCAE         & 0.063\(\pm\)0.044 & 0.090\(\pm\)0.039 & 0.126\(\pm\)0.047 & 0.104\(\pm\)0.045 &0.098\(\pm\)0.060 &0.139 \(\pm\)0.041   \\ 
DCCA\_PRIVATE & 0.264\(\pm\)0.039 & 0.171\(\pm\)0.040 & 0.176\(\pm\)0.042 & 0.168\(\pm\)0.039 & 0.172\(\pm\)0.041 & 0.181\(\pm\)0.035    \\ 
DCCA\_GHA & 0.251\(\pm\)0.049 & 0.249\(\pm\)0.046 & 0.243\(\pm\)0.047 & 0.252\(\pm\)0.052 & 0.268\(\pm\)0.053 & 0.275\(\pm\)0.046    \\ 
DCCA\_EY & 0.195\(\pm\)0.050 & 0.205\(\pm\)0.044 & 0.220\(\pm\)0.041 & 0.214\(\pm\)0.046 & 0.215\(\pm\)0.046 & 0.234\(\pm\)0.041    \\ 

NR-DCCA (ours)  & \textbf{0.311}\(\pm\)0.043 & \textbf{0.314}\(\pm\)0.046 & \textbf{0.306}\(\pm\)0.043 & \textbf{0.309}\(\pm\)0.042 & \textbf{0.313}\(\pm\)0.049 & \textbf{0.322}\(\pm\)0.040 \\ \hline

DGCCA     & 0.231\(\pm\)0.042 & 0.244\(\pm\)0.040 & 0.242\(\pm\)0.040 & 0.211\(\pm\)0.039 & 0.240\(\pm\)0.040 & 0.256\(\pm\)0.037        \\
DGCCAE    & 0.209\(\pm\)0.039 & 0.235\(\pm\)0.042 & 0.240\(\pm\)0.040 & 0.214\(\pm\)0.038 & 0.235\(\pm\)0.041 & 0.254\(\pm\)0.036      \\ 
DGCCA\_PRIVATE & 0.264\(\pm\)0.039 & 0.171\(\pm\)0.040 & 0.176\(\pm\)0.042 & 0.168\(\pm\)0.039 & 0.172\(\pm\)0.042 & 0.181\(\pm\)0.035 \\ 
NR-DGCCA (ours)  & \textbf{0.314}\(\pm\)0.044 & \textbf{0.317}\(\pm\)0.045 & \textbf{0.305}\(\pm\)0.043 & \textbf{0.308}\(\pm\)0.044 & \textbf{0.315}\(\pm\)0.049 & \textbf{0.322}\(\pm\)0.040  \\ \hline
\end{tabular}
}
\label{table:syn}
\end{table}

\begin{table}[ht]
\caption{Performance in real-world datasets} % title of Table
\centering 
\resizebox{\linewidth}{!}{
\begin{tabular}
{c|c|c
|c|
c|c|c}
\hline
F1 Score/Data      & PolyMnist (2)   
& PolyMnist (3)   
& PolyMnist (4)   & PolyMnist (5)  
& CUB   & Caltech101 \\ \hline
CONCAT        & 0.828          
& 0.937          
& 0.964          
& 0.962  & 0.878          & 0.597        \\ \hline
Linear\_CCA & 0.818
& 0.929 & 0.955 & 0.957 & 0.878 & 0.599 \\
Linear\_GCCA & 0.818
& 0.828 & 0.956 & 0.958 & 0.879 & 0.596 \\
PRCCA  & 0.712
& 0.849 & 0.899 & 0.918 & - & - \\ \hline
MVTCAE        & 0.852          
& 0.901          & 0.964          
& 0.964  & 0.900          & 0.284        \\ \hline
DCCA          & 0.865          
& 0.957          & 0.964          
& 0.938   & 0.805          & 0.604       \\
DCCAE         & 0.868          
& 0.956          & 0.961          
& 0.987   & 0.850          & 0.605    \\ 
DCCA\_PRIVATE & \textbf{0.915} 
& 0.959          & 0.968          
& 0.955  & 0.853 & 0.480        \\ 
NR-DCCA (ours)   & 0.910   
& \textbf{0.970} & \textbf{0.991} 
& \textbf{0.993}  & \textbf{0.921}      
& \textbf{0.625} \\ \hline

DGCCA          & 0.875        & 0.964        
& 0.986        &
0.941   & 0.790        &  0.617       \\
DGCCAE         & 0.879        & 0.960       
& 0.988        & 
0.934   & 0.814          & 0.612       \\ 
DGCCA\_PRIVATE & \textbf{0.907}        
& 0.965        
& 0.969        & 0.969         & 0.864         
& 0.476 \\ 
NR-DGCCA(ours)   & 0.903        
& \textbf{0.971}   
& \textbf{0.991}        & \textbf{0.994} 
& \textbf{0.917}      
& \textbf{0.621} \\ \hline
\end{tabular}
}
\label{table:real}
\end{table}

\subsection{Complexity Analysis}
\label{appendix:Complexity Analysis}
In this section, we compare the computational complexity of different DCCA-based methods.
Assuming that we have data from $K$ views, with each view containing $N$ samples and $D$ feature dimensions, then we have the computational complexity of each method in Table~\ref{table:complexity}.

\begin{table}[h]
\caption{Comparisons of computational complexity against baselines} % title of Table
\centering 
\resizebox{\linewidth}{!}{
\begin{tabular}{c|c|c|c|c}
\hline
& DCCA   & DCCAE   & DCCA\_PRIVATE   & NR-DCCA \\ \hline
Generation of Noise & - & - & - & $O(K*N*D)$ \\
MLP Encoder & $O(K*N*L*H^2)$ & $O(K*N*L*H^2)$ & $O(2*K*N*L*H^2)$ & $O(2*K*N*L*H^2)$\\
MLP Decoder & - & $O(K*N*L*H^2)$ & $O(K*N*L*H^2)$ & -\\
Reconstruction Loss  & - & $O(K*N*D)$ & $O(K*N*D)$ & -                               \\
Correlation Maximization & $O((M*K)^3)$ & $O((M*K)^3)$  & $O((M*K)^3)$  & $O((M*K)^3)$ \\
Noise Regularization   & - & - & - & $O(2*K*(M*K)^3)$  \\ \hline

\end{tabular}
}
\label{table:complexity}
\end{table}

\begin{itemize}
    \item \textbf{Complexity of MLP:}  We will use neural networks with the same MLP structure, consisting of $L$ hidden layers, each with $H$ neurons. Therefore, the computational complexity of one pass of the data through the neural networks can be expressed as $O(N*(D*H+D*M+L*H^2))$. To simplify, we use $O(N*L*H^2)$.
    \item \textbf{Complexity of Corr:} During the process of calculating $Cor$ among $K$ views, three main computations are involved. The calculation complexity of the covariance is $O(N*(M*K)^2$. Second, the complexity of the inverse matrix and the eigenvalues are $O((M*K)^3$. As a result, the computational complexity of calculating $Cor$ can be considered as $O((M*K)^3)$.
    \item \textbf{Complexity of reconstruction loss:} The reconstruction loss, also known as the mean squared error (MSE) loss, has a complexity of $O(N*D)$.
   
    % \item \textbf{RQ5:} How do different parameter settings affect our model performance?
\end{itemize}

\end{document}